\documentclass[arxiv]{melba}

\usepackage{mwe} 
\usepackage{formatfig,natbib}
\usepackage{amsmath,amsfonts}
\usepackage{xargs}
\usepackage{algorithmic}
\usepackage{algorithm}
\usepackage{array}
\usepackage[caption=false,font=normalsize,labelfont=sf,textfont=sf]{subfig}
\usepackage{textcomp}
\usepackage{stfloats}
\usepackage{url}
\usepackage{verbatim}
\usepackage{graphicx}
\usepackage{cite}
\usepackage{amsmath,amssymb,algorithm,algorithmic,url}
\usepackage{mathrsfs}
\usepackage{graphicx}
\usepackage{tabularx}
\usepackage{makecell}
\usepackage{xparse}
\usepackage{float}
\usepackage{bbm}
\usepackage{eso-pic}

\def\E{\mathcal{E}}
\def\D{\mathcal{D}}
\def\N{\mathcal{N}}
\newcommand{\scs}{\fontsize{5.6}{4.5}\selectfont}

\usepackage[dvipsnames]{xcolor}
\definecolor{amber}{rgb}{1.0, 0.75, 0.0}
\definecolor{red}{rgb}{1.0, 0, 0.0}
\definecolor{green}{rgb}{0.0, 1.0, 0.0}
\definecolor{magenta}{rgb}{1.0, 0.0, 1.0}
\definecolor{blue}{rgb}{0.0, 0.0, 1.0}
\definecolor{brickred}{rgb}{0.8, 0.25, 0.33}
\definecolor{cyan}{rgb}{0.0, 1.0, 1.0}
\definecolor{black}{rgb}{0.0, 0.0, 0.0}
\definecolor{grey}{rgb}{0.8,0.8,0.8}
\definecolor{blue-violet}{rgb}{0.54, 0.17, 0.89}
\definecolor{darkmagenta}{rgb}{0.55, 0.0, 0.55}
\definecolor{amaranth}{rgb}{0.9, 0.17, 0.31}
\definecolor{darksienna}{rgb}{0.24, 0.08, 0.08}

\definecolor{myo_green}{RGB}{0, 255, 0}
\definecolor{myo_yellow}{RGB}{227, 209, 0}
\definecolor{myo_maroon}{RGB}{255, 0, 255}

\definecolor{neuro_cyan}{RGB}{0, 255, 255}
\definecolor{neuro_black}{RGB}{33, 33, 33}
\definecolor{neuro_green}{RGB}{0, 200, 0}

\definecolor{prostate_maroon}{RGB}{255, 0, 255}
\definecolor{prostate_green}{RGB}{0, 255, 0}
\definecolor{prostate_yellow}{RGB}{227, 209, 0}

\definecolor{fetal_head_maroon}{RGB}{255, 0, 255}
\definecolor{fetal_head_green}{RGB}{0, 255, 0}
\definecolor{fetal_head_yellow}{RGB}{227, 209, 0}

\melbaid{2026:017}  
\doi{10.59275/j.melba.2026-6d54}
\melbaauthors{Pal and Awate}  
\email{pal.jimut@iitb.ac.in}
\volume{2026}
\firstpageno{327}  
\melbayear{2026}  
\datesubmitted{2025-12-08}  
\datepublished{2026-06-14}  

\melbaspecialissue{Medical Imaging with Deep Learning (MIDL) 2020}
\melbaspecialissueeditors{Marleen de Bruijne, Tal Arbel, Ismail Ben Ayed, Hervé Lombaert}

\ShortHeadings{Learning a Sampling-Free Variational DNN Plugin from Tiny Training Sets to Refine OOD
Segmentation}{Pal and Awate}

\title{Learning a Sampling-Free Variational DNN Plugin from Tiny Training Sets to Refine OOD Segmentation
With Uncertainty Estimation}

\author{
\firstname Jimut B. \surname Pal\aff{1}\orcid{0000-0002-1206-7902},
\name Suyash P. \surname Awate\aff{1,2}\orcid{0000-0002-4945-9539}
}
\affiliations{
\num 1 \addr Centre for Machine Intelligence and Data Science (C-MInDS), Indian Institute of Technology (IIT) Bombay, Mumbai \\
\num 2 \addr Computer Science and Engineering (CSE) Department, Indian Institute of Technology (IIT) Bombay,
Mumbai }

\abstract
{
Deep neural networks (DNNs) frequently fail to generalize to out-of-distribution (OOD) medical images
because of variations in scanners and acquisition protocols.
Retraining DNN models to address these distribution shifts is often impractical due to the high cost of
acquiring and annotating new medical datasets.
To address this, we introduce VarDeepPCA, a novel lightweight variational DNN framework designed to
restore/refine degraded segmentation maps by leveraging intrinsic geometric priors.
Unlike existing approaches that require target-domain data or extensive pre-training, our VarDeepPCA
explicitly learns a distribution of valid anatomical geometries using only small in-distribution (ID)
datasets.
Theoretically, our novel variational learning framework leverages a reinterpretation of the softmax mapping
to implicitly perform exact distribution modeling, thereby enabling computationally efficient, sampling-free
learning and inference. This also enables VarDeepPCA to provide uncertainty estimates associated with its
restored segmentation maps.
We empirically validate our framework across 4 distinct clinical applications, using 14 publicly
available datasets, involving segmentation of the myocardium, neuroretinal rim, prostate, and fetal
head. Comparisons against 15 existing methods demonstrate that VarDeepPCA consistently restores
segmentation maps produced by the existing methods on OOD data to (i)~significantly improve anatomical
plausibility of geometries and clinical utility of the segmentations, and (ii)~significantly reduce
errors, without needing any more training data than that used by existing methods.
}

\keywords { Out-of-distribution images, segmentation refinement, plugin, geometric prior learning, small
training set, sampling-free variational learning, uncertainty. }

\begin{document}
\AddToShipoutPictureBG*{
	\AtPageUpperLeft{
		\put(\LenToUnit{0.5\paperwidth},\LenToUnit{-1.0cm}){
			\makebox(0,0)[c]{
				\begin{tabular}{c}
					\normalsize \textbf{Published in Journal of Machine Learning for Biomedical Imaging (MELBA).} \\
					\normalsize \textbf{Link: \textcolor{blue}{\underline{\url{https://www.melba-journal.org/papers/2026:017.html}}}}
				\end{tabular}
			}
		}
	}
}

\twocolumn[\maketitle]

\section{Introduction}

\begin{figure*}[!t]
\centering
%
\fourAcrossLabelsHeight[0]{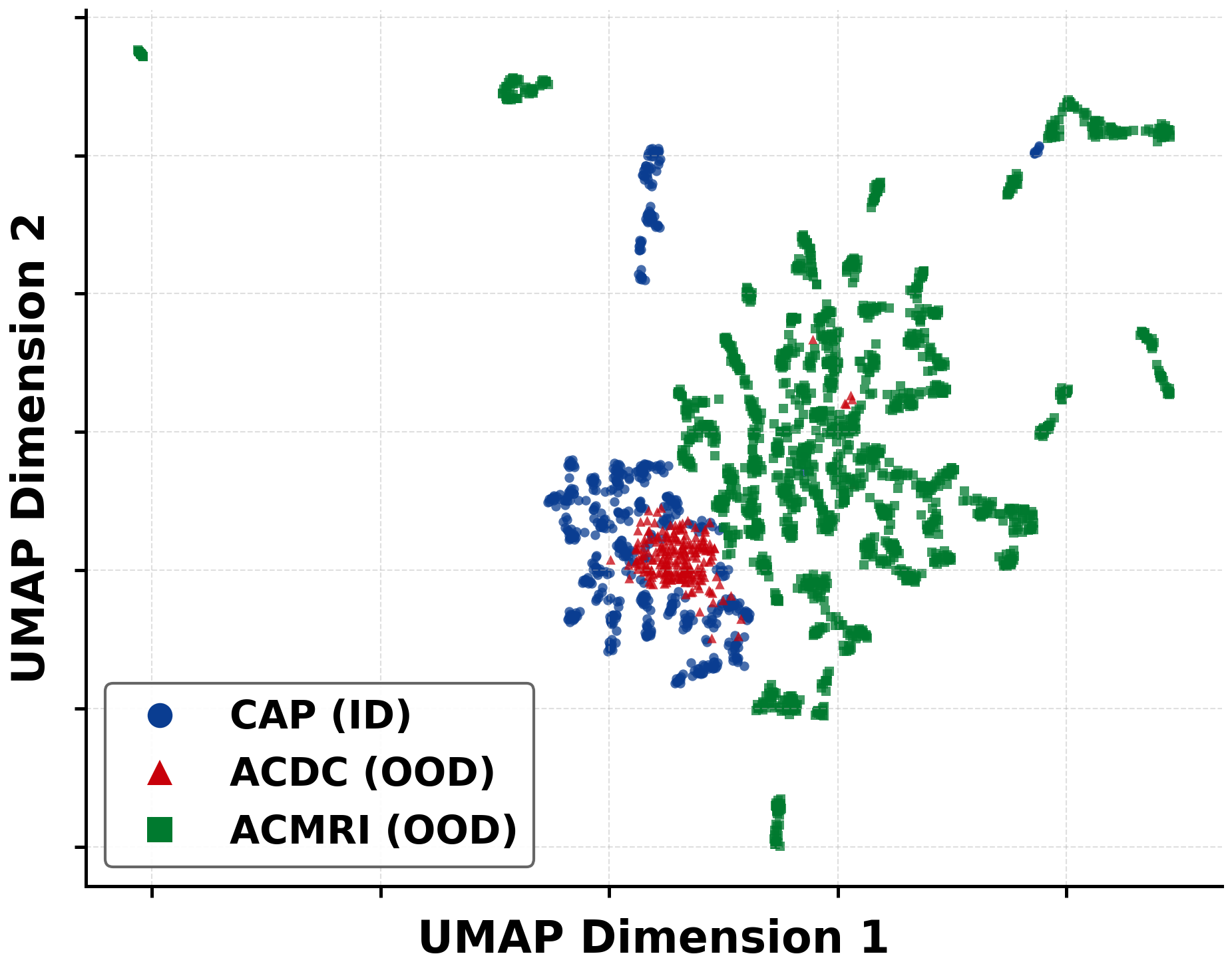}{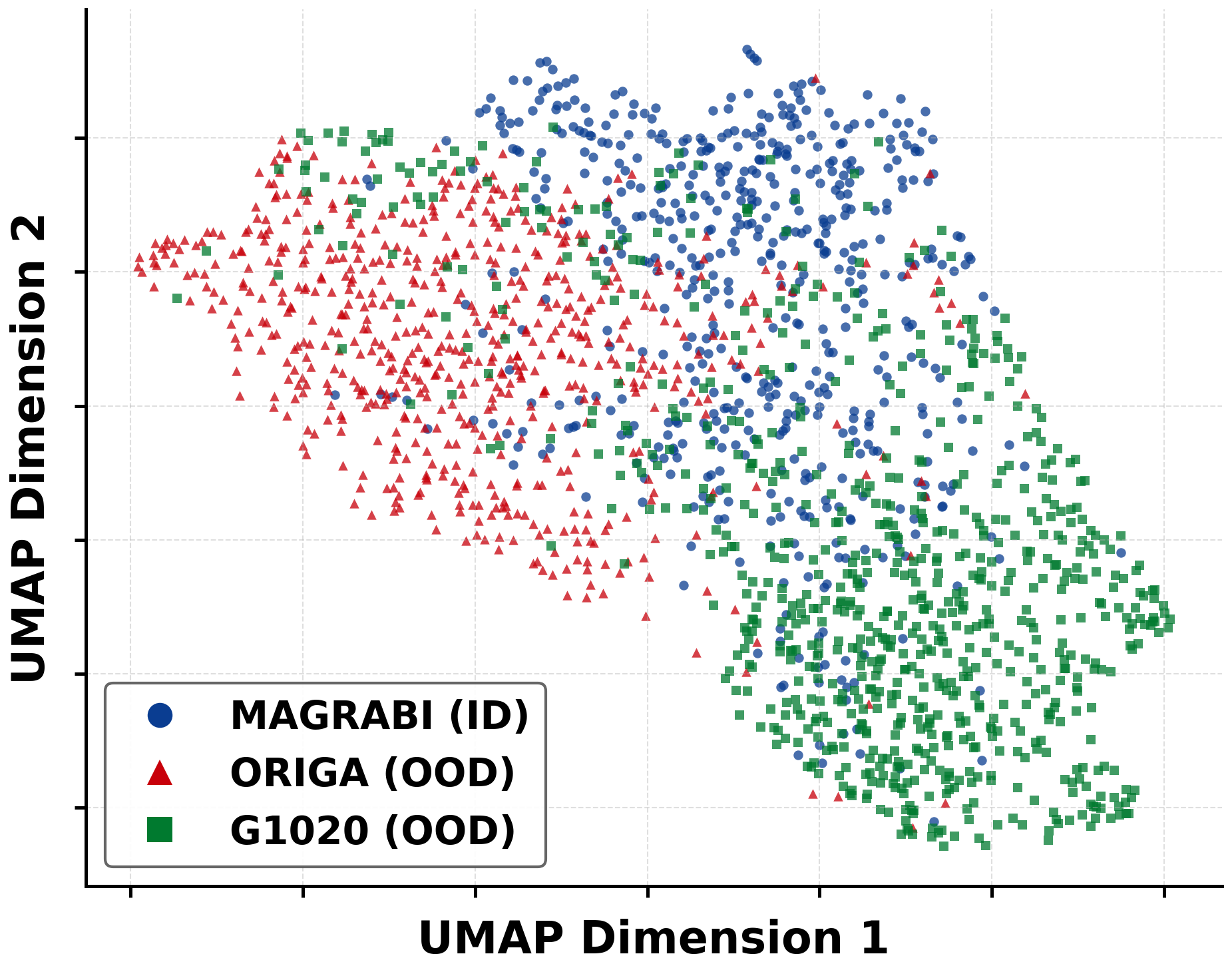}
{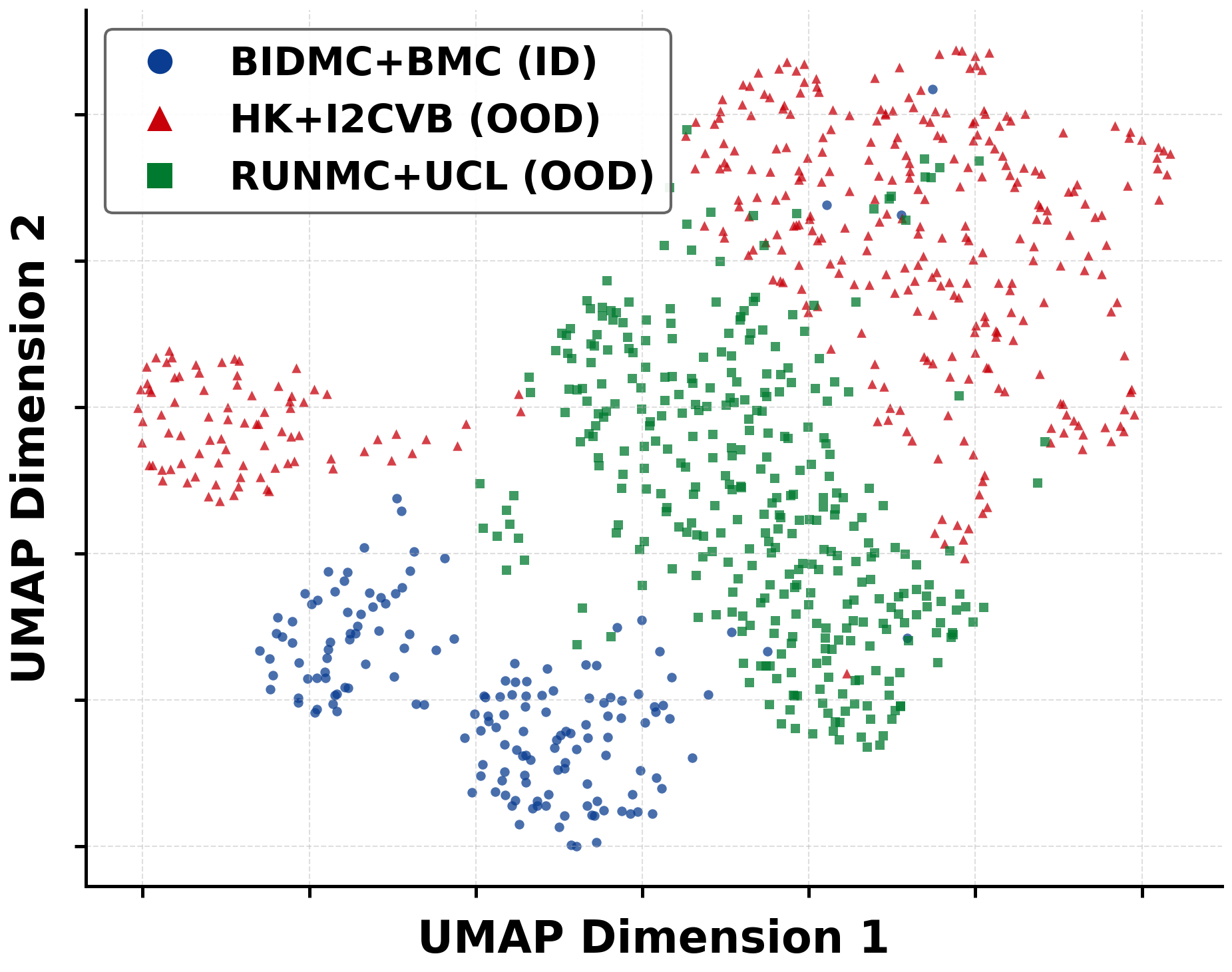}{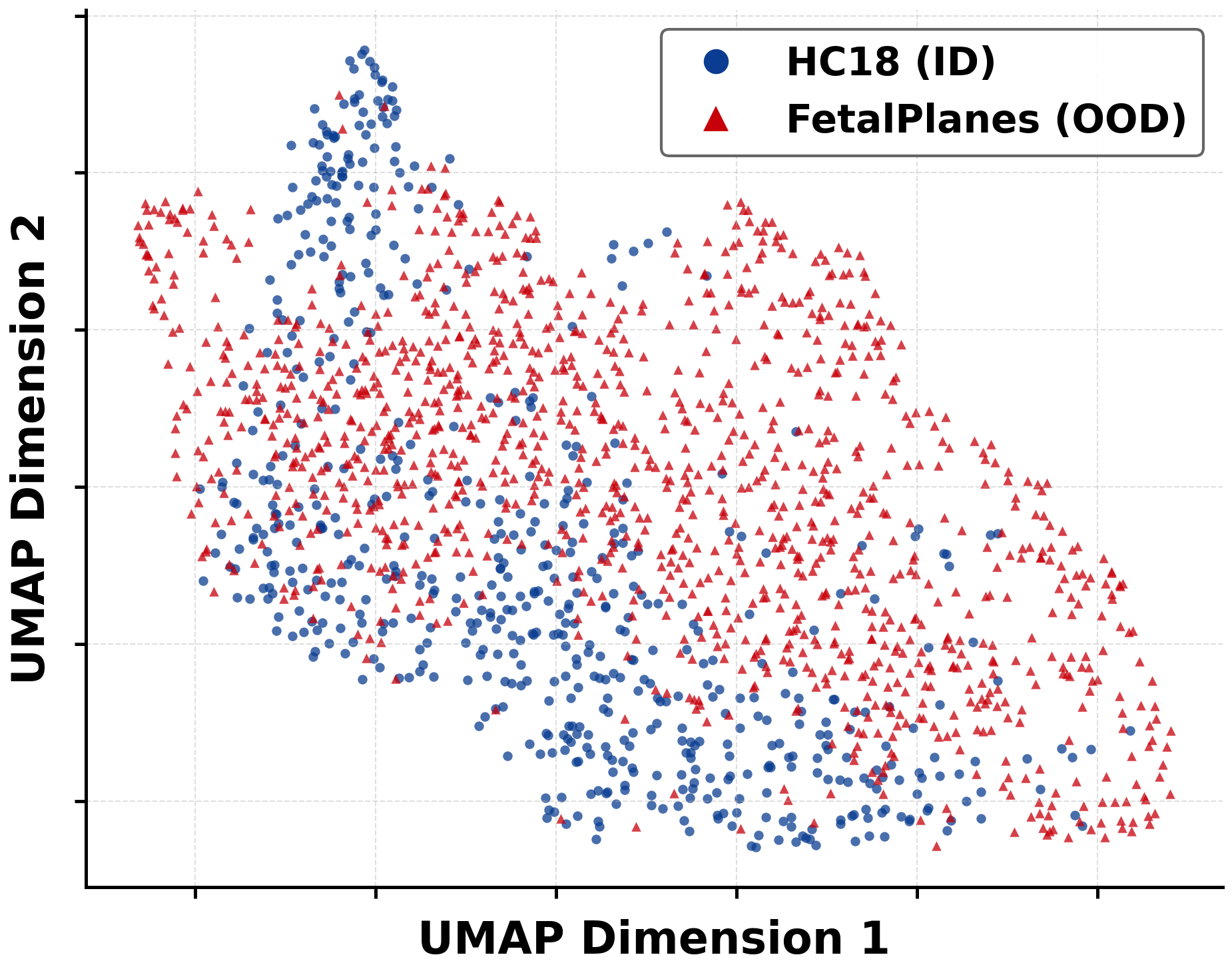}
{\bf(a)~Cardiac MRI}{\bf(b)~Retinal Images}
{\bf(c)~Prostate MRI}{\bf(d)~Fetal Ultrasound}
\caption
{ UMAP \citep{umap} projections of InceptionV3 features of (pooled) ID and OOD image sets across four
medical applications, demonstrating shifts between distributions of ID and OOD images.
Within each application, the details of the specific ID and OOD image sets appear later in
Section~\ref{sec:datasets}.
Each point represents a single acquired medical image, projected from a 2048-dimensional feature space
to 2 dimensions via UMAP.
} 
\label{fig:OOD}
\end{figure*}

\enluminure{D}{eep} neural networks (DNNs) exhibit significant performance degradation when applied to
out-of-distribution (OOD) data \citep{WhatOODis,TranNeurips2020Tutorial}.
Our study addresses OOD data as it appears in real-world clinical scenarios. We define {\em OOD images} to be
of the same anatomical object present in our training/in-distribution (ID) images but acquired from hospitals
or scanners different from those associated with the training images. In this cross-site setting, the
inevitable variations in imaging protocols, devices, and reconstruction schemes give rise to {\em distribution
shifts} \citep{r2_liang_survey_tta,r1_karani, pal2024hard}, as seen in Figure~\ref{fig:OOD}, posing a critical barrier to
the {\em clinical deployment} of DNNs.
OOD data are assumed \citep{WhatOODis,TranNeurips2020Tutorial,r2_liang_survey_tta,r1_karani, MSAttnNet} to be unavailable
during DNN training; retraining is often impractical due to the high cost of acquiring and annotating new
data.

We consider segmentation tasks with a single object of interest per image, i.e., segmenting foreground versus
background; our framework may also be extended to multi-class segmentation problems involving multiple kinds
of objects of interest in an image.
We propose a method to correct the degraded outputs of existing DNNs on OOD data by leveraging {\em priors} on
the {\em inherent geometry} of the object of interest, which remains largely invariant to the aforementioned
OOD variations.
Let a {\em segmentation map} be an image where each pixel value lies within $[0,1]$, representing the
probability of that pixel belonging to the object of interest.
Our novel DNN framework learns the principal modes of variation from ID segmentation maps to explicitly model
a distribution on the manifold of valid anatomical object geometries. This learned low-dimensional
distribution then serves as a powerful geometrical prior to rectify erroneous segmentations produced by
existing DNNs on OOD images.

We apply our method to four diverse medical applications spanning multiple imaging modalities and having
significant diagnostic relevance, i.e., segmenting
(i)~the myocardium from cardiac magnetic resonance images (MRIs),
(ii)~the neuroretinal rim from retinal scans,
(iii)~the prostate from T2-weighted MRI, and 
(iv)~the fetal head from ultrasound images; for which accurate segmentation are clinically vital. 
For instance, {\em myocardial segmentation} in short-axis cardiac
MRI \citep{VentricularCardiacMRI,SegCardiacMRISurvey,SegShortAxisMRI,Wang2015CMIG} is essential for estimating
contractility and tissue strain, which aids in diagnosing infarction, ischemia, and ventricular
dyssynchrony \citep{peng2016review}.
In ophthalmology, optic disc and cup segmentation from retinal scans, i.e., the {\em neuroretinal rim},
enables the calculation of the cup-to-disc ratio \citep{DuaneBook,Lu2011TMI}, a key biomarker for monitoring
glaucoma \citep{almazroa2015optic}.
Similarly, {\em prostate} segmentation in T2-weighted MRI is pivotal for the diagnosis, staging, and treatment
planning of prostate cancer \citep{claus2004pretreatment}.
Finally, {\em fetal head} segmentation in ultrasound images facilitates the measurement of geometric
parameters to assess fetal growth and detect developmental anomalies \citep{zeng2022efficient}.
Since each of these applications demands high-quality segmentation for patient
care \citep{zeng2022efficient,almazroa2015optic,peng2016review}, our work focuses on improving the quality of
(OOD) segmentations to aid the subsequent clinical analysis.

We differentiate our method from a majority of existing segmentation-refinement methods that require the
distribution-shifted data (which they call ``OOD'' data) during their training. Hence, such existing methods
are inapplicable in our OOD setting where distribution-shifted data are unavailable during DNN training.
Accordingly, we compare our method to only those existing approaches that train exclusively on ID data.

Conventional DNNs often require extensive training on large annotated datasets.
Unlike natural images, which are abundant, and relatively easily acquired and annotated, medical images are
typically scarce because they require expensive hardware for acquisition and specialized domain knowledge for
accurate annotation.
Hence, to mitigate this dependency on large training datasets, our DNN framework relies on a {\em lightweight
architecture}. Our DNN trains on a {\em small but representative dataset} of only 100-200 pairs of medical
images and their corresponding segmentation maps, using data-augmentation methods common in medical image
analysis. This highlights our model's capability to learn from tiny training sets, an advantage in clinical
settings.

This paper introduces VarDeepPCA, a novel lightweight variational DNN framework designed to restore/refine
degraded segmentation maps by leveraging intrinsic geometric priors on the anatomical object of interest.
Unlike existing approaches that require target-domain data or extensive pre-training, our VarDeepPCA
explicitly learns a distribution of valid anatomical geometries using only small ID datasets.
Our DNN framework learns the {\em principal modes of variation} in a class of segmentation maps, and models
each segmentation map using a low-dimensional mixture-of-modes latent representation on a simplex.
Theoretically, our novel {\em variational} learning framework leverages a reinterpretation of the softmax
mapping to implicitly perform exact distribution modeling, thereby enabling computationally efficient, {\em
sampling-free} learning and inference.
This also enables VarDeepPCA to provide {\em uncertainty} estimates associated with its restored segmentation
maps.
We empirically validate our framework across 4 distinct clinical applications, using 14 publicly available
datasets, involving segmentation of the myocardium, neuroretinal rim, prostate, and fetal head. Comparisons
against 15 existing methods demonstrate that VarDeepPCA consistently restores segmentation maps produced
by the existing methods on OOD data to (i)~significantly improve anatomical plausibility of geometries and
clinical utility of the segmentations, and (ii)~significantly reduce errors, without needing any more training
data than that used by existing methods.

The rest of the paper is organized as follows.  Section~\ref{sec:related_work} provides a comprehensive
literature review in the segmentation domain, covering the evolution of DNN architectures and loss functions,
variational learning, uncertainty estimation, use of anatomical shape priors, and use of test-time-adaptive
methods for generating robust segmentations.  Section~\ref{sec:methods} describes our proposed method, the
mathematical notations, the PCA-based latent representation, the variational interpretation of the softmax
mapping, sampling-free variational learning, and demonstrates the use of our VarDeepPCA framework to improve
existing segmenters on OOD images.  Section~\ref{sec:results_discussions} talks about the datasets, evaluation
metrics, clinical utility of segmentation maps, baseline methods, implementation details, and the extensive
results and discussions of both segmentation and uncertainty estimation on the four medical applications.
Section~\ref{sec:conclusion} concludes the work by discussing the merits, some of the constraints in our
framework, and future directions.

\section{Related Works in Image Segmentation}
\label{sec:related_work}

DNNs have become the state of the art for many applications medical image segmentation. However, several
challenges remain, particularly in achieving robustness to OOD data, ensuring anatomical plausibility,
maintaining computational efficiency and providing uncertainty estimates, all while training on a small
sample set. This section reviews existing methods in the aforementioned contexts.

\subsection{Evolution of DNN Architectures and Loss Functions}

{\bf Early DNN Methods for Image Segmentation.}
%
%
U-Net \citep{UNet} (here referred to as UNet) employs skip connections from the encoder to fuse context from
the decoder with precise localization of anatomical objects.
Attention U-Net (AttnUNet) \citep{AttnUNet}, a variant of UNet uses gating modules to focus on relevant
regions during training,
ResUNet \citep{ResUNet} uses residual units \citep{He_ResNet}, which is then used as a backbone in
ResUNet++ \citep{ResUNetPP}.
DeepLabV3+ improves upon DeepLab \citep{Chen2018IEEETPAMI} architecture, enabling computationally efficient
training.
ResUNet++ uses squeeze-and-excitation \citep{Squeeze_Excitation_Networks} modules along with atrous spacial
pyramid pooling (ASPP) modules introduced in DeepLabV3+ \citep{DeeplabV3P_Chen} to create a parameter
efficient segmentation architecture.
However, such early DNNs typically lead to poor performance on OOD data because of their relatively
lightweight architectures, straightforward loss functions, and the absence of prior modelling and
pre-training \citep{unet_domain_shift,domain_shift_ood,pretraining_robustness, pal2024advancing}.

{\bf Hybrid Loss Functions for Boundary Enhancement.}
Many DNN methods discussed earlier rely on a single loss term, e.g., binary cross-entropy (BCE) or
soft-Dice \citep{galdranDICE_BCE_Combo}.
Some later DNNs combine multiple loss terms \citep{BASNet_Qin, DoULoss, Kervadec2021MedIA} to focus on
boundary regions. Some methods focus on the difference between the predicted boundary and the ground-truth
segmentation, thereby, focusing on the loss around the segment boundary by employing difference-over-union
(DoU) loss \citep{DoULoss}. Other methods use a boundary loss term by representing a non-symmetric L2 distance
on the space of boundaries/contours as a regional integral.
BASNet \citep{BASNet_Qin} uses a hybrid loss of BCE, structural-similarity (SSIM) \citep{SSIM_Loss_Old,
SSIM_Loss_Zhao}, and intersection over union (IoU) \citep{Loss_Survey_Jadon}.
However, these DNNs are computationally heavier and often require pre-training on ImageNet-1K
\citep{ImageNet_Dataset} dataset. Furthermore, these methods lack the knowledge of high-level segment
characteristics related to the segmented object's geometry or topology \citep{prior_robustness,
  gaikwad2023deep, varma2023adversarial, pal2024miccai_convex, gaikwad2024deep}.

%
{\bf Generative Adversarial DNNs for Image Segmentation.}
Some methods use generative adversarial networks (GANs) \citep{GAN_Goodfellow} towards medical image
segmentation \citep{GAN_Seg_review}, primarily as a generative model for data
augmentation \citep{GAN_Style_Tansfer_Segmentation}.
SegAN \citep{SegAN_Sue} aims to improve segmentations by employing a critic to amplify the differences between
the predicted/generated segmentations and the associated ground-truth segmentations. SegAN uses multi-scale L1
losses to capture both long-range and short-range spatial dependencies.
SegAN aims to solve a min-max optimization problem, involving significant optimization challenges that are
well known~\citep{GAN_Opt_Challenges}.

{\bf Diffusion Models for Image Segmentation.}
Some recent approaches leverage diffusion processes for medical segmentation \citep{DiffReview}.
MedSegDiff \citep{MedSegDiff_Wu} trains using denoising diffusion probabilistic models
(DDPMs) \citep{DiffusionModel_Ho} with dynamic conditional encoding to mitigate
noise.
MedSegDiffV2 \citep{MedSegDiffV2} extends MedSegDiff to segment multi-class objects via a transformer-based
spectrum-space-former architecture.
CIMD \citep{Ambiguous_Med_Diffusion} improves distribution modeling by using multiple expert annotations.
DTAN \citep{DTAN} incorporates text attention on diffusion for segmentation.
However diffusion models for medical image segmentation demand substantial data and computational resources
for optimization, while being more sensitive to hyperparameters \citep{MedSegDiff_Wu,DiffusionModel_Ho}. This
makes OOD segmentation challenging for diffusion models \citep{zhang2025diffuseg,xie2025towards}.

%
{\bf Transformers for Image Segmentation.}  Transformers, leveraging
self-attention \citep{vaswani2017attention} mechanisms, have been applied to image
segmentation \citep{semantic_seg_vit, TransformerSurvey}. Segmenter \citep{segmenter} employs vision
transformers (ViT) \citep{ViT_Dosovitskiy} and processes image patches. SegViT \citep{seg_vit} generates
semantic segmentation masks through attention-to-mask modules. SegViT trains to translate learnable class
tokens and spatial feature maps into segmentation maps. Some transformer architectures for medical image
segmentation domain follow a U-shaped \citep{trans_med_seg} structure.
DSTransUNet \citep{DSTransUNet} improves upon TransUNet \citep{chen2024transunet}\footnote{The original
article appeared on arXiv in 2021 (\url{https://arxiv.org/abs/2102.04306})} by employing hierarchical
swin transformers to capture non-local and multiscale dependencies. It uses dual-scale encoders to extract
coarse and fine-grained features across different semantic classes.
These models are inherently bulky, and need pre-training on large datasets like
ImageNet \citep{ImageNet_Dataset,transformer_pretraining,pinto2021vision}.

{\bf Foundational Models using Vision Transformers.}
Segment Anything Model (SAM) \citep{SAM_Kirillov} leverages foundational models, modeling segmentation as a
promptable task, enabling zero-shot transfer across a wide range of applications.
While SAM demonstrates strong overall performance, it often misses fine structures, occasionally hallucinates
small disconnected components, and may fail to produce crisp boundaries \citep{SAM_Kirillov,
schiappa2024robustness, zhang2024improving}. SAM often needs a lot of human interaction, through text
prompts and image annotations, during inference. Furthermore, SAM was trained on a massive dataset of over one
billion image-mask pairs, making training such models computationally expensive.
MedSAM \citep{SAM_Medical} extends SAM to medical images, but continues to share SAM's limitations.
%
%
%

{\bf State-Space Models for Image Segmentation.}
State-space models (SSMs) \citep{kalman1960new}, inspired by linear state-space equations in control theory,
have recently emerged as an efficient alternative to transformers for modeling long-range dependencies,
scaling linearly with sequence length.
%
%
This includes pure SSM-based methods such as the Vision Mamba UNet (VM-UNet) \citep{ruan2024vmunet}, which
performs better than hybrid SSM-CNN models such as U-Mamba \citep{ma_umamba_2024u},
SwinU-Mamba \citep{liu2024swinmamba}, and SegMamba \citep{xing2024segmamba}.
%
%
However, VM-UNet often struggles with low-contrast regions, is sensitive to image artifacts, and degrades its
segmentation performance with higher image resolution.

\subsection{Variational Learning and Uncertainty Estimation}
\label{rel:uncertainty_quant}

Variational learning in DNNs models distributions in latent space during learning, e.g., the variational
autoencoder (VAE) \citep{VAE_Paper}, vector-quantized VAE (VQ-VAE) \citep{vq_vae_van2017neural}, and
VQ-VAE2 \citep{vqvae_2}.  Conditional VAEs (cVAEs) \citep{cvaes} leverage latent variables to model
conditional distributions on the output. Typical variational-learning methods introduce significant
computational overhead, requiring expensive Monte-Carlo sampling \citep{vae_expensive_mcmc} during both
training and inference.
Probabilistic UNet (ProbUNet) \citep{probUNet} learns a distribution over segmentation maps for a given
input image by combining UNet with cVAEs. Hierarchical ProbUNet (HierProbUNet) \citep{hierarchicalprobUNet}
and PHiRec \citep{r9_fischer_uncertainty} employ a cVAE with a hierarchical latent space. Probabilistic
Hierarchical Segmentation (PHISeg) \citep{phiseg} uses a VAE framework to model the conditional distribution
of segmentation maps, for a given input image, when learning from multiple annotators at different spatial
resolutions.
However, many variational methods fail to provide robustness to OOD data \citep{OOD_bayesian_failure,
  pal2021holistic}, because such models are not explicitly designed for domain shifts
\citep{mehrtash2020confidence,gaikwad2021deep,lennartz2023segmentation}.

Uncertainty-aware methods aim to output per-voxel uncertainty, often modeled as the standard deviation of
a distribution on output segmentations, e.g., \citep{r8_adiga_anatomical_uncertainty}.
Some such approaches use Bayesian modeling and inference \citep{jena2019bayesianMRF,jena2019bayesian}.
Some methods quantify uncertainty estimates using normalized cross correlation (NCC) between the error maps
and the uncertainty maps \citep{r9_fischer_uncertainty}. Another quantitative measure is the unified score
(US; the higher the better) \citep{r10_mehta_qubrats} that combines area-under-curve (AUC) values of
segmentation-performance measures (e.g., filtered true positive, filtered true negative, Dice similarity)
with respect to segmentation-map thresholds. Some metrics focus on uncertainty calibration, e.g., adaptive
calibration error (ACE) and thresholded ACE (TACE) \citep{r11_nixon_calibrationDL}, extending expected
calibration error (ECE) \citep{ece_calibration} using adaptive binning for robustness.

\subsection{Anatomical Shape Priors}


Some DNNs incorporate anatomical information by designing loss terms \citep{Rueckert_ACNN,
Rueckert_ShapeConditionedSegISBI} that entail training on cross-domain data. However, in our (OOD)
setting, data from the newer domain is unavailable during training.
Some models require anatomical landmarks during inference \citep{CrossDomainRepair,
Rueckert_ShapeConditionedSegSTACOM}, which must be provided either by an expert (who is unavailable our
setting) or by a separate DNN (which itself would require training on the OOD data that is unavailable in
our setting).
Post-DAE \citep{r7_larrazabal_postDAE} enforces anatomical plausibility by projecting the predicted
segmentation maps (degraded) generated on ID test data onto a manifold of valid shapes.
Some existing methods leveraging anatomical priors, while not requiring OOD data, suffer from significant
{\em computational and memory} limitations, e.g., a recent VAE-based approach \citep{Nathan_Anatomical}
learns valid cardiac shapes but requires empirically sampling millions of shapes by filtering them with a
domain-specific correctness test, and storing (millions of) the valid ones. At inference it must perform an
expensive search to find the nearest stored shape.
Similarly, pointset-based shape priors \citep{ShapePrior} are often limited by Gaussian assumptions and
expensive brute-force searches.
Our framework avoids such limitations; it does not require any OOD data or annotations during training; by
utilizing sampling-free variational learning, it avoids the need for simulation, storage, or any expensive
inference-time search. Instead, it efficiently optimizes for the closest plausible segmentation using its
DNN decoder via gradient descent on the space of valid segmentation-map geometries. Our model also provides
uncertainty estimates, unlike existing methods using shape-priors for image segmentation.

\subsection{Unsupervised Domain Adaptation}

Existing DNN segmenters often fail in OOD domain because their models over-sensitive to the texture and
spurious correlations within acquired medical images for generating the final predictions \citep{r1_karani}.
To mitigate such effects, global intensity non-linear augmentation (GIN) and interventional
pseudo-correlation augmentation (IPA) methods modify image appearance to break non-causal background
links \citep{r6_ouyang_causality}.
Unsupervised domain adaptation (UDA) methods typically rely on aligning source data to the target data,
without access to target-data labels.
Source-free UDA (SFUDA) \citep{r3_source_free_DA} methods, which are trained to adapt to unlabelled targets
by using pre-trained models, are categorized into parameter-based white-box methods and output-based
black-box methods. Methods such as DeY-Net learn features to encode the anatomy of medical objects for
single domain generalization (SDG).
Test-time adaptation (TTA) methods \citep{r2_liang_survey_tta} adapts to target data using pre-trained
models without access to source data and target-data labels, e.g., denoising TTA (DeTTA)
\citep{r4_denoising_tta}.
Adaptive Mutual Information (AdaMI) \citep{r5_bateson_da} adapts to test data by fine-tuning models using
label-free loss terms and anatomical class-ratio priors.
Recently, SegCNN \citep{r1_karani} adapts to OOD images using image-to-image normalization and
DAEs. However, its performance degrades if the simulated noise during DAE training has a mismatch with the
actual degradations in OOD images, or if the distribution shift renders the image-to-image normalization
module ineffective \citep{r1_karani}.

\begin{figure*}[!t]
\centering
\includegraphics[width=0.98\textwidth]{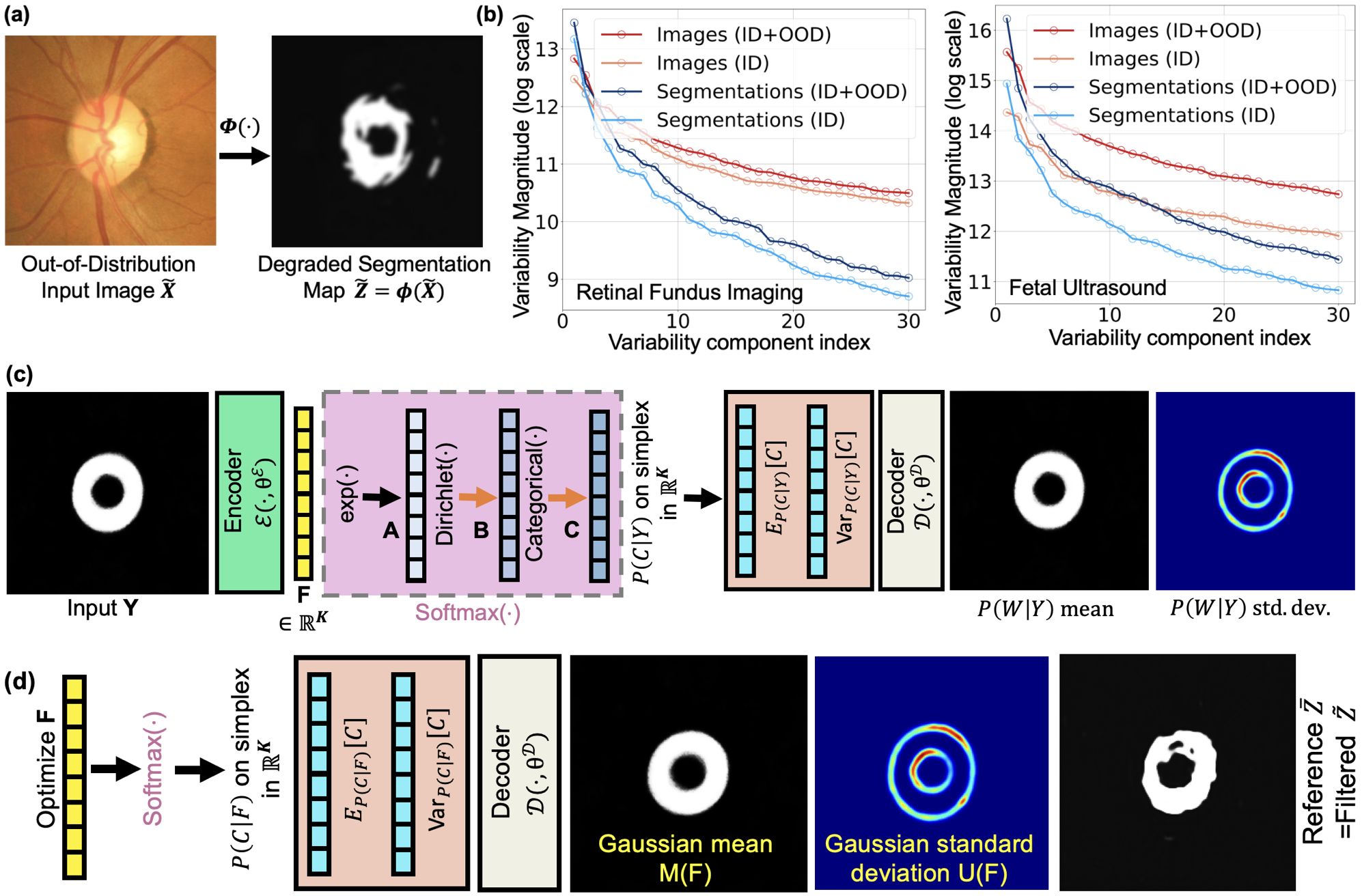}
\caption
{
{\bf (a)}~Existing DNNs segment objects poorly on OOD images.
{\bf (b)}~Principal (log) eigenvalues of covariance matrices of encodings in Inception
DNN \citep{InceptionNet_Szegedy,FID} (in ID sets, as well as ID-union-OOD sets) show the variability
across segmentation maps to be far lower (at least by an order of magnitude) than that across medical
images (image intensities mapped to the range $[0,1]$).
{\bf (c) Sampling-Free Variational Deep Learning of Principal Modes of Variation (VarDeepPCA) in
Segmentation Maps.}
Our VarDeepPCA models and learns principal modes of variation of (ID) segmentation maps. It models each
segmentation map using a low-dimensional mixture-of-modes latent distribution on a simplex
(Section~\ref{sec:simplex}) through a softmax mapping (Section~\ref{sec:softmax}).
Our Bayesian interpretation of the softmax endows a variational model with (i)~closed-form marginalization
enabling sampling-free variational learning (Section~\ref{sec:train}) and (ii)~per-pixel uncertainty
estimates on test images (Section~\ref{sec:train}).
{\bf (d)}~{\bf VarDeepPCA Restores/Refines OOD-Image Segmentation Maps} by first ``filtering'' the
degraded segmentation map, and then ``projecting'' the filtered segmentation map onto VarDeepPCA's learned
principal modes of variation (Section~\ref{sec:restore}).
}
\label{fig:framework}
\end{figure*}

\subsection{Extensions Over Our Preliminary Work}

This paper is a significant extension of our preliminary work in \citep{Pal2025WACV}, advancing both its
theoretical formulation and empirical analysis.
The novel contributions of this extended work are as follows:
(i)~we propose a new mathematical formulation for the learning objective that explicitly models the output
distribution's variance, unlike the formulation in \citep{Pal2025WACV} which disregarded this per-pixel
uncertainty;
(ii)~we validate our method on more clinical applications: prostate segmentation in T2-weighted MRI and fetal
head segmentation in ultrasound, incorporating a total of eight more datasets into our empirical analysis;
(iii)~we provide a new empirical sensitivity analysis for key hyperparameters, i.e., the size of the latent
dimension $K$ and the size of the training set $S$ as discussed in Section~\ref{subsec:sen}, and show that our
VarDeepPCA method performs robustly in all of the settings;
(iv)~we add seven new additional baselines for comparison, including the VMUNet (state-space model), the
MedSAM foundational model specifically tuned for medical image segmentation and three new probabilistic
baselines, i.e., Probabilistic UNet, Hierarchical Probabilistic UNet, and PHISeg, which produce both
segmentation maps and their uncertainty estimates, along with the test-time-adaptation-based SegCNN method
(with and without atlas), which is specifically designed to handle OOD data during test-time;
(v)~we add a comparative study for quantitative evaluation of the clinical utility of the uncertainty
estimates produced by existing methods and our VarDeepPCA framework, using NCC, US, and TACE;
(vi)~we provide a more comprehensive discussion of the related works and methodology, and a more detailed
presentation of the results including extensive qualitative and quantitative examples.

\section{Methods}
\label{sec:methods}

Our work is centered on a novel variational encoder-decoder DNN framework, namely {\em VarDeepPCA}, designed
to learn a robust statistical model of geometry variability from ID segmentation maps.
This learned model is then leveraged as a powerful prior to correct poor segmentations produced by existing
DNNs on OOD data.
The VarDeepPCA framework comprises: 
(i)~a decoder that models the non-linear principal modes of anatomical geometry variation and their mixtures;
(ii)~a low-dimensional latent distribution that models these mixture proportions for a given segmentation map; 
(iii)~an encoder that maps an input segmentation map to its corresponding latent distribution; and 
(iv)~a {\em sampling-free} scheme for variational learning and inference.
After learning this statistical geometry model, we employ a {\em fully automatic} correction process to
improve the degraded segmentations produced by existing DNNs on OOD images.

\subsection{Model Components and Mathematical Notation}
\label{sec:problem_setup}

Let $X$ denote an acquired medical image containing an object of interest, and let $Y$ be the associated
expert-annotated (binary/fuzzy) segmentation map.
We differentiate this from $W$, which represents the (unknown) true anatomical segmentation. 
For a single given $X$, our framework is designed to handle cases with a single $Y$ or multiple expert
segmentations $\{ Y_i \}$, each of which may differ from $W$.
We consider an existing DNN segmenter $\Phi(\cdot)$, e.g., UNet, which is pre-trained on an ID dataset of
$(X, Y)$ pairs, with standard data-augmentation techniques.
Our primary challenge arises when this segmenter is applied to an OOD image $\widetilde{X}$.
By definition, $\widetilde{X}$ is drawn from a different distribution than the data used to train
$\Phi(\cdot)$ stemming from domain shifts (Figure~\ref{fig:OOD}) because of the variation in imaging
protocols across different hospitals and in imaging devices (e.g., different make/models, slight variations
in calibration for identical instruments) \citep{r2_liang_survey_tta}. The domain shift may manifest as
differences in the texture/appearance statistics \citep{r1_karani}.
This distribution shift often causes the segmenter to produce a poor or anatomically implausible segmentation,
$\widetilde{Z} := \Phi(\widetilde{X})$, as illustrated in Figure~\ref{fig:framework}(a).
Our goal is to correct $\widetilde{Z}$ without access to any OOD images $\widetilde{X}$ or their expert
segmentations $\widetilde{Y}$ during the training of our VarDeepPCA model. In contrast, VarDeepPCA relies
solely on the segmentation maps that were used to train $\Phi(\cdot)$, by constructing and employing priors on
the anatomical-object geometry that remains largely invariant to these OOD variations, as illustrated in
Figure~\ref{fig:framework}(b). Indeed, the OOD description applies much more to the acquired medical images,
resulting from variations in imaging protocols, devices, and reconstruction schemes across clinical sites,
rather than variations in human anatomical geometry across the population. Figure~\ref{fig:framework}(b)
quantifies this variability in acquired medical images as well as the segmentation maps through the
eigenvalues of the covariance matrix of latent Inception-DNN encodings.

\subsection{DNN-based PCA Model on Segmentation Maps Using a Latent Representation on a Simplex}
\label{sec:simplex}

The VarDeepPCA framework employs a generic encoder-decoder DNN architecture to learn a statistical model of
variability in segmentation maps for a class of objects (Figure~\ref{fig:framework}(c)).
We hypothesize that this variability can be captured by $K$ {\em principal (non-linear) modes of variation}.
%
%
In a latent space, we represent these discrete modes by a $K$-length {\em one-hot} random vector $C$, where
$C = \mathbbm{1}_k$ indicates the $k$-th mode, with a value of 1 at index $k$.
The spatial-domain representation of these modes of variation is enabled by the mapping underlying our
decoder.
A typical segmentation map $Y$ (in spatial domain) is not associated with a single mode (in latent space) but
rather a {\em mixture} of these $K$ principal modes.
Therefore, we design VarDeepPCA to encode each $Y$ into a {\em latent distribution} $P(C|Y)$.
This distribution is represented by a $K$-dimensional vector of probabilities,
$\left[ P(C=\mathbbm{1}_1 | Y), \dots, P(C=\mathbbm{1}_K | Y) \right]$, which defines the association of $Y$
with each of the $K$ modes.
As this vector's elements are non-negative and sum to 1, it is constrained to lie on a $(K-1)$-dimensional
{\em simplex} in $\mathbb{R}^K$.
To obtain this simplex representation, VarDeepPCA's encoder $\E(\cdot; \theta^{\E})$ first maps an input
segmentation map $Y$ to a $K$-dimensional {\em segmentation-feature} vector $F := \E(Y; \theta^{\E})$.
We then apply the {\em softmax} function to $F$ to produce the probability vector representing $P(C|Y)$. 
This use of the softmax is a critical design choice because it implicitly performs the necessary
variational/distribution modeling (as described in Section~\ref{sec:softmax}) and, crucially enables {\em
sampling-free} variational learning (as described in Section~\ref{sec:train}).
Finally, VarDeepPCA's decoder $\D(\cdot; \theta^{\D})$ maps the latent simplex representation $P(C|Y)$ back
to a distribution $P(W|Y)$ on the segmentation maps.

\subsection{Reinterpreting Softmax Mapping in a Variational Setup}
\label{sec:softmax}

VarDeepPCA reinterprets the softmax function underlying the mapping $P(C|Y) \equiv \text{Softmax}(F)$ using
Bayesian principles to illuminate the implicit variational modeling and distribution on the simplex in
$\mathbb{R}^K$ (Figure~\ref{fig:framework}(c)).
We model $P(C|F)$ as the \emph{posterior-predictive distribution} on $C$ arising from
(i)~a \emph{Categorical}-distribution \emph{likelihood} $P(C | \cdot)$ on the modes of variation indicated by
$C$, coupled with
(ii)~a \emph{Dirichlet}-distribution (conjugate) \emph{prior} $P(\cdot | F)$.
Let the random vector $A$ have elements $A_k := \exp(F_k) > 0$ for all $1 \le k \le K$, such that $A$
parameterizes a Dirichlet distribution $\text{Dir}(B; A)$ of a hidden random vector $B$ residing on the
$(K-1)$-dimensional simplex.
Since the mapping from $Y \to F \to A$ is deterministic, the following equivalence between
posterior-predictive distributions holds:
$P(C | Y = y) \equiv P(C| F = \E(y; \theta^\E)) \equiv P(C| A = \exp(\E(y; \theta^\E)))$.
Consider a categorical distribution $\text{Cat}(C; B)$ on one-hot vectors $C$, which is parameterized by the
hidden random vector $B$ that is sampled from its {\em conjugate} (prior) distribution $\text{Dir}(B; A)$.
The posterior-predictive distribution
\begin{flalign}
P (C | A)
=
\int_{b} P (C | b) P (b | A) db 
\end{flalign}
which equals
\begin{flalign}
\int_{b} \text{Cat} (C; b) \text{Dir}(b; A) db
\end{flalign}
which equals
\begin{flalign}
\int_{b}
\bigg(
\prod_k (b_k)^{C_k}
\bigg)
\bigg(
\frac{1}{\eta(A)} \prod_k (b_k)^{A_k - 1}
\bigg)
db
,
\end{flalign}
where the normalizing constant for the Dirichlet distribution is
$\eta(A) := \prod_k \Gamma(A_k) / \Gamma(\sum_k A_k)$, and $\Gamma(\cdot)$ denotes the Gamma function. This
yields
\begin{flalign}
P (C | A)
&
=
\frac{1}{\eta(A)}
\int_{b}
\prod_k (b_k)^{C_k+A_k-1} db
=
\frac {\eta(A+C)} {\eta(A)}
\\
&
=
\frac
{\prod_k \Gamma(A_k + C_k) / \Gamma\left(\sum_k (A_k + C_k)\right)}
{\prod_k \Gamma(A_k) / \Gamma(\sum_k A_k)}
.
\end{flalign}

Now, consider a specific instance $C = \mathbbm{1}_k$ (i.e., the $k$-th mode). Because $C$ is a one-hot vector,
$\sum_k C_k = 1$. Using the property $\Gamma(g+1) = g \Gamma(g)$, for gamma functions, we simplify the
posterior-predictive distribution as
\begin{flalign}
&
P (C = \mathbbm{1}_k | A)
\nonumber
\\
&
=
\frac
{ \Gamma(A_k+1) \prod_{j \neq k} \Gamma(A_j) / \Gamma(\sum_j A_j + 1) }
{ \prod_j \Gamma(A_j) / \Gamma(\sum_j A_j) }
\\
&
=
\frac
{ A_k \Gamma(A_k) \prod_{j \neq k} \Gamma(A_j) }
{ \prod_j \Gamma(A_j) }
\cdot
\frac
{ \Gamma(\sum_j A_j) }
{ (\sum_j A_j) \Gamma(\sum_j A_j) }
\\
&
=
\frac
{ A_k }
{ \sum_{j=1}^K A_j }
=
\frac
{ \exp (F_k) }
{ \sum_{j=1}^K \exp (F_j) }
,
\end{flalign}
which is the $k$-th element of the $\text{Softmax}(F)$ vector.
Thus, while the softmax mapping from $F$ to the latent distribution $P(C|Y)$ is deterministic, it implicitly
(i)~subsumes variational modeling by defining the (prior) distribution $P(B|\exp(F)) \equiv P(B|A)$ and the
(likelihood) distribution $P(C|B)$, and then
(ii)~marginalizes out the random variable $B$ via Bayesian inference to produce the \emph{analytically exact}
posterior-predictive distribution $P(C | F)$ in \emph{closed form}.

\subsection{Sampling-Free Variational Learning}
\label{sec:train}

Let the training set of $N$ segmentation maps be $\{ Y_n \}_{n=1}^N$.
For an input segmentation map $Y$, VarDeepPCA's internal low-dimensional representations ($F$ and $P(C|F)$)
are designed to model $Y$ using only the top $K$ modes of variation, thereby filtering out the remaining
variation that arises from sources such as segmentation errors and discretization artifacts in $Y$.
This is because the low-dimensional ($K$-dimensional) latent space in our autoencoder acts as a
(well-studied) bottleneck with limited capacity \citep{laakom2024reducing}, which forces our autoencoder to
model/represent mainly those dominant shapes/structures of the segmentation maps that were present in its
training set of high-quality segmentation maps \citep{cho2013simple,creswell2018denoising}.
For a given input $Y$, the variational model underlying VarDeepPCA produces a latent distribution $P(C|Y)$ by
implicitly modeling the categorical distribution $P(C|B)$ and the Dirichlet distribution
$P(B | Y=y) \equiv P(B | A = \exp(\E(y; \theta^{\E}))$.
This enables VarDeepPCA to sample $c \sim P(C|Y)$ through the following procedure: (i)~map input $Y$ to
$F \leftarrow \E(Y; \theta^{\E})$, (ii)~map $F$ to $A \leftarrow \exp(F)$, (iii)~sample
$b \sim \text{Dir}(B; A)$, and (iv)~sample $c \sim \text{Cat}(C; b)$.
The decoder then {\em outputs a distribution over segmentation maps} by mapping the latent distribution
$P(C|Y)$ through the decoder $\D(\cdot)$.
Specifically, the decoder maps each $c \sim P(C|Y)$ to a segmentation map, where we ensure that each per-pixel
output lies within the range $[0,1]$ by incorporating a sigmoid layer as the final output layer of the
decoder.
For the (posterior-predictive) categorical distribution $P(C|Y)$, the mean and variance are available
analytically in closed form:

\begin{flalign}
C^{\text{mean}}
&
:= \mathbb{E}_{P(C|Y; \theta^\E)} [C]
\\
&
= [ P(C=\mathbbm{1}_1 | Y), \ldots, P(C=\mathbbm{1}_K | Y) ]
\\
&
= \text{Softmax}(\E(Y; \theta^\E))
\end{flalign}
(as per Section~\ref{sec:softmax}), and the $k$-th element of the variance is given by
\begin{flalign}
C^{\text{var}}_k := C^{\text{mean}}_k (1 - C^{\text{mean}}_k).
\end{flalign}
We model the decoder-output distribution by propagating the mean and variance of $P(C|Y)$ through the decoder,
and approximating the output as a Gaussian distribution $\N(\cdot)$ characterized by a {\em mean segmentation
map} $M := \D(C^{\text{mean}}; \theta^\D)$ and a {\em variance map} $V$, which we describe next.
Let $\D_i(\cdot)$ denote the decoder mapping to the $i$-th pixel.
For pixel $i$ in $V$, we model the variance $V_i$ using (i)~the variances $C^{\text{var}}_k$ and (ii)~the
Jacobian of the decoder mapping $\D(L; \theta^\D)$ (where $L$ is a dummy variable) evaluated at
$C^{\text{mean}}$. Thus,
\begin{flalign}
M
&
:= \D(C^{\text{mean}}; \theta^\D) \text{, and}
\\
V_i
&
:=
\sum_{k=1}^K
C^{\text{var}}_k
\bigg(
\frac{\partial \D_i(L)}{\partial L_k}
\bigg|_{L := C^{\text{mean}} = \text{Softmax}(\E(Y; \theta^\E))}
\bigg)^2.
\end{flalign}
Our choice of modeling the output distribution as Gaussian stems from the Gaussian being the maximum-entropy
(most general, in a sense) \citep{CoverThomas} distribution across all distributions constrained by a fixed
mean $M$ and a fixed variance $V$.
From an alternative perspective, VarDeepPCA's output can be interpreted as (i)~the representative segmentation
$M$ together with (ii)~an underlying per-pixel {\em uncertainty} $U$ (Figure~\ref{fig:framework}(c)) given by
the per-pixel square root of the values in $V$.

We formulate the variational learning objective to maximize, over parameters $\theta$, the likelihood of the
observed reference segmentation $Y$ under the Gaussian distribution $\N(\cdot; M,V)$ output by the decoder.
VarDeepPCA's variational learning formulation is therefore
\begin{flalign}
&
\arg \max_\theta
\prod_{n=1}^N
\N(Y_n; M(Y_n;\theta), V(Y_n;\theta))
\equiv
\\
&
\arg \min_\theta
\sum_{n=1}^N
\sum_{i=1}^I
\frac{(Y_{ni} - M_i(Y_n;\theta))^2}{V_i(Y_n;\theta) + \epsilon}
+
\log(V_i(Y_n;\theta) + \epsilon),
\end{flalign}
where $M_i(\cdot)$ and $V_i(\cdot)$ denote, respectively, the values at the $i$-th pixel in the mean
segmentation map $M$ and the variance map $V$; $\epsilon > 0$ is a small regularization parameter for
numerical stability.
Thus, our VarDeepPCA learning formulation, despite explicitly modeling (i)~a latent distribution $P(C|Y)$ and
(ii)~distributions $P(C|B)$ and $P(B|Y)$ implicitly within the softmax parameterization, eliminates the need
for Monte Carlo sampling and the associated reparameterization that becomes necessary in typical variational
deep networks (e.g., VAEs) due to the intractability of their underlying integrals.

\subsection{VarDeepPCA to Improve Existing Segmenters on OOD Images}
\label{sec:restore}

We propose a novel two-stage algorithm (Figure~\ref{fig:framework}(d)) to leverage the learned VarDeepPCA
model for restoring the poor segmentation maps $\widetilde{Z}$ produced by existing DNNs $\Phi(\cdot)$ on OOD
images $\widetilde{X}$.
In the first stage, we pass $\widetilde{Z}$ through the encoder-decoder of VarDeepPCA to ``filter'' out the
non-principal components of variability from $\widetilde{Z}$, producing the ``filtered'' segmentation map
\begin{flalign}
\overline{Z} := \D(\text{Softmax}(\E(\widetilde{Z}; \theta^\E)); \theta^\D).
\end{flalign}
In the second stage, we explicitly ``project'' $\overline{Z}$ onto the learned space of principal modes of
variation by (i)~fixing $\overline{Z}$ as the output reference, (ii)~optimizing the segmentation-feature
vector in $\mathbb{R}^K$ as
\begin{flalign}
F^* := \arg \max_F \N(\overline{Z}; M(F;\theta^\D), V(F;\theta^\D))
\end{flalign}
using gradient ascent, and (iii)~obtaining the restored segmentation
\begin{flalign}
M^* := \D(\text{Softmax}(F^*); \theta^\D)
\end{flalign}
with the associated per-pixel uncertainties given by $U^*_i := \sqrt{V^*_i}$ (Section~\ref{sec:train}).

In summary, VarDeepPCA introduces a principled variational framework that learns non-linear geometrical priors
from ID data through a novel simplex-based latent representation, enabling sampling-free inference via softmax
mappings that implicitly perform exact Bayesian marginalization.
The framework's decoder outputs both a mean segmentation and per-pixel uncertainty estimates, providing
interpretable measures of model confidence.
We deploy this learned prior for the fully automatic correction of moderately degraded OOD segmentations by
leveraging gradient-based projection onto the learned manifold of anatomically plausible geometries.
The full procedure is detailed in Algorithm \ref{fig:algorithm}.
We now proceed to empirically validate this framework on multiple clinical applications.

\begin{algorithm}[!t]
\centering
\oneWidth{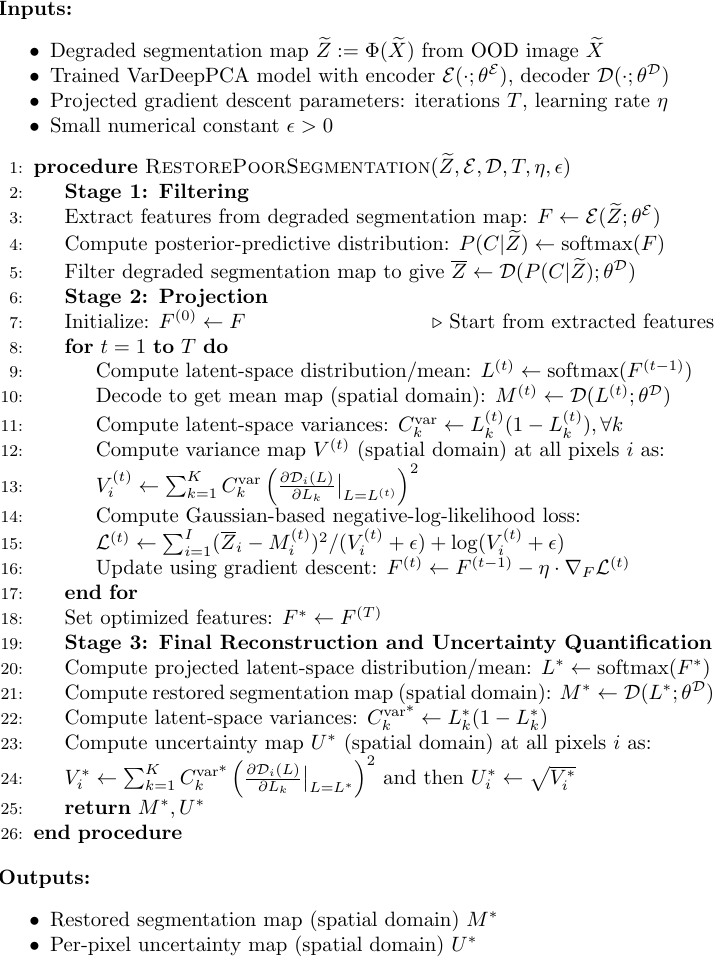}{1}
\caption{VarDeepPCA: Restoring OOD Segmentation Maps with Uncertainty Estimation.}
\label{fig:algorithm}
\end{algorithm}

\section{Results and Discussions}
\label{sec:results_discussions}

\subsection{Datasets}
\label{sec:datasets}

We evaluate our framework across four distinct medical imaging applications:
(i)~segmenting the myocardium in MRI, 
(ii)~segmenting the neuroretinal rim in retinal fundus images, 
(iii)~segmenting the prostate in MRI, and
(iv)~segmenting the fetal head in ultrasound.
These applications span diverse anatomical geometries, including genus-0 and genus-1 topologies.
%
%
For each application, we train all models on a single dataset (ID training set) and test on one or more
separate datasets (i.e., ID test set and OOD dataset).
This cross-dataset evaluation scheme mimics a typical clinical scenario, testing robustness across domain
shifts \citep{r2_liang_survey_tta,r1_karani} caused by variations in imaging equipment, acquisition protocols,
pathologies, etc.
We pre-process images by cropping/padding and resampling image size to 256$\times$256 pixels, applying data
augmentation, and rescaling the intensities to the range $[0,1]$.
An overview of the datasets appears in Table~\ref{tab:datasets_summary}.
%
%

\begin{figure*}[!t]
\centering
\fiveAcrossLabelsHeightFirst
{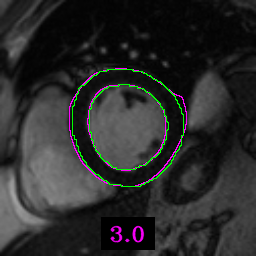}
{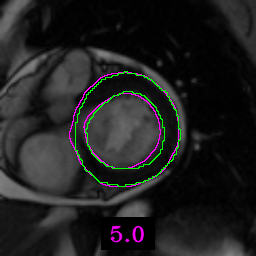}
{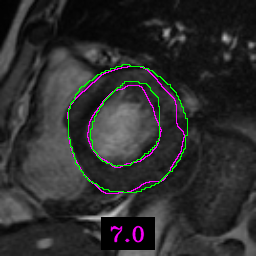}
{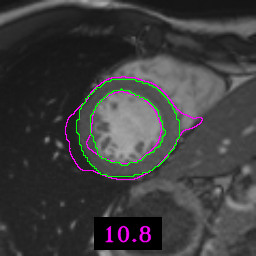}
{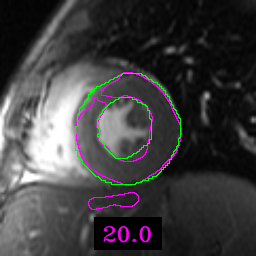}
{\bf \scs (a1)~UNet (ID)}
{\bf \scs (a2)~UNet (ID)}
{\bf \scs (a3)~UNet (ID)}
{\bf \scs (a4)~UNet (OOD)}
{\bf \scs (a5)~UNet (OOD)}
\fiveAcrossLabelsHeightFirst
{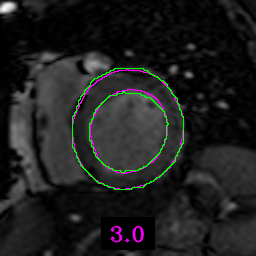}
{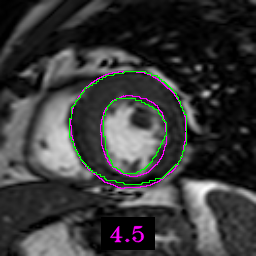}
{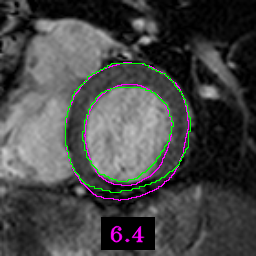}
{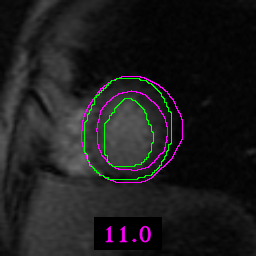}
{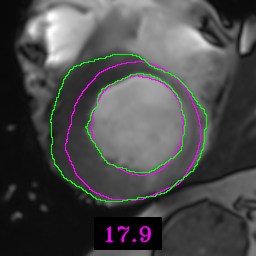}
{\bf \scs (b1)~PHISeg (ID)}
{\bf \scs (b2)~PHISeg (ID)}
{\bf \scs (b3)~PHISeg (ID)}
{\bf \scs (b4)~PHISeg (OOD)}
{\bf \scs (b5)~PHISeg (OOD)}
\fiveAcrossLabelsHeightFirst
{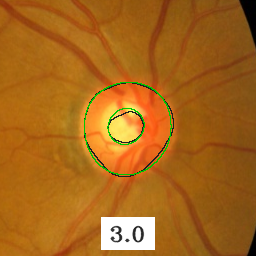}
{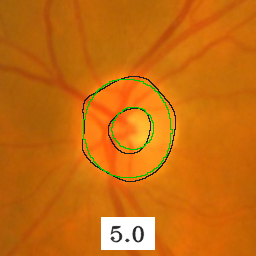}
{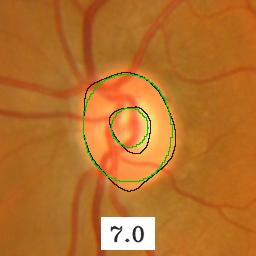}
{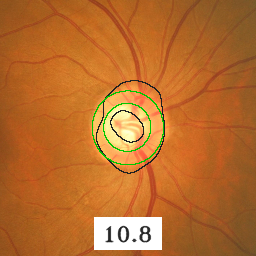}
{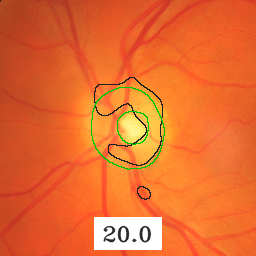}
{\bf \scs (c1)~ProbUNet (ID)}
{\bf \scs (c2)~ProbUNet (ID)}
{\bf \scs (c3)~ProbUNet (ID)}
{\bf \scs (c4)~ProbUNet (OOD)}
{\bf \scs (c5)~ProbUNet (OOD)}
\fiveAcrossLabelsHeightFirst
{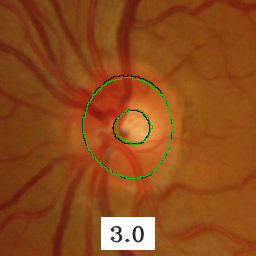}
{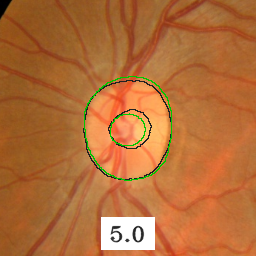}
{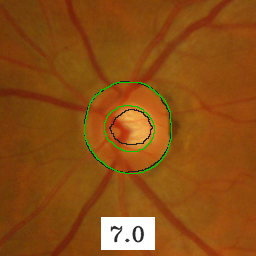}
{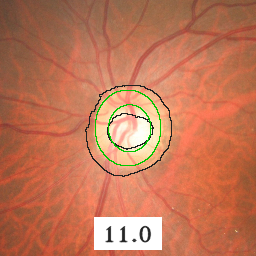}
{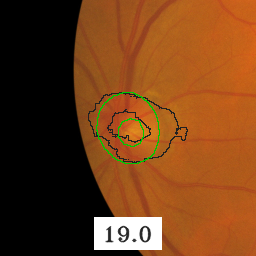}
{\bf \scs (d1)~VMUNet (ID)}
{\bf \scs (d2)~VMUNet (ID)}
{\bf \scs (d3)~VMUNet (ID)}
{\bf \scs (d4)~VMUNet (OOD)}
{\bf \scs (d5)~VMUNet (OOD)}
\fiveAcrossLabelsHeightFirst
{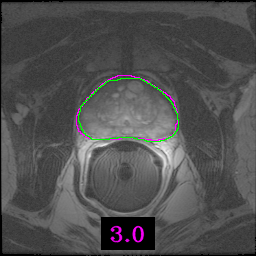}
{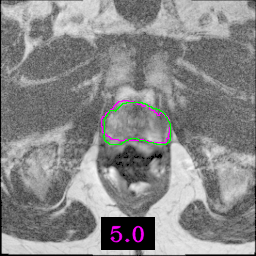}
{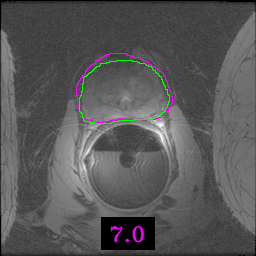}
{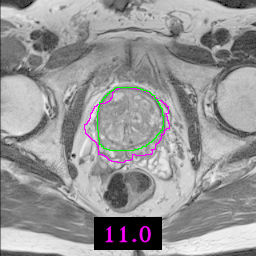}
{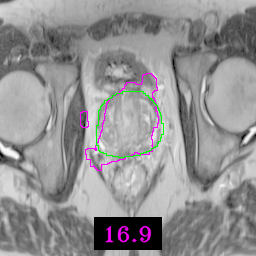}
{\bf \scs (e1)~MedSegDiff (ID)}
{\bf \scs (e2)~MedSegDiff (ID)}
{\bf \scs (e3)~MedSegDiff (ID)}
{\bf \scs (e4)~MedSegDiff (OOD)}
{\bf \scs (e5)~MedSegDiff (OOD)}
\fiveAcrossLabelsHeightFirst
{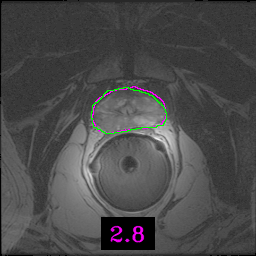}
{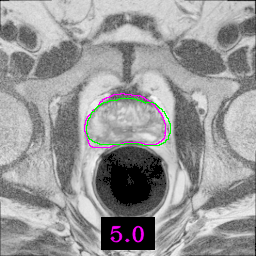}
{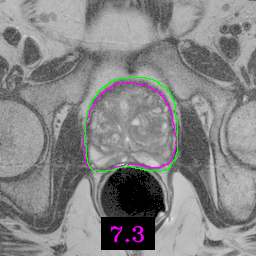}
{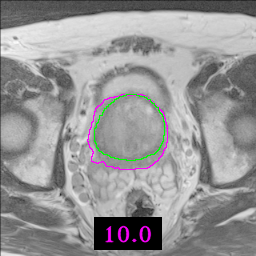}
{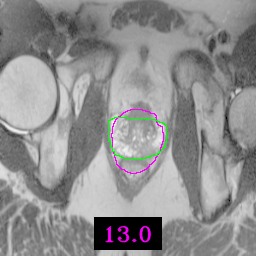}
{\bf \scs (f1)~PHISeg (ID)}
{\bf \scs (f2)~PHISeg (ID)}
{\bf \scs (f3)~PHISeg (ID)}
{\bf \scs (f4)~PHISeg (OOD)}
{\bf \scs (f5)~PHISeg (OOD)}
\fiveAcrossLabelsHeightFirst
{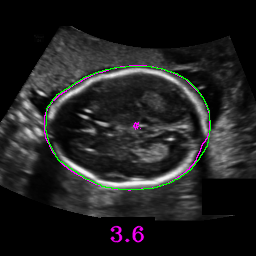}
{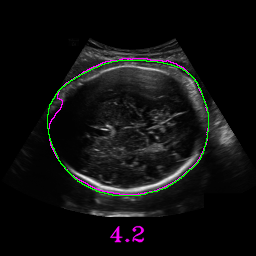}
{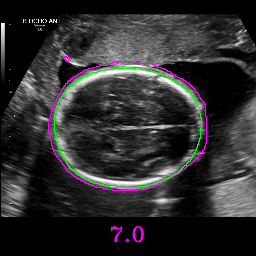}
{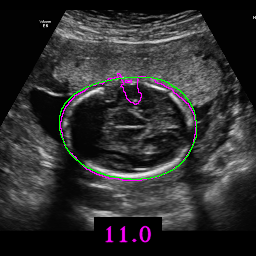}
{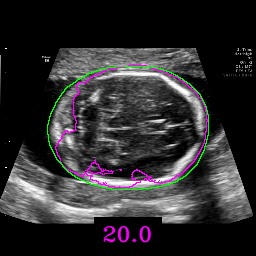}
{\bf \scs (g1)~UNet (ID)}
{\bf \scs (g2)~UNet (ID)}
{\bf \scs (g3)~UNet (OOD)}
{\bf \scs (g4)~UNet (OOD)}
{\bf \scs (g5)~UNet (OOD)}
\fiveAcrossLabelsHeightFirst
{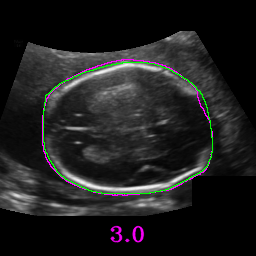}
{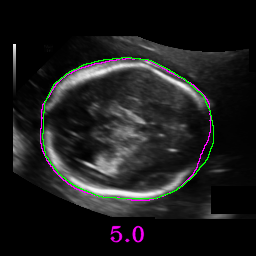}
{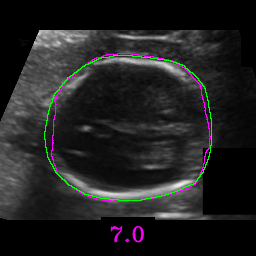}
{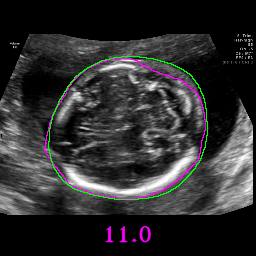}
{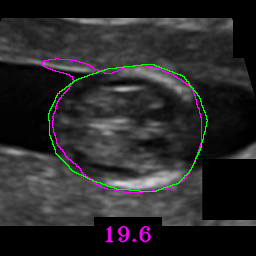}
{\bf \scs (h1)~BASNet (ID)}
{\bf \scs (h2)~BASNet (ID)}
{\bf \scs (h3)~BASNet (ID)}
{\bf \scs (h4)~BASNet (OOD)}
{\bf \scs (h5)~BASNet (OOD)}
\vspace{-8pt}
\caption
{
{\bf Qualitative: Establishing Clinical Utility of Segmentation Maps.}
Segmentation results across diverse baselines for:
(i)~myocardium on {\bf (a1)-(a5)} UNet and {\bf (b1)-(b5)} PHISeg;
(ii)~neuroretinal rim on {\bf (c1)-(c5)} ProbUNet and {\bf (d1)-(d5)} VMUNet;
(iii)~prostate on {\bf (e1)-(e5)} MedSegDiff and {\bf (f1)-(f5)} PHISeg;
(iv)~fetal head on {\bf (g1)-(g5)} UNet and {\bf (f1)-(f5)} BASNet.
Ground-truth segmentation appears in green.
The numbers show HD95 values. 
}
\label{fig:clinical_utility}
\end{figure*}

\begin{table*}[!t]
  \caption{
    {\bf Datasets.} Overview of the datasets used across the four medical imaging applications. 
  }
\centering \renewcommand{\arraystretch}{1.4} \resizebox{0.95\textwidth}{!}{%
\begin{tabular}{@{} p{3.0cm} p{4.2cm} c c c p{4.5cm} p{4.5cm} @{}}
\toprule
\textbf{Application} & \textbf{ID Dataset} & \textbf{ID} & \textbf{ID} & \textbf{ID} & \textbf{OOD Dataset 1} & \textbf{OOD Dataset 2} \\
\textbf{(Modality)} & & \textbf{Train} & \textbf{Val} & \textbf{Test} & \textbf{(Test Size)} & \textbf{(Test Size)} \\
\midrule

Myocardium \newline (MRI) &
CAP~\citep{CAP1,CAP2,CAP3} &
150 & 70 & 634 &
ACDC~\citep{ACDC} (220 samples) &
ACMRI~\citep{Andreopoulos_CMRI} (1722 samples) \\
\addlinespace

Neuroretinal \newline Rim (Fundus) &
Magrabi~\citep{Magrabi} &
150 & 63 & 620 &
ORIGA~\citep{ORIGA} (637 samples) &
G1020~\citep{G1020} (788 samples)\\
\addlinespace

Prostate \newline (MRI) &
BIDMC+BMC~\citep{UCL_BIDMC_HK,RUNMC_BMC_1,RUNMC_BMC_2} &
150 & 18 & 213 &
HK+I2CVB~\citep{UCL_BIDMC_HK,I2CVB} (366 samples) &
RUNMC+UCL~\citep{RUNMC_BMC_1,RUNMC_BMC_2,UCL_BIDMC_HK} (348 samples)\\
\addlinespace

Fetal Head \newline (Ultrasound) &
HC18~\citep{HC18} &
150 & 68 & 666 &
FetalPlanes~\citep{FetalPlanes} (1250 samples)&
-- \\
\bottomrule
\end{tabular}%
}
\label{tab:datasets_summary}
\end{table*}

\textbf{Myocardium.}
We utilize three publicly available short-axis cardiac MRI datasets. 
The CAP dataset \citep{CAP1,CAP2,CAP3}, with 854 images, serves as the ID data, from which we use 150 images
for training.
The remaining data is split into a validation set (10\% of the remainder, 70 images) and the rest (634 images)
as an ID test set.
For OOD evaluation, we employ the ACDC dataset \citep{ACDC} (220 images) and the A-CMRI
dataset \citep{Andreopoulos_CMRI} (1722 images).
%

\textbf{Neuroretinal Rim.}
For the neuroretinal rim segmentation task, we utilize three publicly available retinal fundus image
datasets.
The Magrabi dataset \citep{Magrabi}, with 833 images, serves as the ID data.
From this dataset, we use 150 images for training, 63 for validation, and 620 for ID testing.
For OOD evaluation, we employed the ORIGA dataset \citep{ORIGA} (637 images) and the G1020 dataset
\citep{G1020} (788 images).
%

\textbf{Prostate.}
We use six publicly available T2-weighted MRI datasets from
a multi-institutional collection for prostate segmentation \citep{Qunade_Liu_Multisite}, with a small number
of images per institution.
Within ID and OOD datasets, we desire to have a sufficient number of images and a comparable number of
images. Thus, we design the ID and OOD datasets as follows: (i)~the ID dataset combines BIDMC
\citep{UCL_BIDMC_HK} and BMC \citep{RUNMC_BMC_1, RUNMC_BMC_2}, together having 381 images,
(ii)~the first OOD dataset combines HK \citep{UCL_BIDMC_HK} and I2CVB \citep{I2CVB}, together having 366
images, and
(iii)~the second OOD dataset combines RUNMC \citep{RUNMC_BMC_1, RUNMC_BMC_2} and UCL \citep{UCL_BIDMC_HK},
together having 348 images.
From the ID dataset, we use 150 images for training, 18 images for validation, and 213 images for ID testing.

\textbf{Fetal Head.}
We utilize two publicly available ultrasound datasets.
The HC18 dataset \citep{HC18}, comprising 884 images, serves as the ID dataset.
From this set, we use 150 images for training, 68 images for validation, and 666 images for ID testing.
The FetalPlanes dataset \citep{FetalPlanes}, containing 1250 images, is the OOD test set.

\textbf{Curation of ID Training Set, ID Validation Set, ID Test Set.}
From the ID dataset, we aim to create an ID training subset that has a small size (say, 150) mimicking
clinical scenarios, and that is representative of the diversity of anatomical shapes in the application. For
this purpose, we propose the following strategy. 
First, in the ID dataset, for each binarized ground-truth segmentation map, we extract six features that
collectively characterize the anatomical geometry \citep{clustering_2024}, i.e., region area, bounding-box
area, extent, eccentricity, solidity, and orientation.
Second, in this 6-dimensional feature space, we perform K-means clustering using 50 clusters.
Third, we randomly select 3 segmentation maps from each of the 50 clusters, giving a total of 150 images (and
their acquired medical images) in the ID training subset (used by all methods/baselines).
From the remaining ID data, we use 10\% as the ID validation set (to tune the hyperparameters for all models),
and 90\% as the ID test set.
We train all the baselines, as well as VarDeepPCA, using the training dataset of these curated 150
image-mask pairs.
We use the validation set to find the optimal latent-dimension size (i.e., $K$) for training our VarDeepPCA
framework.

\textbf{Data Augmentation}.
To enhance model generalization and mitigate overfitting, we employ data augmentation
\citep{buslaev2020albumentations} during learning for VarDeepPCA as well as for all the baseline methods.
The augmentation pipeline uses geometric and pixel-level transformations.
Geometric augmentations, specifically horizontal flip, vertical flip, and affine transformations (comprising
rotation, scaling, and translation), apply synchronously to medical images and their corresponding
segmentation maps.
Pixel-level augmentations include random brightness-contrast, random gamma, and blur.

\begin{figure}[!t]
\centering
\fourAcrossLabelsHeightFirst[0]
{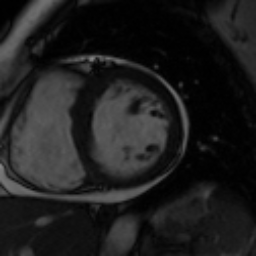}
{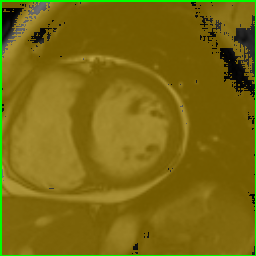}
{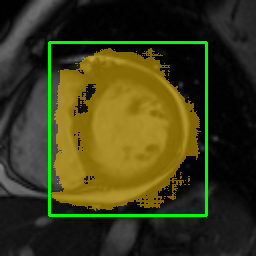}
{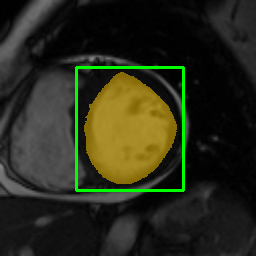}
{\bf \scs (a1)~Cardiac MRI}
{\bf \scs (a2)~Full BBox}
{\bf \scs (a3)~Half BBox}
{\bf \scs (a4)~Small BBox}
\fourAcrossLabelsHeightFirst[0]
{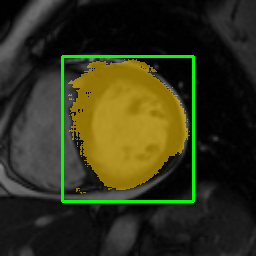}
{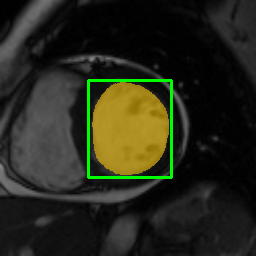}
{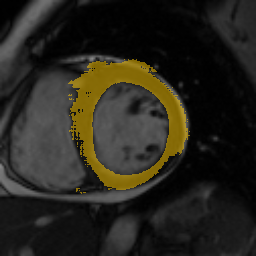}
{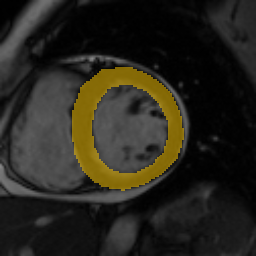}
{\bf \scs (b1)~Epicardium BBox}
{\bf \scs (b2)~Endocardium BBox}
{\bf \scs (b3)~Myocardium}
{\bf \scs (b4)~GT}
\fourAcrossLabelsHeightFirst[0]
{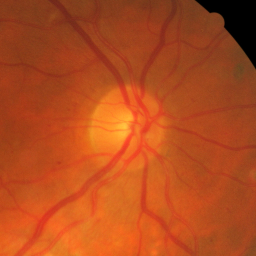}
{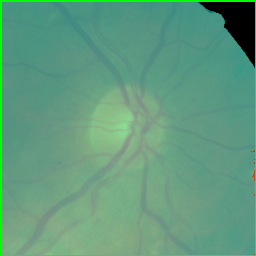}
{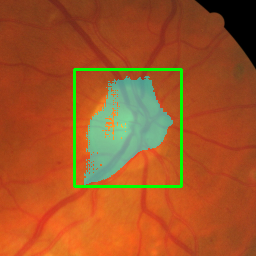}
{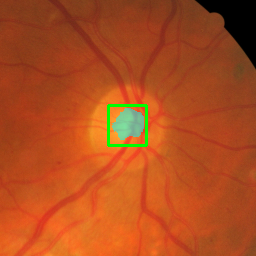}
{\bf \scs (c1)~Fundus image}
{\bf \scs (c2)~Full BBox}
{\bf \scs (c3)~Half BBox}
{\bf \scs (c4)~Small BBox}
\fourAcrossLabelsHeightFirst[0]
{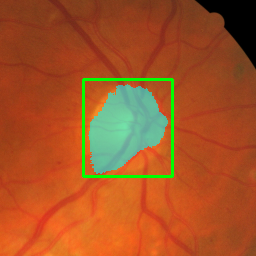}
{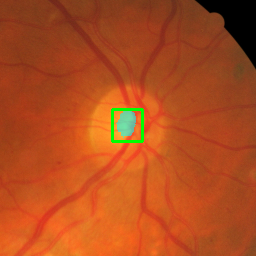}
{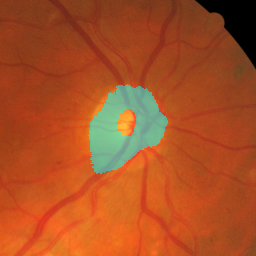}
{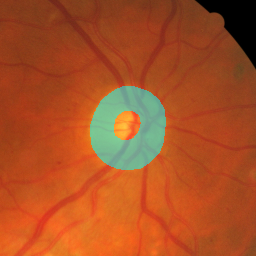}
{\bf \scs (d1)~Optic Disc BBox}
{\bf \scs (d2)~Optic Cup BBox}
{\bf \scs (d3)~Neuroretinal Rim}
{\bf \scs (d4)~GT}
\fourAcrossLabelsHeightFirst[0]
{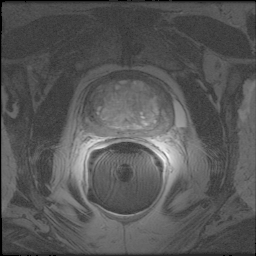}
{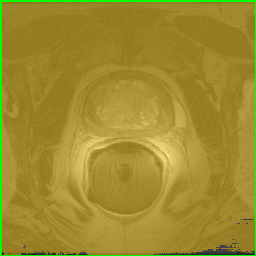}
{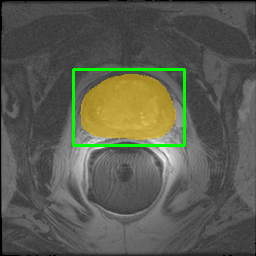}
{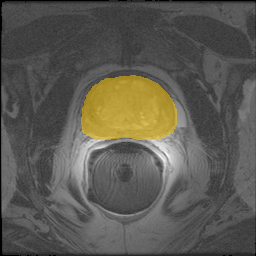}
{\bf \scs (e1)~Prostate MRI}
{\bf \scs (e2)~Full BBox}
{\bf \scs (e3)~Prostate BBox}
{\bf \scs (e4)~GT}
\fourAcrossLabelsHeightFirst[0]
{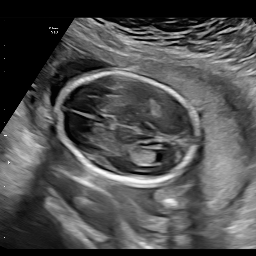}
{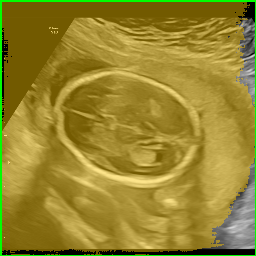}
{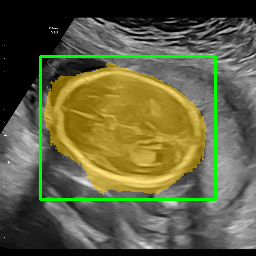}
{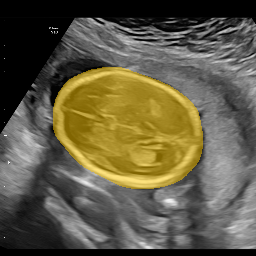}
{\bf \scs (f1)~Ultrasound}
{\bf \scs (f2)~Full BBox}
{\bf \scs (f3)~Fetal Head BBox}
{\bf \scs (f4)~GT}
\caption
{
{\bf MedSAM-based Segmentation using a Bounding Box (BBox) Prompt.}
{\bf (a1)--(b4)}~Segmenting myocardium in cardiac MRI.
{\bf (c1)--(d4)}~Segmenting neuroretinal rim in retinal fundus images.
{\bf (e1)--(e4)}~Segmenting prostate in T2-weighted MRI.
{\bf (f1)--(f4)}~Segmenting fetal head in ultrasound images.
}
\label{fig:medsam_segmentation}
\end{figure}

\begin{table*}[!t]
\caption{{\bf Model Sizes, Training Times, Inference Times, Hyperparameters.}
  The model size (number of parameters; in millions), training time (in minutes), and per-image inference time
  (in milliseconds) are the same across all four applications. MedSegDiff needed many more epochs during
  training, compared to other models. Learning rate (LR), weight decay (WD), batch size (BS), and loss
    functions are for training.  }  \centering \resizebox{1\textwidth}{!}{%
  \footnotesize \renewcommand{\arraystretch}{1.1}
\begin{tabular}{@{}lccccc@{}}
  \toprule
  \textbf{Models} & \textbf{Params (M)} & \textbf{Train. Time (mins)} & \textbf{Inf. Time (ms)} &
                                                                                                  \textbf{Train. Epochs}
  & \textbf{Key Hyperparameters, Loss Functions} \\
  \midrule

UNet & 31.03 & 6.8 & \makebox[3.5em][r]{9.9} $\pm$ \makebox[2.5em][l]{16.7} & 200 & Adam: LR 1e-4, WD 5e-4; \\
\scriptsize\citep{UNet} & & & & & BS 64; Loss: SoftDice \\
\addlinespace

AttnUNet & 34.87 & 9.2 & \makebox[3.5em][r]{10.6} $\pm$ \makebox[2.5em][l]{27.3} & 200 & Adam: LR 1e-4, WD 5e-4; \\
\scriptsize\citep{AttnUNet} & & & & & BS 32; Loss: SoftDice \\
\addlinespace

ResUNet++ & 4.06 & 7.3 & \makebox[3.5em][r]{9.1} $\pm$ \makebox[2.5em][l]{20.2} & 200 & Adam: LR 1e-4, WD 5e-4; \\
\scriptsize\citep{ResUNetPP} & & & & & BS 64; Loss: SoftDice \\
\addlinespace

DeepLabV3+ & 59.33 & 5.6 & \makebox[3.5em][r]{10.1} $\pm$ \makebox[2.5em][l]{23.2} & 200 & Adam: LR 1e-4, WD 5e-4; \\
\scriptsize\citep{DeeplabV3P_Chen} & & & & & BS 64; Loss: SoftDice \\
\addlinespace

BASNet & 87.06 & 21.4 & \makebox[3.5em][r]{18.2} $\pm$ \makebox[2.5em][l]{9.5} & 200 & Adam: LR 1e-4, WD 5e-4; \\
\scriptsize\citep{BASNet_Qin} & & & & & BS 24; Loss: BCE, SSIM, IoU \\
\addlinespace

SegAN & 216.44 & 19.3 & \makebox[3.5em][r]{14.1} $\pm$ \makebox[2.5em][l]{21.3} & 200 & Adam: LR 1e-4, WD 0, $\beta$ 0.999; \\
\scriptsize\citep{SegAN_Sue} & & & & & BS 24; Loss: SoftDice, Adv. \\
\addlinespace

MedSegDiff & 129.40 & 2120.0 & \makebox[3.5em][r]{6.8e4} $\pm$ \makebox[2.5em][l]{150.1} & 10000 & Adam: LR 1e-4, WD 0; EMA: 0.999; \\
\scriptsize\citep{MedSegDiff_Wu,MedSegDiffV2} & & & & & BS 14; Loss: MSE, Calib. \\
\addlinespace

DSTransUNet & 171.44 & 17.8 & \makebox[3.5em][r]{61.2} $\pm$ \makebox[2.5em][l]{29.5} & 200 & Adam: LR 1e-4, WD 5e-4; \\
\scriptsize\citep{DSTransUNet} & & & & & BS 24; Loss: Structure \\
\addlinespace

VMUNet & 44.27 & 21.1 & \makebox[3.5em][r]{20.7} $\pm$ \makebox[2.5em][l]{79.7} & 200 & Adam: LR 1e-4, WD 5e-4; \\
\scriptsize\citep{ruan2024vmunet} & & & & & BS 8; Loss: SoftDice \\
\addlinespace

MedSAM & 93.73 & {\em NA} & \makebox[3.5em][r]{1204.8} $\pm$ \makebox[2.5em][l]{30.3} & {\em NA} & {\em NA} \\
\scriptsize\citep{SAM_Medical} & & & & & \\
\addlinespace

PHISeg & 99.21 & 12.9 & \makebox[3.5em][r]{183.7} $\pm$ \makebox[2.5em][l]{29.2} & 200 & Adam: LR 1e-4; BS 16; \\
\scriptsize\citep{phiseg} & & & & & Loss: Hierarchical KLD, CE \\
\addlinespace

ProbUNet & 5.0 & 7.6 & \makebox[3.5em][r]{21.0} $\pm$ \makebox[2.5em][l]{20.4} & 200 & Adam: LR 1e-4, WD 0; Lat. Dim: 6; \\
\scriptsize\citep{probUNet} & & & & & $\beta$: 10.0; BS 32; Loss: L2, ELBO \\
\addlinespace

HierProbUNet & 65.36 & 22.1 & \makebox[3.5em][r]{199.3} $\pm$ \makebox[2.5em][l]{40.0} & 200 & Adam: LR 1e-4, WD 1e-5; Lat. Dim: 4; \\
\scriptsize\citep{hierarchicalprobUNet} & & & & & BS 8; Loss: BCE, GECO \\
\addlinespace

SegCNN+TTA+DAE & 2.25 & 8.08 & \makebox[3.5em][r]{3703.1} $\pm$ \makebox[2.5em][l]{272.9} & 200 & Adam: Train LR 1e-4, WD 5e-4; BS 32; \\
(+Atlas)~\scriptsize \citep{r1_karani} & & & & & TTA LR 1e-3, steps 1000; Loss: SoftDice \\
\bottomrule
\end{tabular}
}
\label{tab:model_params_time}
\end{table*}

\subsection{Evaluation Metrics}

{\bf Quantifying Segmentation Performance}.
Our evaluation prioritizes boundary-based metrics, which quantify distances between the predicted boundary
points and the ground-truth boundary points.
The Dice similarity coefficient (DSC)~\citep{r12_dice1945,r13_sorensen,r14_zijdenbos_morphometric} is a
popular metric for image segmentation, but it loses sensitivity when boundary-prediction errors are small
relative to the size of the object \citep{dice_issues}.
Since our work focuses on precise object-boundary delineation, instead of localization, we primarily rely on
metrics that are more sensitive to delineation errors \citep{optic_surface}.
Our chosen metrics are the 95th-percentile Hausdorff distance (HD95) \citep{Hausdorff_Distance_Original} and
the average surface distance (ASD) \citep{ASD}; both of these metrics are calculated in pixel distances, and
lower values are better.
These metrics are well-established for segmentation tasks in cardiac MRI \citep{ACDC,CAP1}, retinal images
\citep{optic_surface}, prostate MRI \citep{prostate_surface, Qunade_Liu_Multisite}, and fetal head ultrasound
\citep{fetal_surface1, fetal_surface2}.
We also report DSC (higher is better) in percentage, for reference.

{\bf Quantifying Uncertainty-Estimation Performance}.
We seek uncertainty estimates that correlate with, or are well-calibrated with respect to, actual errors in
segmentation.
For this purpose, we use three metrics (described in Section~\ref{rel:uncertainty_quant}):
(i)~NCC \citep{r9_fischer_uncertainty}, (ii)~US \citep{r10_mehta_qubrats}, and
(iii)~TACE \citep{r11_nixon_calibrationDL}.

\subsection{Establishing Clinical Utility Based on Segmentation-Map HD95}
\label{sec:UpperBoundClinUtil}

We decide on the clinical utility of a given segmentation map generated by the DNN based on its HD95
value with respect to the ground-truth segmentation map, as follows.
First, for the ID dataset in each application, we pool the segmentation results across all baselines and
compute the histogram of the HD95 values.
Subsequently, for each ID dataset, we analyze the histogram of HD95 values to find the typical range of HD95
values that can be considered as a clinically-acceptable performance (assuming most baselines work well on ID
datasets).
Through this quantitative analysis, we find that:
(i)~for the cardiac CAP dataset: HD95 $\leq$ 8 pixels accounted for around 92\% of the segmentations;
(ii)~for the retinal MAGRABI dataset: HD95 $\leq$ 8 pixels accounted for around 72\% of the segmentations;
(iii)~for the prostate BIDMC+BMC dataset: HD95 $\leq$ 8 pixels accounted for around 75\% of the segmentations;
(iv)~for the fetal HC18 dataset: HD95 $\leq$ 8 pixels accounted for around 59\% of the segmentations.
Moreover, qualitative analysis in Figure~\ref{fig:clinical_utility} shows that HD95 values below 7-8 pixels
typically indicate minor inconsistencies in the segmentation maps. Such inconsistencies may also arise from
the variability across inter-expert and intra-expert annotations. On the other hand, HD95 values more than 11
typically indicate significant deviations from the ground truth.
Thus, we consider a segmentation with a HD95 values $\leq$ 8 as having clinical utility.

\subsection{Baseline Methods}
We compare our method against a comprehensive suite of baselines (denoted $\Phi(\cdot)$ earlier).
%
%
As representative of standard encoder-decoder architectures (early DNN methods for medical image
segmentation), we include UNet \citep{UNet}, AttnUNet \citep{AttnUNet}, ResUNet++ \citep{ResUNetPP}, and
DeepLabV3+ \citep{DeeplabV3P_Chen}.
To represent hybrid loss functions, we use BASNet \citep{BASNet_Qin} that is designed for boundary-aware
segmentation.
To represent transformer architectures, we utilize DSTransUNet \citep{DSTransUNet}, a pre-trained model that
has been shown to surpass TransUNet \citep{chen2024transunet}.
Our generative baselines include the adversarial-based SegAN \citep{SegAN_Sue} and the diffusion-based
MedSegDiff \citep{MedSegDiff_Wu}.
We also use some very recent strategies, i.e., VMUNet \citep{ruan2024vmunet} representing state-space models,
and MedSAM \citep{SAM_Medical} representing foundation models. We evaluate MedSAM in a zero-shot setting using
prompts (specific strategy described later), as it is a foundation model trained on over one million medical
image-mask pairs.

\begin{table*}[!t]
\caption {
{\bf Results--Quantitative: Myocardium Segmentation.}
All models were trained on CAP (ID), and evaluated on ACDC and ACMRI (both OOD).
For each method-dataset combination, we report mean (top row) and standard deviation (bottom row, in
\textcolor{gray}{gray}) for DSC($\uparrow$), HD95($\downarrow$), and ASD($\downarrow$).
Augmenting each baseline with our VarDeepPCA consistently improves performance.
{\bf Bold-font} values in the columns indicate a statistically significant improvement of
the Baseline+VarDeepPCA method over the underlying Baseline method, using a one-tailed paired-sample
t-test (p $<$ 0.05).
}
\centering
\resizebox{0.96\textwidth}{!}{%
\begin{minipage}{\textwidth}
\centering
\footnotesize
\renewcommand{\arraystretch}{0.95}
\setlength{\tabcolsep}{2.5pt}
\begin{tabular}{@{}l| *{6}{c} @{\hspace{8pt}}| *{6}{c} @{\hspace{8pt}}| *{6}{c}@{}}		
\toprule
\multirow{3}{*}{} & 
\multicolumn{6}{c}{\textbf{CAP (ID)}} & 
\multicolumn{6}{c}{\textbf{ACDC (OOD)}} & 
\multicolumn{6}{c}{\textbf{ACMRI (OOD)}} \\

\cmidrule(lr){2-7} \cmidrule(lr){8-13} \cmidrule(lr){14-19} 
& \multicolumn{3}{c}{Baseline} & \multicolumn{3}{c}{Baseline +} &
\multicolumn{3}{c}{Baseline} & \multicolumn{3}{c}{Baseline +} &
\multicolumn{3}{c}{Baseline} & \multicolumn{3}{c}{Baseline +} \\

& \multicolumn{3}{c}{} & \multicolumn{3}{c}{\bf VarDeepPCA} &
\multicolumn{3}{c}{} & \multicolumn{3}{c}{\bf VarDeepPCA} &
\multicolumn{3}{c}{} & \multicolumn{3}{c}{\bf VarDeepPCA} \\
\cmidrule(lr){2-4} \cmidrule(lr){5-7} \cmidrule(lr){8-10} \cmidrule(lr){11-13} \cmidrule(lr){14-16} \cmidrule(lr){17-19}
& DSC & HD95 & ASD & DSC & HD95 & ASD & 
DSC  & HD95 & ASD & DSC & HD95 & ASD & 
DSC & HD95 & ASD & DSC & HD95 & ASD \\
\midrule
UNet & 
89.6 & 6.2 & 2.3 & \textbf{90.0} & \textbf{3.8} & \textbf{1.5} & 
73.5 & 28.6 & 7.7 & \textbf{76.8} & \textbf{7.2} & \textbf{2.9} & 
75.2 & 26.0 & 7.9 & \textbf{80.4} & \textbf{8.1} & \textbf{3.6} \\
& \textcolor{gray}{3.7} & \textcolor{gray}{9.6} & \textcolor{gray}{1.8} & \textbf{\textcolor{gray}{3.4}} & \textbf{\textcolor{gray}{1.3}} & \textbf{\textcolor{gray}{0.5}} &
\textcolor{gray}{11.6} & \textcolor{gray}{19.0} & \textcolor{gray}{4.3} & \textbf{\textcolor{gray}{9.9}} & \textbf{\textcolor{gray}{3.7}} & \textbf{\textcolor{gray}{1.2}} &
\textcolor{gray}{11.4} & \textcolor{gray}{20.1} & \textcolor{gray}{5.3} & \textbf{\textcolor{gray}{8.1}} & \textbf{\textcolor{gray}{3.5}} & \textbf{\textcolor{gray}{1.8}} \\
\addlinespace
AttnUNet & 
90.9 & 3.6 & 1.5 & 90.9 & 3.5 & \textbf{1.4} & 
78.4 & 15.3 & 4.3 & \textbf{79.4} & \textbf{6.1} & \textbf{2.6} & 
75.5 & 17.7 & 5.6 & \textbf{78.6} & \textbf{8.2} & \textbf{3.4} \\
& \textcolor{gray}{3.1} & \textcolor{gray}{1.9} & \textcolor{gray}{0.8} & \textcolor{gray}{3.0} & \textcolor{gray}{1.3} & \textbf{\textcolor{gray}{0.5}} &
\textcolor{gray}{10.9} & \textcolor{gray}{14.5} & \textcolor{gray}{3.0} & \textbf{\textcolor{gray}{9.1}} & \textbf{\textcolor{gray}{3.0}} & \textbf{\textcolor{gray}{1.1}} &
\textcolor{gray}{10.0} & \textcolor{gray}{14.1} & \textcolor{gray}{3.3} & \textbf{\textcolor{gray}{8.5}} & \textbf{\textcolor{gray}{3.3}} & \textbf{\textcolor{gray}{1.5}} \\
\addlinespace
ResUNet++ & 
89.1 & 4.2 & 1.6 & \textbf{89.7} & \textbf{3.8} & \textbf{1.5} & 
77.1 & 9.3 & 2.9 & 77.4 & \textbf{7.3} & \textbf{2.8} & 
78.2 & 11.8 & 4.1 & \textbf{79.6} & \textbf{8.5} & \textbf{3.5} \\
& \textcolor{gray}{3.6} & \textcolor{gray}{2.5} & \textcolor{gray}{0.7} & \textbf{\textcolor{gray}{3.5}} & \textbf{\textcolor{gray}{1.3}} & \textbf{\textcolor{gray}{0.5}} &
\textcolor{gray}{8.5} & \textcolor{gray}{6.5} & \textcolor{gray}{1.3} & \textcolor{gray}{8.8} & \textbf{\textcolor{gray}{3.5}} & \textbf{\textcolor{gray}{1.1}} &
\textcolor{gray}{7.8} & \textcolor{gray}{9.4} & \textcolor{gray}{2.2} & \textbf{\textcolor{gray}{7.4}} & \textbf{\textcolor{gray}{3.1}} & \textbf{\textcolor{gray}{1.3}} \\
\addlinespace
DeepLabV3+ & 
88.4 & 4.4 & 1.8 & \textbf{88.8} & \textbf{4.1} & \textbf{1.7} & 
70.1 & 13.6 & 5.1 & 70.5 & \textbf{9.8} & \textbf{4.1} & 
69.9 & 11.8 & 4.5 & \textbf{71.3} & \textbf{9.3} & \textbf{3.9} \\
& \textcolor{gray}{3.9} & \textcolor{gray}{1.9} & \textcolor{gray}{0.6} & \textbf{\textcolor{gray}{3.9}} & \textbf{\textcolor{gray}{1.2}} & \textbf{\textcolor{gray}{0.5}} &
\textcolor{gray}{12.9} & \textcolor{gray}{9.0} & \textcolor{gray}{2.6} & \textcolor{gray}{13.6} & \textbf{\textcolor{gray}{4.4}} & \textbf{\textcolor{gray}{1.7}} &
\textcolor{gray}{10.5} & \textcolor{gray}{7.4} & \textcolor{gray}{2.0} & \textbf{\textcolor{gray}{10.4}} & \textbf{\textcolor{gray}{2.8}} & \textbf{\textcolor{gray}{1.1}} \\
\addlinespace
BASNet & 
91.3 & 3.2 & 1.4 & 91.4 & 3.1 & 1.3 & 
80.2 & 10.0 & 3.5 & 80.2 & \textbf{5.6} & \textbf{2.5} & 
81.1 & 9.1 & 3.4 & 81.2 & \textbf{7.2} & \textbf{3.1} \\
& \textcolor{gray}{2.9} & \textcolor{gray}{1.2} & \textcolor{gray}{0.6} & \textcolor{gray}{3.0} & \textcolor{gray}{1.2} & \textcolor{gray}{0.4} &
\textcolor{gray}{8.5} & \textcolor{gray}{14.0} & \textcolor{gray}{2.9} & \textcolor{gray}{8.1} & \textbf{\textcolor{gray}{1.8}} & \textbf{\textcolor{gray}{0.9}} &
\textcolor{gray}{7.7} & \textcolor{gray}{9.4} & \textcolor{gray}{2.1} & \textcolor{gray}{7.7} & \textbf{\textcolor{gray}{2.8}} & \textbf{\textcolor{gray}{1.3}} \\
\addlinespace
SegAN & 
91.6 & 3.3 & 1.3 & 91.7 & 3.2 & 1.3 & 
72.3 & 11.7 & 4.3 & \textbf{72.7} & \textbf{8.3} & \textbf{3.6} & 
71.0 & 10.9 & 4.0 & \textbf{72.3} & \textbf{8.5} & \textbf{3.4} \\
& \textcolor{gray}{3.5} & \textcolor{gray}{1.4} & \textcolor{gray}{0.5} & \textcolor{gray}{3.5} & \textcolor{gray}{1.3} & \textcolor{gray}{0.5} &
\textcolor{gray}{12.7} & \textcolor{gray}{8.8} & \textcolor{gray}{2.4} & \textbf{\textcolor{gray}{12.8}} & \textbf{\textcolor{gray}{3.2}} & \textbf{\textcolor{gray}{1.3}} &
\textcolor{gray}{13.3} & \textcolor{gray}{8.3} & \textcolor{gray}{1.9} & \textbf{\textcolor{gray}{12.7}} & \textbf{\textcolor{gray}{3.1}} & \textbf{\textcolor{gray}{1.2}} \\
\addlinespace
MedSegDiff & 
87.1 & 4.3 & 1.9 & \textbf{88.1} & \textbf{4.2} & \textbf{1.9} & 
69.7 & 12.2 & 4.7 & \textbf{71.0} & \textbf{8.9} & \textbf{4.1} & 
71.7 & 11.5 & 4.7 & \textbf{73.0} & \textbf{9.5} & \textbf{4.3} \\
& \textcolor{gray}{7.3} & \textcolor{gray}{1.6} & \textcolor{gray}{0.7} & \textbf{\textcolor{gray}{4.4}} & \textbf{\textcolor{gray}{1.3}} & \textbf{\textcolor{gray}{0.6}} &
\textcolor{gray}{10.0} & \textcolor{gray}{8.7} & \textcolor{gray}{2.3} & \textbf{\textcolor{gray}{10.1}} & \textbf{\textcolor{gray}{2.8}} & \textbf{\textcolor{gray}{1.5}} &
\textcolor{gray}{12.1} & \textcolor{gray}{8.4} & \textcolor{gray}{2.3} & \textbf{\textcolor{gray}{11.3}} & \textbf{\textcolor{gray}{3.0}} & \textbf{\textcolor{gray}{1.6}} \\
\addlinespace
DSTransUNet & 
91.5 & 3.5 & 1.3 & \textbf{91.8} & \textbf{3.0} & \textbf{1.2} & 
77.6 & 11.6 & 3.8 & \textbf{79.2} & \textbf{6.7} & \textbf{2.7} & 
81.0 & 8.3 & 3.2 & \textbf{81.8} & \textbf{7.4} & 3.2 \\
& \textcolor{gray}{3.0} & \textcolor{gray}{3.8} & \textcolor{gray}{0.9} & \textbf{\textcolor{gray}{2.8}} & \textbf{\textcolor{gray}{1.2}} & \textbf{\textcolor{gray}{0.5}} &
\textcolor{gray}{9.8} & \textcolor{gray}{10.2} & \textcolor{gray}{2.5} & \textbf{\textcolor{gray}{9.0}} & \textbf{\textcolor{gray}{3.5}} & \textbf{\textcolor{gray}{1.2}} &
\textcolor{gray}{8.6} & \textcolor{gray}{6.0} & \textcolor{gray}{1.7} & \textbf{\textcolor{gray}{7.9}} & \textbf{\textcolor{gray}{3.7}} & \textcolor{gray}{1.7} \\
\addlinespace
VMUNet & 
90.4 & 3.5 & 1.4 & 90.6 & 3.4 & 1.4 & 
77.9 & 9.2 & 3.2 & 78.0 & \textbf{6.9} & \textbf{2.8} & 
78.2 & 8.3 & 3.1 & 78.3 & \textbf{7.9} & 3.1 \\
& \textcolor{gray}{2.9} & \textcolor{gray}{1.2} & \textcolor{gray}{0.5} & \textcolor{gray}{2.9} & \textcolor{gray}{1.2} & \textcolor{gray}{0.5} &
\textcolor{gray}{10.4} & \textcolor{gray}{9.2} & \textcolor{gray}{2.4} & \textcolor{gray}{10.2} & \textbf{\textcolor{gray}{3.3}} & \textbf{\textcolor{gray}{1.1}} &
\textcolor{gray}{10.4} & \textcolor{gray}{4.8} & \textcolor{gray}{1.4} & \textcolor{gray}{10.5} & \textbf{\textcolor{gray}{3.8}} & \textcolor{gray}{1.3} \\
\addlinespace
MedSAM & 
62.9 & 10.4 & 4.1 & \textbf{68.6} & \textbf{6.5} & \textbf{3.8} & 
60.0 & 9.5 & 3.1 & \textbf{63.3} & \textbf{6.3} & \textbf{2.3} & 
71.3 & 8.8 & 3.3 & \textbf{74.1} & \textbf{6.2} & \textbf{2.1} \\
& \textcolor{gray}{13.6} & \textcolor{gray}{3.7} & \textcolor{gray}{1.4} & \textbf{\textcolor{gray}{13.1}} & \textbf{\textcolor{gray}{3.1}} & \textbf{\textcolor{gray}{1.5}} &
\textcolor{gray}{21.5} & \textcolor{gray}{4.9} & \textcolor{gray}{1.2} & \textbf{\textcolor{gray}{20.6}} & \textbf{\textcolor{gray}{4.1}} & \textbf{\textcolor{gray}{0.2}} &
\textcolor{gray}{15.8} & \textcolor{gray}{3.4} & \textcolor{gray}{0.9} & \textbf{\textcolor{gray}{14.0}} & \textbf{\textcolor{gray}{3.1}} & \textbf{\textcolor{gray}{0.3}} \\
\addlinespace
PHISeg & 
88.9 & 3.9 & 1.6 & 88.9 & 3.9 & 1.5 & 
74.6 & 7.3 & 2.9 & 74.8 & \textbf{6.7} & 2.9 & 
72.9 & 8.7 & 3.2 & 73.0 & \textbf{8.3} & 3.2 \\
& \textcolor{gray}{4.2} & \textcolor{gray}{1.6} & \textcolor{gray}{0.6} & \textcolor{gray}{4.2} & \textcolor{gray}{1.4} & \textcolor{gray}{0.6} &
\textcolor{gray}{14.2} & \textcolor{gray}{4.2} & \textcolor{gray}{1.2} & \textcolor{gray}{14.3} & \textbf{\textcolor{gray}{3.0}} & \textcolor{gray}{1.1} &
\textcolor{gray}{11.1} & \textcolor{gray}{3.6} & \textcolor{gray}{1.0} & \textcolor{gray}{10.9} & \textbf{\textcolor{gray}{3.0}} & \textcolor{gray}{1.0} \\
\addlinespace
ProbUNet & 
89.3 & 4.4 & 1.8 & \textbf{90.1} & \textbf{3.6} & \textbf{1.5} & 
74.3 & 21.5 & 6.1 & \textbf{78.1} & \textbf{6.4} & \textbf{2.6} & 
73.4 & 21.3 & 6.4 & \textbf{77.9} & \textbf{7.6} & \textbf{3.2} \\
& \textcolor{gray}{3.9} & \textcolor{gray}{3.2} & \textcolor{gray}{1.0} & \textbf{\textcolor{gray}{3.6}} & \textbf{\textcolor{gray}{1.3}} & \textbf{\textcolor{gray}{0.5}} &
\textcolor{gray}{10.8} & \textcolor{gray}{17.0} & \textcolor{gray}{3.9} & \textbf{\textcolor{gray}{9.5}} & \textbf{\textcolor{gray}{3.0}} & \textbf{\textcolor{gray}{0.9}} &
\textcolor{gray}{10.4} & \textcolor{gray}{15.0} & \textcolor{gray}{3.7} & \textbf{\textcolor{gray}{8.8}} & \textbf{\textcolor{gray}{2.8}} & \textbf{\textcolor{gray}{1.5}} \\
\addlinespace
HierProbUNet & 
86.7 & 6.2 & 2.3 & \textbf{89.2} & \textbf{3.9} & \textbf{1.6} & 
60.4 & 39.1 & 12.0 & \textbf{70.0} & \textbf{9.4} & \textbf{3.5} & 
71.0 & 25.7 & 7.8 & \textbf{76.6} & \textbf{8.8} & \textbf{3.7} \\
& \textcolor{gray}{4.9} & \textcolor{gray}{5.4} & \textcolor{gray}{1.4} & \textbf{\textcolor{gray}{4.0}} & \textbf{\textcolor{gray}{1.4}} & \textbf{\textcolor{gray}{0.5}} &
\textcolor{gray}{9.1} & \textcolor{gray}{15.4} & \textcolor{gray}{4.5} & \textbf{\textcolor{gray}{8.2}} & \textbf{\textcolor{gray}{2.9}} & \textbf{\textcolor{gray}{1.2}} &
\textcolor{gray}{9.6} & \textcolor{gray}{14.1} & \textcolor{gray}{3.4} & \textbf{\textcolor{gray}{8.6}} & \textbf{\textcolor{gray}{3.0}} & \textbf{\textcolor{gray}{1.5}} \\
\addlinespace
SegCNN+DAE & 
90.1 & 3.8 & 1.7 & 90.2 & 3.7 & 1.6 & 
78.2 & 8.3 & 3.2 & \textbf{79.7} & \textbf{5.9} & \textbf{2.2} & 
79.3 & 10.3 & 3.7 & \textbf{80.4} & \textbf{7.8} & \textbf{2.7} \\
+TTA & \textcolor{gray}{3.2} & \textcolor{gray}{1.4} & \textcolor{gray}{0.5} & \textcolor{gray}{3.2} & \textcolor{gray}{1.4} & \textcolor{gray}{0.4} &
\textcolor{gray}{10.1 } & \textcolor{gray}{7.5}  & \textcolor{gray}{2.2} & \textbf{\textcolor{gray}{9.7}} & \textbf{\textcolor{gray}{2.8}} & \textbf{\textcolor{gray}{1.3}} &
\textcolor{gray}{9.8} & \textcolor{gray}{9.4} & \textcolor{gray}{2.8} & \textbf{\textcolor{gray}{9.1}} & \textbf{\textcolor{gray}{3.5}} & \textbf{\textcolor{gray}{2.1}} \\
\addlinespace
SegCNN+DAE & 
90.6 & 3.6 & 1.4 & 90.6 & 3.6 & 1.4 & 
79.7 & 7.3 & 2.8 & \textbf{80.6} & \textbf{5.8} & \textbf{2.1} & 
80.4 & 9.3 & 3.4 & \textbf{81.9} & \textbf{7.5} & \textbf{2.6} \\
+TTA+Atlas & \textcolor{gray}{3.2} & \textcolor{gray}{1.3} & \textcolor{gray}{0.4} & \textcolor{gray}{3.2} & \textcolor{gray}{1.3} & \textcolor{gray}{0.4} &
\textcolor{gray}{9.8} & \textcolor{gray}{7.4} & \textcolor{gray}{1.9} & \textbf{\textcolor{gray}{9.5}} & \textbf{\textcolor{gray}{2.6}} & \textbf{\textcolor{gray}{1.1}} &
\textcolor{gray}{9.0} & \textcolor{gray}{8.9} & \textcolor{gray}{2.5} & \textbf{\textcolor{gray}{8.7}} & \textbf{\textcolor{gray}{3.3}} & \textbf{\textcolor{gray}{1.9}} \\
\midrule
Mean of Baselines & 
87.7 & 4.6 & 1.8 & \textbf{88.6} & \textbf{3.8} & \textbf{1.6} & 
74.5 & 13.6 & 4.4 & \textbf{75.8} & \textbf{7.2} & \textbf{3.0} & 
75.5 & 13.2 & 4.5 & \textbf{77.5} & \textbf{8.0} & \textbf{3.3} \\
& \textcolor{gray}{8.9} & \textcolor{gray}{4.0} & \textcolor{gray}{1.2} & \textbf{\textcolor{gray}{7.6}} & \textbf{\textcolor{gray}{1.7}} & \textbf{\textcolor{gray}{0.9}} &
\textcolor{gray}{12.2} & \textcolor{gray}{13.6} & \textcolor{gray}{3.3} & \textbf{\textcolor{gray}{11.4}} & \textbf{\textcolor{gray}{3.4}} & \textbf{\textcolor{gray}{1.3}} &
\textcolor{gray}{11.4} & \textcolor{gray}{12.2} & \textcolor{gray}{3.1} & \textbf{\textcolor{gray}{10.4}} & \textbf{\textcolor{gray}{3.3}} & \textbf{\textcolor{gray}{1.5}} \\
\bottomrule
\end{tabular}
\end{minipage}%
}
\label{tab:myo}
\end{table*}

\begin{figure}[!t]
\threeAcrossLabelsWidth{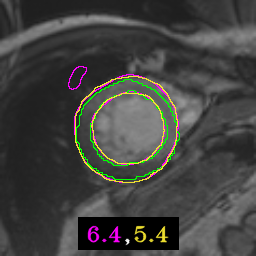}
{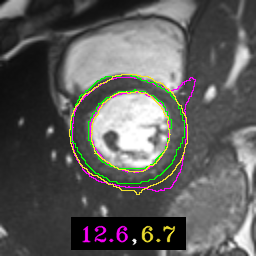}
{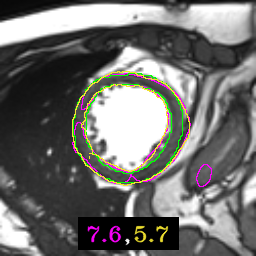}
{\scs{\bf(a)}~SegCNN+DAE+TTA+Atlas}{\scs{\bf(b)}~VMUNet}{\scs{\bf(c)}~ResUNet++}{0.48}
\threeAcrossLabelsWidth{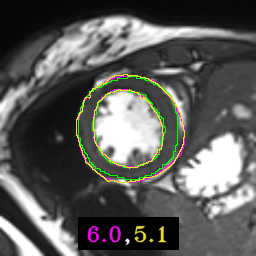}
{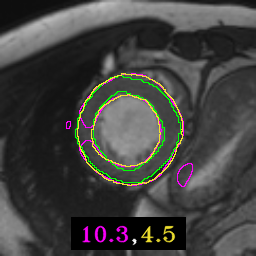}
{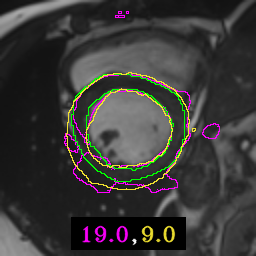}
{\scs{\bf(d)}~PHISeg}{\scs{\bf(e)}~ProbUNet}{\scs{\bf(f)}~HierProbUNet}{0.48}
\threeAcrossLabelsWidth{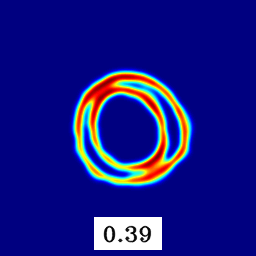}
{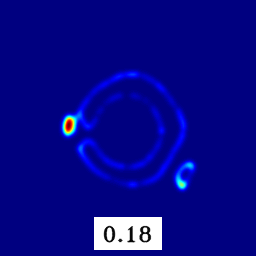}
{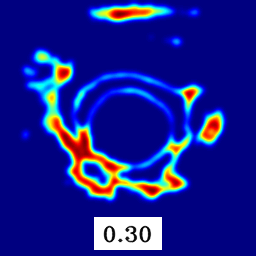}
{\scs{\bf(g)}~Unc. for (d)~PHISeg}{\scs{\bf(h)}~Unc. for (e)~ProbUNet}{\scs{\bf(i)}~Unc. for (f)~HierProbUNet}{0.48}
\threeAcrossLabelsWidth{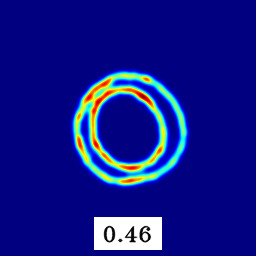}
{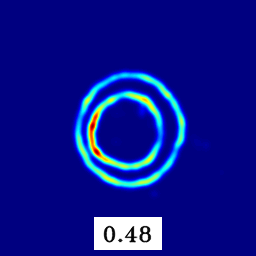}
{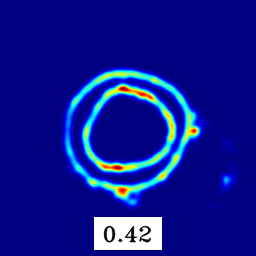}
{\scs{\bf(j)}~Unc. for (d)+VarDeepPCA}{\scs{\bf(k)}~Unc. for (e)+VarDeepPCA}{\scs{\bf(l)}~Unc. for (f)+VarDeepPCA}{0.48}
\caption
{
{\bf Results--Qualitative: Myocardium Segmentation Restoration on ACDC (OOD) data.}
{\bf (a)--(c)}~Results on images for the best non-variational baselines.
{\bf (d)--(f)}~Results on images for the variational baselines.
{\bf (g)--(i)}~Uncertainty maps produced using variational baselines.
{\bf (j)--(l)}~Uncertainty maps produced using VarDeepPCA when plugged into the associated baselines
(d)--(f).
Color scheme in (a)--(f):
\textcolor{myo_maroon}{Baseline};
\textcolor{myo_yellow}{Baseline+VarDeepPCA (Ours)};
\textcolor{myo_green}{Ground Truth}.
HD95 numbers in (a)--(f) indicate that the examples were representative of the test set, because the HD95
values were close to the mean HD95 reported in Table~\ref{tab:myo}.
NCC numbers in (g)--(l) indicate that the examples were representative of the test set, because the NCC values were close to the mean of the NCC reported in Table~\ref{tab:myo_calibration}.}
\label{fig:myo_acdc_auto}
\end{figure}

\begin{figure}[!t]
\threeAcrossLabelsWidth{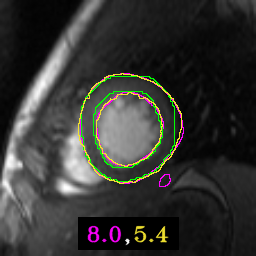}
{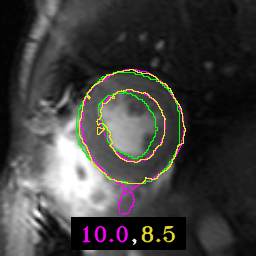}
{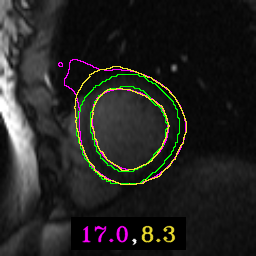}
{\scs{\bf(a)}~DSTransUNet}{\scs{\bf(b)}~VMUNet}{\scs{\bf(c)}~BASNet}{0.48}
\threeAcrossLabelsWidth{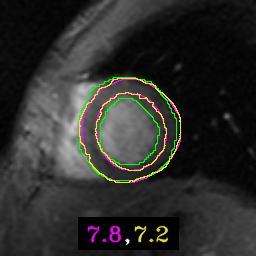}
{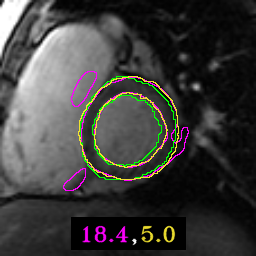}
{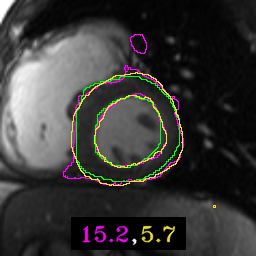}
{\scs{\bf(d)}~PHISeg}{\scs{\bf(e)}~ProbUNet}{\scs{\bf(f)}~HierProbUNet}{0.48}
\threeAcrossLabelsWidth{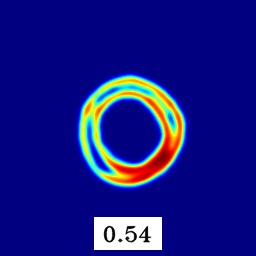}
{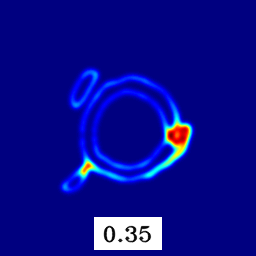}
{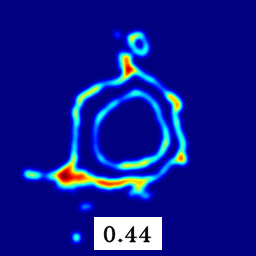}
{\scs{\bf(g)}~Unc. for (d)~PHISeg}{\scs{\bf(h)}~Unc. for (e)~ProbUNet}{\scs{\bf(i)}~Unc. for (f)~HierProbUNet}{0.48}

\threeAcrossLabelsWidth{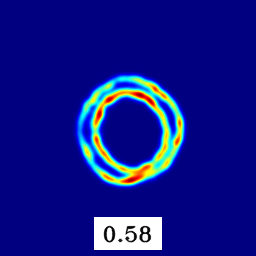}
{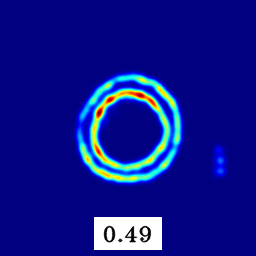}
{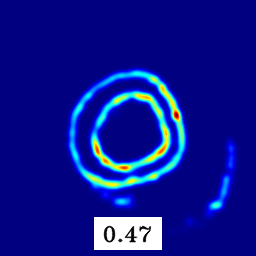}
{\scs{\bf(j)}~Unc. for (d)+VarDeepPCA}{\scs{\bf(k)}~Unc. for (e)+VarDeepPCA}{\scs{\bf(l)}~Unc. for (f)+VarDeepPCA}{0.48}
\caption
{
{\bf Results--Qualitative: Myocardium Segmentation Restoration on ACMRI (OOD) data.}
{\bf (a)--(c)}~Results on images for the best non-variational baselines.
{\bf (d)--(f)}~Results on images for the variational baselines. 
{\bf (g)--(i)}~Uncertainty maps produced using variational baselines.
{\bf (j)--(l)}~Uncertainty maps produced using VarDeepPCA when plugged into the associated baselines
(d)--(f).
Color scheme in (a)--(f):
\textcolor{myo_maroon}{Baseline};
\textcolor{myo_yellow}{Baseline+VarDeepPCA (Ours)};
\textcolor{myo_green}{Ground Truth}.
HD95 numbers in (a)--(f) indicate that the examples were representative of the test set, because the HD95
values were close to the mean HD95 reported in Table~\ref{tab:myo}.
NCC numbers in (g)--(l) indicate that the examples were representative of the test set, because
the NCC values were close to the mean of the NCC reported in Table~\ref{tab:myo_calibration}.} 
\label{fig:myo_acmri_auto}
\end{figure}

Within the class of variational-learning methods (which also produce uncertainty estimates), we use
ProbUNet \citep{probUNet}, HierProbUNet \citep{hierarchicalprobUNet}, and PHISeg \citep{phiseg}. For each
variational method, we generated 20 stochastic samples per input image at inference time; these samples lead
to the mean segmentation map and the per-pixel standard deviation (uncertainty) map. We utilize the mean
prediction generated by each variational method as the input for VarDeepPCA refinement.
Within the class of TTA-based methods, we use the method by \citep{r1_karani} that has (i)~an
image-to-image translator for normalization, which is adapted to the test image at test time, (ii)~a
UNet-based segmenter, and (iii)~a denoising autoencoder (DAE) that removes minor inconsistencies in the
segmentation masks.
\citep{r1_karani} uses jigsaw-based random-patch \citep{r1_karani} replacement to train the DAE using 2D
masks. We use two versions of the method by \citep{r1_karani}, i.e., SegCNN+TTA+DAE, and SegCNN+TTA+DAE+Atlas
where the atlas constrains the predicted segmentation map to be close to typical ID segmentation maps.

{\bf Model Size, Training Time, and Inference Time.}
Table~\ref{tab:model_params_time} gives the model size, training time, and inference time.
All experiments use an NVIDIA RTX A6000 GPU.
The training and inference times remain the same across applications because they use the same number of
training images, same image size, and same number of epochs.
Among the baselines, DeepLabV3+ exhibited the fastest training (5.6 min), likely due to its efficient atrous
convolution modules \citep{DeeplabV3P_Chen}.
ResUNet++ was the fastest during inference.
The model in SegCNN+TTA+DAE had the smallest parameter count.
Our VarDeepPCA plugin framework is lightweight. Its smallest configuration (latent dimension $K = 3$)
comprises 1.02M parameters, which is approximately 50\% of the smallest baseline (SegCNN+TTA+DAE). Its
largest configuration (latent dimension $K = 16$) comprises 2.72M parameters; this contrasts sharply with
the largest baseline (SegAN with 216.44M params).
Because of VarDeepPCA's architectural simplicity, it has very small training times (under 2 minutes for all
latent vectors from $K = 2$ to $K = 16$) and inference times (around $3.56$ ms per image). In this way, the
VarDeepPCA plugin typically leads to (very) small overheads in terms of model size and training/inference
time.

\begin{table*}[!t]
\caption
{
{\bf Results--Quantitative: Neuroretinal Rim Segmentation.}
All models were trained on MAGRABI (ID), and evaluated on G1020 and ORIGA (both OOD).
For each method-dataset combination, we report mean (top row) and standard deviation (bottom row, in
\textcolor{gray}{gray}) for DSC($\uparrow$), HD95($\downarrow$), and ASD($\downarrow$).
Augmenting each baseline with our VarDeepPCA consistently improves performance.
{\bf Bold-font} values in the columns indicate a statistically significant improvement of
the Baseline+VarDeepPCA method over the underlying Baseline method, using a one-tailed paired-sample
t-test (p $<$ 0.05).
}
\centering
\resizebox{0.96\textwidth}{!}{%
\begin{minipage}{\textwidth}
\centering
\footnotesize
\renewcommand{\arraystretch}{0.95}
\setlength{\tabcolsep}{2.5pt}
\begin{tabular}{@{}l| *{6}{c} @{\hspace{8pt}}| *{6}{c} @{\hspace{8pt}}| *{6}{c}@{}}     
\toprule
\multirow{3}{*}{\textbf{Models}} & 
\multicolumn{6}{c}{\textbf{MAGRABI (ID)}} & 
\multicolumn{6}{c}{\textbf{G1020 (OOD)}} & 
\multicolumn{6}{c}{\textbf{ORIGA (OOD)}} \\
\cmidrule(lr){2-7} \cmidrule(lr){8-13} \cmidrule(lr){14-19}
& \multicolumn{3}{c}{Baseline} & \multicolumn{3}{c}{Baseline +} &
\multicolumn{3}{c}{Baseline} & \multicolumn{3}{c}{Baseline +} &
\multicolumn{3}{c}{Baseline} & \multicolumn{3}{c}{Baseline +} \\
& \multicolumn{3}{c}{} & \multicolumn{3}{c}{\bf VarDeepPCA} &
\multicolumn{3}{c}{} & \multicolumn{3}{c}{\bf VarDeepPCA} &
\multicolumn{3}{c}{} & \multicolumn{3}{c}{\bf VarDeepPCA} \\
\cmidrule(lr){2-4} \cmidrule(lr){5-7} \cmidrule(lr){8-10} \cmidrule(lr){11-13} \cmidrule(lr){14-16} \cmidrule(lr){17-19}
& DSC & HD95 & ASD & DSC & HD95 & ASD & 
DSC  & HD95 & ASD & DSC & HD95 & ASD & 
DSC & HD95 & ASD & DSC & HD95 & ASD \\
\midrule
UNet & 
88.1 & 7.7 & 3.3 & \textbf{89.4} & \textbf{6.0} & \textbf{2.8} & 
79.3 & 16.2 & 5.7 & \textbf{82.6} & \textbf{7.3} & \textbf{3.4} & 
71.7 & 21.1 & 7.6 & \textbf{76.3} & \textbf{7.6} & \textbf{4.3} \\
& \textcolor{gray}{4.7} & \textcolor{gray}{6.3} & \textcolor{gray}{1.9} & \textbf{\textcolor{gray}{4.6}} & \textbf{\textcolor{gray}{2.1}} & \textbf{\textcolor{gray}{1.2}} &
\textcolor{gray}{10.8} & \textcolor{gray}{19.3} & \textcolor{gray}{5.9} & \textbf{\textcolor{gray}{8.3}} & \textbf{\textcolor{gray}{2.2}} & \textbf{\textcolor{gray}{1.3}} &
\textcolor{gray}{10.2} & \textcolor{gray}{20.5} & \textcolor{gray}{6.2} & \textbf{\textcolor{gray}{6.4}} & \textbf{\textcolor{gray}{1.9}} & \textbf{\textcolor{gray}{1.1}} \\
\addlinespace
AttnUNet & 
92.0 & 5.5 & 2.0 & \textbf{92.9} & \textbf{4.9} & \textbf{1.8} & 
78.3 & 9.8 & 4.2 & \textbf{82.2} & \textbf{7.7} & \textbf{3.4} & 
68.6 & 11.1 & 5.4 & \textbf{69.3} & \textbf{9.3} & 5.3 \\
& \textcolor{gray}{4.0} & \textcolor{gray}{2.4} & \textcolor{gray}{0.8} & \textbf{\textcolor{gray}{3.7}} & \textbf{\textcolor{gray}{2.0}} & \textbf{\textcolor{gray}{0.7}} &
\textcolor{gray}{9.5} & \textcolor{gray}{7.1} & \textcolor{gray}{1.9} & \textbf{\textcolor{gray}{8.3}} & \textbf{\textcolor{gray}{2.0}} & \textbf{\textcolor{gray}{1.1}} &
\textcolor{gray}{7.3} & \textcolor{gray}{4.2} & \textcolor{gray}{1.3} & \textbf{\textcolor{gray}{7.6}} & \textbf{\textcolor{gray}{1.6}} & \textcolor{gray}{1.1} \\
\addlinespace
ResUNet++ & 
76.9 & 23.7 & 2.2 & 77.2 & \textbf{7.1} & \textbf{1.9} & 
73.9 & 32.2 & 8.1 & \textbf{76.6} & \textbf{10.8} & \textbf{5.3} & 
47.8 & 26.3 & 8.8 & \textbf{55.7} & \textbf{10.4} & \textbf{5.7} \\
& \textcolor{gray}{21.9} & \textcolor{gray}{12.2} & \textcolor{gray}{1.8} & \textcolor{gray}{21.5} &\textbf{ \textcolor{gray}{1.1}} & \textbf{\textcolor{gray}{1.5}} &
\textcolor{gray}{10.6} & \textcolor{gray}{16.9} & \textcolor{gray}{7.3} & \textbf{\textcolor{gray}{7.8}} & \textbf{\textcolor{gray}{0.9}} & \textbf{\textcolor{gray}{1.3}} &
\textcolor{gray}{25.4} & \textcolor{gray}{15.9} & \textcolor{gray}{6.3} & \textbf{\textcolor{gray}{23.6}} & \textbf{\textcolor{gray}{1.5}} & \textbf{\textcolor{gray}{2.1}} \\
\addlinespace
DeepLabV3+ & 
92.2 & 5.5 & 2.0 & 92.2 & \textbf{5.3} & 2.0 & 
74.7 & 13.0 & 6.3 & \textbf{76.3} & \textbf{9.2} & \textbf{5.5} & 
62.0 & 11.4 & 7.6 & \textbf{63.3} & \textbf{8.6} & \textbf{6.5} \\
& \textcolor{gray}{3.7} & \textcolor{gray}{2.5} & \textcolor{gray}{0.9} & \textcolor{gray}{3.6} & \textbf{\textcolor{gray}{2.1}} & \textcolor{gray}{0.8} &
\textcolor{gray}{7.4} & \textcolor{gray}{8.4} & \textcolor{gray}{2.4} & \textbf{\textcolor{gray}{6.2}} & \textbf{\textcolor{gray}{1.6}} & \textbf{\textcolor{gray}{1.4}} &
\textcolor{gray}{7.4} & \textcolor{gray}{4.0} & \textcolor{gray}{1.3} & \textbf{\textcolor{gray}{7.0}} & \textbf{\textcolor{gray}{1.0}} & \textbf{\textcolor{gray}{1.0}} \\
\addlinespace
BASNet & 
93.7 & 4.4 & 1.6 & 93.8 & \textbf{4.3} & \textbf{1.6} & 
79.5 & 11.7 & 5.2 & \textbf{80.5} & \textbf{8.0} & \textbf{4.4} & 
65.1 & 13.0 & 7.1 & \textbf{67.6} & \textbf{10.1} & \textbf{6.5} \\
& \textcolor{gray}{3.0} & \textcolor{gray}{1.9} & \textcolor{gray}{0.7} & \textcolor{gray}{2.9} & \textbf{\textcolor{gray}{1.9}} & \textbf{\textcolor{gray}{0.6}} &
\textcolor{gray}{7.1} & \textcolor{gray}{10.4} & \textcolor{gray}{2.8} & \textbf{\textcolor{gray}{6.4}} & \textbf{\textcolor{gray}{2.0}} & \textbf{\textcolor{gray}{1.4}} &
\textcolor{gray}{6.8} & \textcolor{gray}{8.9} & \textcolor{gray}{2.2} & \textbf{\textcolor{gray}{6.7}} & \textbf{\textcolor{gray}{1.2}} & \textbf{\textcolor{gray}{0.9}} \\
\addlinespace
SegAN & 
77.2 & 23.0 & 1.7 & 77.7 & \textbf{7.2} & 1.7 & 
79.1 & 26.3 & 4.8 & \textbf{83.0} & \textbf{10.7} & 4.6 & 
46.5 & 15.3 & 5.1 & \textbf{53.9} & \textbf{10.5} & 5.9 \\
& \textcolor{gray}{23.0} & \textcolor{gray}{10.9} & \textcolor{gray}{0.7} & \textcolor{gray}{22.9} & \textbf{\textcolor{gray}{0.4}} & \textcolor{gray}{1.0} &
\textcolor{gray}{9.0} & \textcolor{gray}{3.2} & \textcolor{gray}{1.6} & \textbf{\textcolor{gray}{8.6}} & \textbf{\textcolor{gray}{1.0}} & \textcolor{gray}{1.5} &
\textcolor{gray}{19.2} & \textcolor{gray}{7.8} & \textcolor{gray}{1.8} & \textbf{\textcolor{gray}{19.0}} & \textbf{\textcolor{gray}{1.2}} & \textcolor{gray}{1.3} \\
\addlinespace
MedSegDiff & 
92.5 & 5.2 & 1.9 & \textbf{92.6} & \textbf{5.0} & \textbf{1.8} & 
77.8 & 9.8 & 5.3 & 78.0 & \textbf{8.5} & \textbf{5.0} & 
62.9 & 10.8 & 7.0 & \textbf{63.5} & \textbf{9.1} & 7.0 \\
& \textcolor{gray}{3.2} & \textcolor{gray}{2.0} & \textcolor{gray}{0.7} & \textbf{\textcolor{gray}{3.2}} & \textbf{\textcolor{gray}{2.0}} & \textbf{\textcolor{gray}{0.7}} &
\textcolor{gray}{7.2} & \textcolor{gray}{8.1} & \textcolor{gray}{2.5} & \textcolor{gray}{7.1} & \textbf{\textcolor{gray}{1.9}} & \textbf{\textcolor{gray}{1.6}} &
\textcolor{gray}{7.2} & \textcolor{gray}{2.5} & \textcolor{gray}{1.2} & \textbf{\textcolor{gray}{7.2}} & \textbf{\textcolor{gray}{1.2}} & \textcolor{gray}{1.0} \\
\addlinespace
DSTransUNet & 
92.4 & 5.5 & 2.0 & \textbf{93.1} & \textbf{4.7} & \textbf{1.7} & 
78.5 & 14.4 & 5.8 & \textbf{80.7} & \textbf{7.9} & \textbf{4.1} & 
62.3 & 21.8 & 8.6 & \textbf{68.2} & \textbf{9.6} & \textbf{6.0} \\
& \textcolor{gray}{3.3} & \textcolor{gray}{4.3} & \textcolor{gray}{1.0} & \textbf{\textcolor{gray}{3.2}} & \textbf{\textcolor{gray}{1.8}} & \textbf{\textcolor{gray}{0.7}} &
\textcolor{gray}{8.6} & \textcolor{gray}{16.5} & \textcolor{gray}{5.3} & \textbf{\textcolor{gray}{7.0}} & \textbf{\textcolor{gray}{2.1}} & \textbf{\textcolor{gray}{1.4}} &
\textcolor{gray}{8.1} & \textcolor{gray}{15.4} & \textcolor{gray}{3.6} & \textbf{\textcolor{gray}{7.1}} & \textbf{\textcolor{gray}{1.3}} & \textbf{\textcolor{gray}{1.0}} \\
\addlinespace
VMUNet & 
93.0 & 4.8 & 1.8 & \textbf{93.2} & \textbf{4.6} & \textbf{1.7} & 
79.1 & 10.0 & 4.8 & \textbf{80.1} & \textbf{8.3} & \textbf{4.4} & 
61.8 & 12.7 & 7.2 & 62.0 & \textbf{10.4} & \textbf{6.8} \\
& \textcolor{gray}{3.2} & \textcolor{gray}{2.0} & \textcolor{gray}{0.7} & \textbf{\textcolor{gray}{3.3}} & \textbf{\textcolor{gray}{2.0}} & \textbf{\textcolor{gray}{0.7}} &
\textcolor{gray}{7.1} & \textcolor{gray}{8.0} & \textcolor{gray}{2.1} & \textbf{\textcolor{gray}{6.8}} & \textbf{\textcolor{gray}{1.9}} & \textbf{\textcolor{gray}{1.4}} &
\textcolor{gray}{7.1} & \textcolor{gray}{7.3} & \textcolor{gray}{1.7} & \textcolor{gray}{7.2} & \textbf{\textcolor{gray}{1.2}} & \textbf{\textcolor{gray}{1.0}} \\
\addlinespace
MedSAM & 
78.2 & 11.8 & 5.0 & \textbf{84.6} & \textbf{7.8} & \textbf{3.3} & 
80.3 & 9.2 & 3.4 & \textbf{85.1} & \textbf{6.8} & \textbf{2.9} & 
77.4 & 6.5 & 2.6 & \textbf{81.1} & \textbf{5.5} & \textbf{2.4} \\
& \textcolor{gray}{7.6} & \textcolor{gray}{3.2} & \textcolor{gray}{1.5} & \textbf{\textcolor{gray}{6.0}} & \textbf{\textcolor{gray}{2.8}} & \textbf{\textcolor{gray}{1.7}} &
\textcolor{gray}{8.5} & \textcolor{gray}{4.0} & \textcolor{gray}{1.0} & \textbf{\textcolor{gray}{6.9}} & \textbf{\textcolor{gray}{2.4}} & \textbf{\textcolor{gray}{1.0}} &
\textcolor{gray}{11.6} & \textcolor{gray}{2.1} & \textcolor{gray}{0.7} & \textbf{\textcolor{gray}{10.0}} & \textbf{\textcolor{gray}{1.9}} & \textbf{\textcolor{gray}{0.7}} \\
\addlinespace
PHISeg & 
92.3 & 5.0 & 1.9 & \textbf{93.1} & \textbf{4.8} & \textbf{1.7} & 
74.6 & 10.7 & 4.2 & \textbf{78.1} & \textbf{8.2} & \textbf{4.2} & 
63.6 & 10.4 & 6.3 & 63.5 & \textbf{9.8} & 6.4 \\
& \textcolor{gray}{4.4} & \textcolor{gray}{2.0} & \textcolor{gray}{0.7} & \textbf{\textcolor{gray}{3.2}} & \textbf{\textcolor{gray}{1.9}} & \textbf{\textcolor{gray}{0.6}} &
\textcolor{gray}{16.0} & \textcolor{gray}{8.0} & \textcolor{gray}{1.3} & \textbf{\textcolor{gray}{9.7}} & \textbf{\textcolor{gray}{1.9}} & \textbf{\textcolor{gray}{1.3}} &
\textcolor{gray}{7.8} & \textcolor{gray}{1.4} & \textcolor{gray}{1.1} & \textcolor{gray}{7.7} & \textbf{\textcolor{gray}{1.3}} & \textcolor{gray}{1.1} \\
\addlinespace
ProbUNet & 
82.9 & 10.1 & 4.3 & \textbf{85.4} & \textbf{6.8} & \textbf{3.7} & 
66.0 & 21.8 & 7.8 & \textbf{74.0} & \textbf{9.3} & \textbf{4.3} & 
62.5 & 30.7 & 10.9 & \textbf{70.8} & \textbf{9.7} & \textbf{3.7} \\
& \textcolor{gray}{8.3} & \textcolor{gray}{5.0} & \textcolor{gray}{1.7} & \textbf{\textcolor{gray}{7.4}} & \textbf{\textcolor{gray}{2.4}} & \textbf{\textcolor{gray}{1.6}} &
\textcolor{gray}{13.3} & \textcolor{gray}{12.7} & \textcolor{gray}{4.2} & \textbf{\textcolor{gray}{11.5}} & \textbf{\textcolor{gray}{1.8}} & \textbf{\textcolor{gray}{1.2}} &
\textcolor{gray}{14.0} & \textcolor{gray}{15.9} & \textcolor{gray}{6.1} & \textbf{\textcolor{gray}{13.6}} & \textbf{\textcolor{gray}{1.6}} & \textbf{\textcolor{gray}{0.9}} \\
\addlinespace
HierProbUNet & 
84.6 & 10.9 & 4.4 & \textbf{88.7} & \textbf{6.3} & \textbf{3.0} & 
72.1 & 28.0 & 10.3 & \textbf{80.8} & \textbf{8.6} & \textbf{4.0} & 
67.5 & 14.9 & 6.4 & \textbf{72.0} & \textbf{9.1} & \textbf{4.4} \\
& \textcolor{gray}{6.1} & \textcolor{gray}{8.1} & \textcolor{gray}{3.0} & \textbf{\textcolor{gray}{4.7}} & \textbf{\textcolor{gray}{2.3}} & \textbf{\textcolor{gray}{1.1}} &
\textcolor{gray}{11.1} & \textcolor{gray}{26.8} & \textcolor{gray}{9.8} & \textbf{\textcolor{gray}{8.5}} & \textbf{\textcolor{gray}{2.1}} & \textbf{\textcolor{gray}{1.4}} &
\textcolor{gray}{8.7} & \textcolor{gray}{12.6} & \textcolor{gray}{5.2} & \textbf{\textcolor{gray}{8.1}} & \textbf{\textcolor{gray}{1.8}} & \textbf{\textcolor{gray}{1.2}} \\
\addlinespace
SegCNN+DAE & 
89.3 & 6.8 & 2.9 & \textbf{90.7} & \textbf{6.3} & \textbf{2.7} & 
79.6 & 9.7 & 4.9 & \textbf{80.8} & \textbf{8.1} & \textbf{4.3} & 
67.9 & 10.5 & 6.6 & \textbf{68.5} & \textbf{9.5} & \textbf{5.1} \\
+TTA & \textcolor{gray}{4.3} & \textcolor{gray}{2.6} & \textcolor{gray}{1.3} & \textbf{\textcolor{gray}{3.6}} & \textbf{\textcolor{gray}{2.1}} & \textbf{\textcolor{gray}{1.7}} &
\textcolor{gray}{8.5} & \textcolor{gray}{7.3} & \textcolor{gray}{3.1} & \textbf{\textcolor{gray}{7.5}} & \textbf{\textcolor{gray}{2.6}} & \textbf{\textcolor{gray}{1.6}} &
\textcolor{gray}{8.3} & \textcolor{gray}{4.9} & \textcolor{gray}{1.9} & \textbf{\textcolor{gray}{7.9}} & \textbf{\textcolor{gray}{2.1}} & \textbf{\textcolor{gray}{1.8}} \\
\addlinespace
SegCNN+DAE & 
90.3 & 6.2 & 2.5 & \textbf{91.2} & \textbf{5.7} & \textbf{2.1} & 
80.1 & 8.9 & 4.2 & \textbf{81.1} & \textbf{7.7} & \textbf{3.8} & 
69.1 & 9.9 & 5.4 & \textbf{70.2} & \textbf{9.0} & \textbf{4.9} \\
+TTA+Atlas & \textcolor{gray}{4.0} & \textcolor{gray}{2.4} & \textcolor{gray}{1.0} & \textbf{\textcolor{gray}{3.9}} & \textbf{\textcolor{gray}{2.1}} & \textbf{\textcolor{gray}{1.0}} &
\textcolor{gray}{7.3} & \textcolor{gray}{6.0} & \textcolor{gray}{2.0} & \textbf{\textcolor{gray}{7.1}} & \textbf{\textcolor{gray}{2.0}} & \textbf{\textcolor{gray}{1.2}} &
\textcolor{gray}{7.6} & \textcolor{gray}{4.1} & \textcolor{gray}{1.5} & \textbf{\textcolor{gray}{7.5}} & \textbf{\textcolor{gray}{1.6}} & \textbf{\textcolor{gray}{1.1}} \\
\midrule
Mean of Baselines & 
89.5 & 6.8 & 2.7 & \textbf{90.9} & \textbf{5.6} & \textbf{2.3} & 
77.6 & 12.6 & 5.2 & \textbf{80.4} & \textbf{8.1} & \textbf{4.1} & 
67.5 & 14.1 & 6.5 & \textbf{70.2} & \textbf{8.8} & \textbf{5.0} \\
& \textcolor{gray}{6.6} & \textcolor{gray}{4.5} & \textcolor{gray}{1.7} & \textbf{\textcolor{gray}{5.2}} & \textbf{\textcolor{gray}{2.3}} & \textbf{\textcolor{gray}{1.3}} &
\textcolor{gray}{10.3} & \textcolor{gray}{12.5} & \textcolor{gray}{4.0} & \textbf{\textcolor{gray}{8.3}} & \textbf{\textcolor{gray}{2.2}} & \textbf{\textcolor{gray}{1.5}} &
\textcolor{gray}{10.5} & \textcolor{gray}{12.2} & \textcolor{gray}{3.9} & \textbf{\textcolor{gray}{10.1}} & \textbf{\textcolor{gray}{2.2}} & \textbf{\textcolor{gray}{1.8}} \\
\bottomrule
\end{tabular}
\end{minipage}%
}
\label{tab:neuroretinal_rim}
\end{table*}

{\bf MedSAM Bounding-Box Strategy.}
First, MedSAM's extensive training on over 1.5 million medical image-mask pairs \citep{SAM_Medical} leads to a
high risk of overlap between its training set and our test sets (giving it a potential undue
advantage). Second, MedSAM leads to fundamental limitations in our applications:
(i)~MedSAM is highly sensitive to the placement and size of its bounding-box prompt, and
(ii)~MedSAM fails to generalize to anatomical topologies (e.g., ring shapes for myocardium and neuroretinal
rim) that were absent in its training data, because its training was heavily biased towards genus-0 \citep{genus_0_2024} objects.
Consequently, a naively designed bounding-box prompt produces unusable results
(Figure~\ref{fig:medsam_segmentation}).
Nevertheless, to establish a competitive baseline, we used an ``oracle prompting'' strategy that utilizes
ground-truth information (of course, this is unfair to all other baselines) as follows:
(i)~to segment ring-like structures, we use MedSAM using a two-stage strategy that models the ring as a
``subtraction'' of an inner genus-0 segment from an outer genus-0 segment
(Figure~\ref{fig:medsam_segmentation}(a1)-(b4) for myocardium; Figure~\ref{fig:medsam_segmentation}(c1)-(d4)
for neuroretinal rim);
(ii)~for genus-0 structures, e.g., the prostate and fetal head, we use a single bounding box
(Figure~\ref{fig:medsam_segmentation}(e1)-(f4));
(iii)~our bounding-box prompts use bounding boxes as expanded versions of the ground-truth bounding box, where
the expansion adds a random margin of 10-30\% of the box dimensions.

\begin{table*}[!t]
\caption
{
{\bf Results--Quantitative: Myocardium -- Measuring Calibration between Per-Pixel Segmentation
Uncertainty and Per-Pixel Segmentation Error.}
All models were trained on CAP (ID), and evaluated on ACDC and ACMRI (both OOD).
For each method-dataset combination, we report the mean (top row) and standard deviation (bottom row; in
\textcolor{gray}{gray}) for NCC~($\uparrow$), US~($\uparrow$), and TACE~($\downarrow$) metrics.
Augmenting each baseline with our VarDeepPCA shows better calibration.
{\bf Bold-font} values in the columns indicate a statistically significant improvement of
the Baseline+VarDeepPCA method over the underlying Baseline method, using a one-tailed paired-sample
t-test (p $<$ 0.05).
}
\centering
\resizebox{.96\textwidth}{!}{%
\begin{minipage}{\textwidth}
\centering
\footnotesize
\renewcommand{\arraystretch}{0.95}
\setlength{\tabcolsep}{2.5pt}
\begin{tabular}{@{}l| *{6}{c} @{\hspace{8pt}}| *{6}{c} @{\hspace{8pt}}| *{6}{c}@{}}        
\toprule
\multirow{3}{*}{} & 
\multicolumn{6}{c}{\textbf{CAP (ID)}} & 
\multicolumn{6}{c}{\textbf{ACDC (OOD)}} & 
\multicolumn{6}{c}{\textbf{ACMRI (OOD)}} \\

\cmidrule(lr){2-7} \cmidrule(lr){8-13} \cmidrule(lr){14-19} 
& \multicolumn{3}{c}{Baseline} & \multicolumn{3}{c}{Baseline +} &
\multicolumn{3}{c}{Baseline} & \multicolumn{3}{c}{Baseline +} &
\multicolumn{3}{c}{Baseline} & \multicolumn{3}{c}{Baseline +} \\

& \multicolumn{3}{c}{} & \multicolumn{3}{c}{\bf VarDeepPCA} &
\multicolumn{3}{c}{} & \multicolumn{3}{c}{\bf VarDeepPCA} &
\multicolumn{3}{c}{} & \multicolumn{3}{c}{\bf VarDeepPCA} \\
\cmidrule(lr){2-4} \cmidrule(lr){5-7} \cmidrule(lr){8-10} \cmidrule(lr){11-13} \cmidrule(lr){14-16} \cmidrule(lr){17-19}
& NCC & US & TACE & NCC & US & TACE & 
NCC & US & TACE & NCC & US & TACE & 
NCC & US & TACE & NCC & US & TACE \\
\midrule
PHISeg & 
0.53 & 0.83 & 0.23 & \textbf{0.58} & \textbf{0.85} & \textbf{0.18} & 
0.41 & 0.73 & 0.35 & \textbf{0.48} & \textbf{0.88} & \textbf{0.16} & 
0.52 & 0.75 & 0.32 & \textbf{0.56} & \textbf{0.85} & \textbf{0.10} \\
& \textcolor{gray}{0.04} & \textcolor{gray}{0.03} & \textcolor{gray}{0.05} & \textbf{\textcolor{gray}{0.03}} & \textbf{\textcolor{gray}{0.02}} & \textbf{\textcolor{gray}{0.03}} &
\textcolor{gray}{0.05} & \textcolor{gray}{0.08} & \textcolor{gray}{0.08} & \textbf{\textcolor{gray}{0.05}} & \textbf{\textcolor{gray}{0.07}} & \textbf{\textcolor{gray}{0.03}} &
\textcolor{gray}{0.04} & \textcolor{gray}{0.07} & \textcolor{gray}{0.06} & \textbf{\textcolor{gray}{0.05}} & \textbf{\textcolor{gray}{0.06}} & \textbf{\textcolor{gray}{0.02}} \\
\addlinespace
ProbUNet & 
0.34 & 0.75 & 0.29 & \textbf{0.50} & \textbf{0.91} & \textbf{0.19} & 
0.23 & 0.63 & 0.31 & \textbf{0.46} & \textbf{0.89} & \textbf{0.21} & 
0.33 & 0.73 & 0.34 & \textbf{0.45} & \textbf{0.87} & \textbf{0.12} \\
& \textcolor{gray}{0.06} & \textcolor{gray}{0.01} & \textcolor{gray}{0.06} & \textbf{\textcolor{gray}{0.03}} & \textbf{\textcolor{gray}{0.01}} & \textbf{\textcolor{gray}{0.03}} &
\textcolor{gray}{0.05} & \textcolor{gray}{0.04} & \textcolor{gray}{0.12} & \textbf{\textcolor{gray}{0.04}} & \textbf{\textcolor{gray}{0.04}} & \textbf{\textcolor{gray}{0.03}} &
\textcolor{gray}{0.08} & \textcolor{gray}{0.04} & \textcolor{gray}{0.10} & \textbf{\textcolor{gray}{0.04}} & \textbf{\textcolor{gray}{0.05}} & \textbf{\textcolor{gray}{0.02}} \\
\addlinespace
HPUNet & 
0.45 & 0.73 & 0.18 & \textbf{0.50} & \textbf{0.85} & \textbf{0.10} & 
0.34 & 0.61 & 0.49 & \textbf{0.43} & \textbf{0.75} & \textbf{0.26} & 
0.42 & 0.57 & 0.31 & \textbf{0.47} & \textbf{0.82} & \textbf{0.19} \\
& \textcolor{gray}{0.03} & \textcolor{gray}{0.02} & \textcolor{gray}{0.07} & \textbf{\textcolor{gray}{0.03}} & \textbf{\textcolor{gray}{0.02}} & \textbf{\textcolor{gray}{0.03}} &
\textcolor{gray}{0.04} & \textcolor{gray}{0.04} & \textcolor{gray}{0.07} & \textbf{\textcolor{gray}{0.05}} & \textbf{\textcolor{gray}{0.04}} & \textbf{\textcolor{gray}{0.02}} &
\textcolor{gray}{0.05} & \textcolor{gray}{0.05} & \textcolor{gray}{0.09} & \textbf{\textcolor{gray}{0.04}} & \textbf{\textcolor{gray}{0.04}} & \textbf{\textcolor{gray}{0.02}} \\
\addlinespace
\bottomrule
\end{tabular}
\end{minipage}%
}
\label{tab:myo_calibration}
\end{table*}

\subsection{Implementation Details for Our VarDeepPCA}

We demonstrate the benefits of VarDeepPCA using 14 publicly-available datasets and 15 existing DNN
segmenters.
VarDeepPCA's architecture uses:
(i)~an encoder $\mathcal{E}(\cdot; \theta^\mathcal{E})$ having a sequence of convolution and max-pooling
layers that progressively increases the number of feature channels
(32$\rightarrow$64$\rightarrow$128$\rightarrow$256) while reducing spatial dimensions by 2$\times$ at each
pooling stage
(256$\rightarrow$128$\rightarrow$64$\rightarrow$32$\rightarrow$16), followed by a fully-connected bottleneck
layer that projects the flattened 256$\times$16$\times$16 feature map to a low-dimensional latent vector
$F \in 1\times 1\times \mathbb{R}^K$ (in this paper, $K$ is tunable and set to the optimal value for each
application),
and (ii)~a corresponding decoder $\mathcal{D}(\cdot; \theta^\mathcal{D})$ that applies a softmax activation to
$F$ before using a fully-connected expansion layer and transpose-convolutions for upsampling back to the
original 256$\times$256 resolution.
Training uses Adam~\citep{Kingma2015AdamAM}, a batch normalization after each convolution layer, and four
independent runs from which we choose the best model.
This architecture achieves dimensionality reduction by compressing high-dimensional spatial features into a
$K$-dimensional representation, enabling efficient latent space analysis while learning to preserve
reconstruction quality through the decoder pathway.
The specific model configurations were tailored to each application: $K = 16$ (2.72M parameters) for the myocardium; $K = 8$ (1.67M parameters) for both the neuroretinal rim and prostate; and $K = 3$ (1.02M parameters) for the fetal head.
All implementations use PyTorch~\citep{paszke2019pytorch}.
In the interest of reproducibility, the code and models used in this study will be made publicly available
after publication.

\begin{table*}[!b]
\caption
{
{\bf Results--Quantitative: Neuroretinal Rim -- Measuring Calibration between Per-Pixel Segmentation
Uncertainty and Per-Pixel Segmentation Error.}
All models were trained on MAGRABI (ID), and evaluated on G1020 and ORIGA (both OOD).
For each method-dataset combination, we report the mean (top row) and standard deviation (bottom row; in
\textcolor{gray}{gray}) for NCC~($\uparrow$), US~($\uparrow$), and TACE~($\downarrow$) metrics.
Augmenting each baseline with our VarDeepPCA shows better calibration.
{\bf Bold-font} values in the columns indicate a statistically significant improvement of
the Baseline+VarDeepPCA method over the underlying Baseline method, using a one-tailed paired-sample
t-test (p $<$ 0.05).
}
\centering
\resizebox{0.96\textwidth}{!}{%
\begin{minipage}{\textwidth}
\centering
\footnotesize
\renewcommand{\arraystretch}{0.95}
\setlength{\tabcolsep}{2.5pt}
\begin{tabular}{@{}l| *{6}{c} @{\hspace{8pt}}| *{6}{c} @{\hspace{8pt}}| *{6}{c}@{}}        
\toprule
\multirow{3}{*}{} & 
\multicolumn{6}{c}{\textbf{MAGRABI (ID)}} & 
\multicolumn{6}{c}{\textbf{G1020 (OOD)}} & 
\multicolumn{6}{c}{\textbf{ORIGA (OOD)}} \\

\cmidrule(lr){2-7} \cmidrule(lr){8-13} \cmidrule(lr){14-19} 
& \multicolumn{3}{c}{Baseline} & \multicolumn{3}{c}{Baseline +} &
\multicolumn{3}{c}{Baseline} & \multicolumn{3}{c}{Baseline +} &
\multicolumn{3}{c}{Baseline} & \multicolumn{3}{c}{Baseline +} \\

& \multicolumn{3}{c}{} & \multicolumn{3}{c}{\bf VarDeepPCA} &
\multicolumn{3}{c}{} & \multicolumn{3}{c}{\bf VarDeepPCA} &
\multicolumn{3}{c}{} & \multicolumn{3}{c}{\bf VarDeepPCA} \\
\cmidrule(lr){2-4} \cmidrule(lr){5-7} \cmidrule(lr){8-10} \cmidrule(lr){11-13} \cmidrule(lr){14-16} \cmidrule(lr){17-19}
& NCC & US & TACE & NCC & US & TACE & 
NCC & US & TACE & NCC & US & TACE & 
NCC & US & TACE & NCC & US & TACE \\
\midrule
PHISeg & 
0.56 & 0.93 & 0.21 & \textbf{0.58} & \textbf{0.95} & \textbf{0.12} & 
0.48 & 0.82 & 0.32 & \textbf{0.57} & \textbf{0.87} & \textbf{0.13} & 
0.54 & 0.82 & 0.48 & \textbf{0.56} & \textbf{0.86} & \textbf{0.18} \\
& \textcolor{gray}{0.05} & \textcolor{gray}{0.03} & \textcolor{gray}{0.04} & \textbf{\textcolor{gray}{0.04}} & \textbf{\textcolor{gray}{0.02}} & \textbf{\textcolor{gray}{0.01}} &
\textcolor{gray}{0.07} & \textcolor{gray}{0.11} & \textcolor{gray}{0.06} & \textbf{\textcolor{gray}{0.05}} & \textbf{\textcolor{gray}{0.06}} & \textbf{\textcolor{gray}{0.01}} &
\textcolor{gray}{0.05} & \textcolor{gray}{0.04} & \textcolor{gray}{0.06} & \textbf{\textcolor{gray}{0.04}} & \textbf{\textcolor{gray}{0.03}} & \textbf{\textcolor{gray}{0.01}} \\
\addlinespace
ProbUNet & 
0.42 & 0.91 & 0.27 & \textbf{0.49} & \textbf{0.94} & \textbf{0.12} & 
0.37 & 0.85 & 0.39 & \textbf{0.43} & \textbf{0.88} & \textbf{0.13} & 
0.33 & 0.86 & 0.55 & \textbf{0.39} & \textbf{0.88} & \textbf{0.16} \\
& \textcolor{gray}{0.07} & \textcolor{gray}{0.04} & \textcolor{gray}{0.08} & \textbf{\textcolor{gray}{0.06}} & \textbf{\textcolor{gray}{0.03}} & \textbf{\textcolor{gray}{0.02}} &
\textcolor{gray}{0.08} & \textcolor{gray}{0.06} & \textcolor{gray}{0.12} & \textbf{\textcolor{gray}{0.07}} & \textbf{\textcolor{gray}{0.06}} & \textbf{\textcolor{gray}{0.01}} &
\textcolor{gray}{0.08} & \textcolor{gray}{0.07} & \textcolor{gray}{0.11} & \textbf{\textcolor{gray}{0.05}} & \textbf{\textcolor{gray}{0.06}} & \textbf{\textcolor{gray}{0.01}} \\
\addlinespace
HPUNet & 
0.45 & 0.92 & 0.19 & \textbf{0.53} & \textbf{0.95} & \textbf{0.10} & 
0.44 & 0.88 & 0.40 & \textbf{0.49} & \textbf{0.92} & \textbf{0.13} & 
0.43 & 0.88 & 0.45 & \textbf{0.53} & \textbf{0.93} & \textbf{0.12} \\
& \textcolor{gray}{0.04} & \textcolor{gray}{0.03} & \textcolor{gray}{0.09} & \textbf{\textcolor{gray}{0.05}} & \textbf{\textcolor{gray}{0.02}} & \textbf{\textcolor{gray}{0.01}} &
\textcolor{gray}{0.07} & \textcolor{gray}{0.06} & \textcolor{gray}{0.12} & \textbf{\textcolor{gray}{0.07}} & \textbf{\textcolor{gray}{0.04}} & \textbf{\textcolor{gray}{0.01}} &
\textcolor{gray}{0.05} & \textcolor{gray}{0.04} & \textcolor{gray}{0.09} & \textbf{\textcolor{gray}{0.05}} & \textbf{\textcolor{gray}{0.04}} & \textbf{\textcolor{gray}{0.01}} \\
\addlinespace
\bottomrule
\end{tabular}
\end{minipage}%
}
\label{tab:fundus_calibration}
\end{table*}

\begin{figure}[!t]
\threeAcrossLabelsWidth{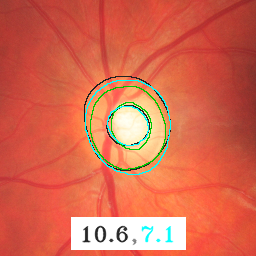}
{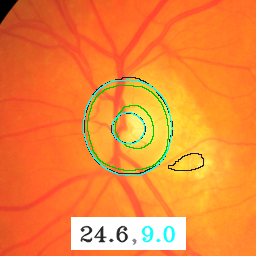}
{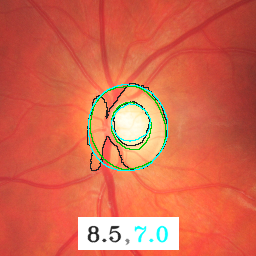}
{\scs{\bf(a)}~SegCNN+DAE+TTA+Atlas}{\scs{\bf(b)}~MedSegDiff}{\scs{\bf(c)}~MedSAM}{0.48}
\threeAcrossLabelsWidth{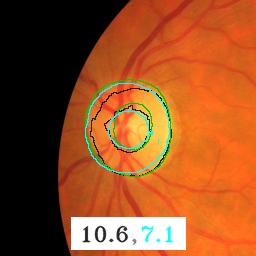}
{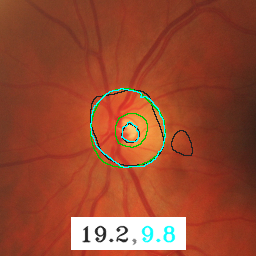}
{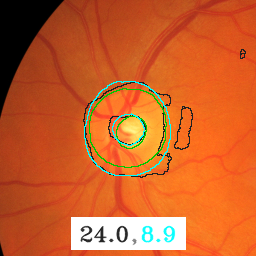}
{\scs{\bf(d)}~PHISeg}{\scs{\bf(e)}~ProbUNet}{\scs{\bf(f)}~HierProbUNet}{0.48}
\threeAcrossLabelsWidth{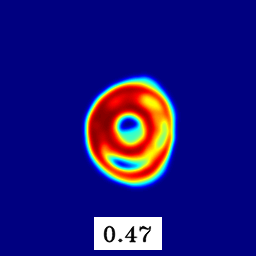}
{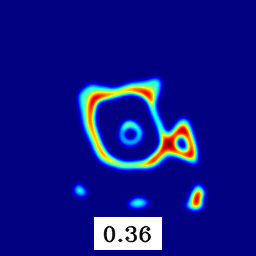}
{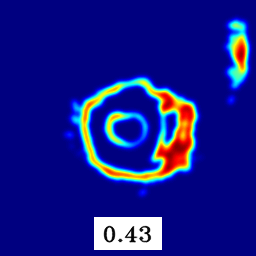}
{\scs{\bf(g)}~Unc. for (d)~PHISeg}{\scs{\bf(h)}~Unc. for (e)~ProbUNet}{\scs{\bf(i)}~Unc.  for (f)~HierProbUNet}{0.48}
\threeAcrossLabelsWidth{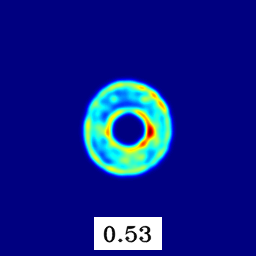}
{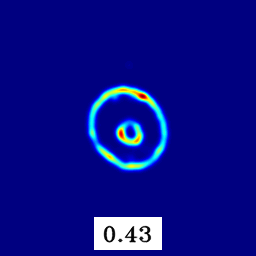}
{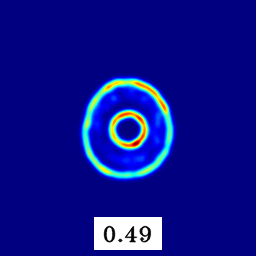}
{\scs{\bf(j)}~Unc. for (d)+VarDeepPCA}{\scs{\bf(k)}~Unc. for (e)+VarDeepPCA}{\scs{\bf(l)}~Unc. for (f)+VarDeepPCA}{0.48}
\caption
{
{\bf Results--Qualitative: Neuroretinal Rim Segmentation Restoration on G1020 (OOD) data. }
{\bf (a)--(c)}~Results on images for the best non-variational baselines.
{\bf (d)--(f)}~Results on images for the variational baselines. 
{\bf (g)--(i)}~Uncertainty maps produced using variational baselines.
{\bf (j)--(l)}~Uncertainty maps produced using VarDeepPCA when plugged into the associated baselines
(d)--(f).
Color scheme in (a)--(f):
\textcolor{neuro_black}{Baseline};
\textcolor{neuro_cyan}{Baseline+VarDeepPCA (Ours)};
\textcolor{neuro_green}{Ground Truth}.
HD95 numbers in (a)--(f) indicate that the examples were representative of the test set, because the HD95
values were close to the mean HD95 reported in Table~\ref{tab:neuroretinal_rim}.
NCC numbers in (g)--(l) indicate that the examples were representative of the test set, because the NCC
values were close to the mean of the NCC reported in Table~\ref{tab:fundus_calibration}.
}
\label{fig:neurorim_g1020_auto}
\end{figure}

\begin{figure}[!t]
\threeAcrossLabelsWidth{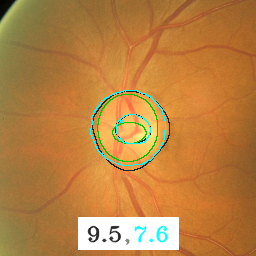}
{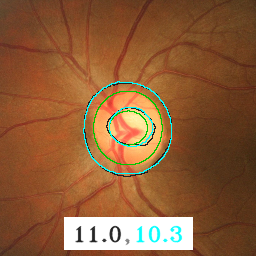}
{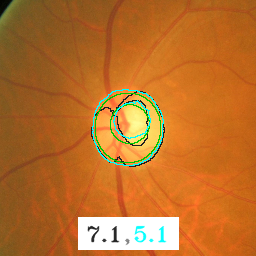}
{\scs{\bf(a)}~SegCNN+DAE+TTA+Atlas}{\scs{\bf(b)}~MedSegDiff}{\scs{\bf(c)}~MedSAM}{0.48}
\threeAcrossLabelsWidth{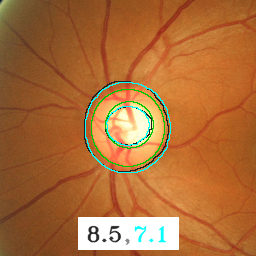}
{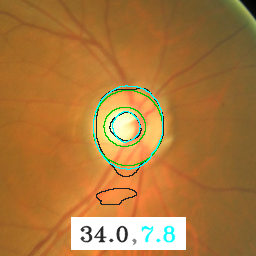}
{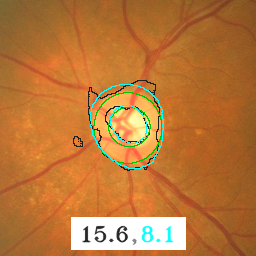}
{\scs{\bf(d)}~PHISeg}{\scs{\bf(e)}~ProbUNet}{\scs{\bf(f)}~HierProbUNet}{0.48}
\threeAcrossLabelsWidth{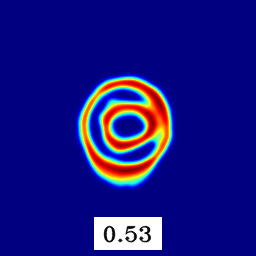}
{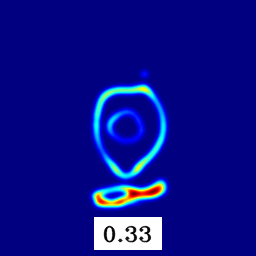}
{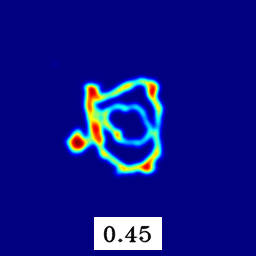}
{\scs{\bf(g)}~Unc. for (d)~PHISeg}{\scs{\bf(h)}~Unc. for (e)~ProbUNet}{\scs{\bf(i)}~Unc.  for (f)~HierProbUNet}{0.48}
\threeAcrossLabelsWidth{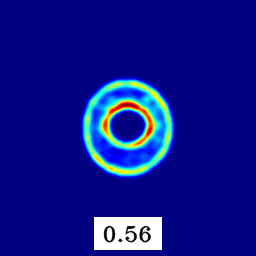}
{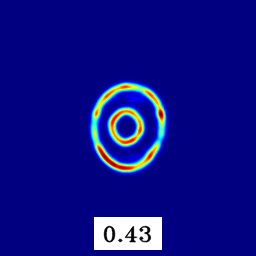}
{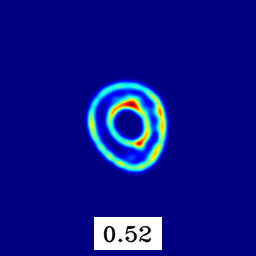}
{\scs{\bf(j)}~Unc. for (d)+VarDeepPCA}{\scs{\bf(k)}~Unc. for (e)+VarDeepPCA}{\scs{\bf(l)}~Unc. for (f)+VarDeepPCA}{0.48}
\caption
{
{\bf Results--Qualitative: Neuroretinal Rim Segmentation Restoration on ORIGA (OOD) data.}
{\bf (a)--(c)}~Results on images for the best non-variational baselines. 
{\bf (d)--(f)}~Results on images for the variational baselines.
{\bf (g)--(i)}~Uncertainty maps produced using variational baselines.
{\bf (j)--(l)}~Uncertainty maps produced using VarDeepPCA when plugged into the associated baselines
(d)--(f).
Color scheme in (a)--(f):
\textcolor{neuro_black}{Baseline};
\textcolor{neuro_cyan}{Baseline+VarDeepPCA (Ours)};
\textcolor{neuro_green}{Ground Truth}.
HD95 numbers in (a)--(f) indicate that the examples were representative of the test set, because the HD95
values were close to the mean HD95 reported in Table~\ref{tab:neuroretinal_rim}.
NCC numbers in (g)--(l) indicate that the examples were representative of the test set, because the NCC values were close to the mean of the NCC reported in Table~\ref{tab:fundus_calibration}.
}
\label{fig:neurorim_origa_auto}
\end{figure}

\subsection{Results: Quantitative and Qualitative}

A comprehensive analysis across all four application domains reveals degraded OOD performance for many
early-DNN methods and some variational methods.
While the variational model PHISeg demonstrated considerable robustness to OOD data and provided meaningful
uncertainty maps, other large architectures, e.g., BASNet, MedSegDiff, and DSTransUNet, showed only occasional
robustness depending on the specific application and the specific OOD dataset.
We find that SegCNN+DAE+TTA+Atlas shows robust segmentation performance on OOD data in some applications.
%
Our VarDeepPCA framework consistently refined the results, with often significantly improving the results
produced by the baseline methods.
In datasets where baseline performance had poor accuracy and poor precision (exhibiting high variance) in
terms of HD95, VarDeepPCA not only improved the accuracy but also the precision. The reduction in variance may
be attributed to VarDeepPCA's ability to learn the manifold/distribution of anatomically valid geometries.
For qualitative analysis, we show representative examples for which the HD95 value comes close to the mean
HD95 value across the entire test set.

{\bf Myocardium Segmentation.}
%
%
On the CAP (ID) test set, most baseline models perform well (Table~\ref{tab:myo}), achieving a mean DSC $\geq$
86\% and HD95 $\leq$ 6.5; this implies that the models fitted well (avoiding overfitting) to the training
data.
As anticipated, when the same models are applied to OOD datasets (ACDC and ACMRI), we observe a significant
degradation in performance across all metrics; this confirms the baseline models' lack of robustness to domain
shifts.
Our primary finding is that plugging in our VarDeepPCA framework consistently and significantly lowers the
HD95 and ASD values across all baselines on these OOD datasets.
For these experiments, we use a latent dimension of $K = 16$ for VarDeepPCA, as justified in our sensitivity
analysis in Section~\ref{subsec:sen}.
This demonstrates that VarDeepPCA successfully filters the degraded segmentation maps and projects them onto
the learned manifold of anatomically plausible geometries.
Notably, VarDeepPCA also improves the boundary metrics (HD95, ASD) even on the ID test set. 
This suggests that our model is not just correcting for large OOD shifts but also filtering minor anatomical
inaccuracies present in the baselines' ID outputs.
While the corresponding gains in DSC are more modest (since this metric is less sensitive to the boundary
errors that are the focus of our work), we argue that the substantial improvement in boundary-based metrics
brings the segmentations closer to the ground truth, achieving a level of accuracy more suitable for clinical
applications.
Indeed, for the OOD datasets of ACDC and ACMRI, the mean HD95 values after employing the VarDeepPCA
plugin reduce from an average (across all methods; before VarDeepPCA) within 13.2-13.6 to an average (across
all methods; after VarDeepPCA) within 7.2-8, which is far more clinically acceptable as per our analysis in
Section~\ref{sec:UpperBoundClinUtil}.
We also present the quantitative results for the uncertainty calibration metrics, i.e., the NCC, US, and
TACE scores in Table~\ref{tab:myo_calibration}. For the PHISeg, ProbUNet, and HPUNet baselines,
incorporating the VarDeepPCA plugin leads to a significant improvement of these scores in both ID and OOD
datasets.

\begin{table*}[!t]
\caption
{
{\bf Results--Quantitative: Prostate Segmentation.}
All models were trained on BIDMC+BMC (ID), and evaluated on HK+I2CVB and RUNMC+UCL (both OOD).
For each method-dataset combination, we report mean (top row) and standard deviation (bottom row, in
\textcolor{gray}{gray}) for DSC($\uparrow$), HD95($\downarrow$), and ASD($\downarrow$).
Augmenting each baseline with our VarDeepPCA consistently improves performance.
{\bf Bold-font} values in the columns indicate a statistically significant improvement of
the Baseline+VarDeepPCA method over the underlying Baseline method, using a one-tailed paired-sample
t-test (p $<$ 0.05).
}
\centering
\resizebox{0.96\textwidth}{!}{%
\begin{minipage}{\textwidth}
\centering
\footnotesize
\renewcommand{\arraystretch}{0.95}
\setlength{\tabcolsep}{2.5pt}
\begin{tabular}{@{}l| *{6}{c} @{\hspace{8pt}}| *{6}{c} @{\hspace{8pt}}| *{6}{c}@{}}     
\toprule
\multirow{3}{*}{\textbf{Models}} & 
\multicolumn{6}{c}{\textbf{BIDMC+BMC (ID)}} & 
\multicolumn{6}{c}{\textbf{HK+I2CVB (OOD)}} & 
\multicolumn{6}{c}{\textbf{RUNMC+UCL (OOD)}} \\
\cmidrule(lr){2-7} \cmidrule(lr){8-13} \cmidrule(lr){14-19}
& \multicolumn{3}{c}{Baseline} & \multicolumn{3}{c}{Baseline +} &
\multicolumn{3}{c}{Baseline} & \multicolumn{3}{c}{Baseline +} &
\multicolumn{3}{c}{Baseline} & \multicolumn{3}{c}{Baseline +} \\
& \multicolumn{3}{c}{} & \multicolumn{3}{c}{\bf VarDeepPCA} &
\multicolumn{3}{c}{} & \multicolumn{3}{c}{\bf VarDeepPCA} &
\multicolumn{3}{c}{} & \multicolumn{3}{c}{\bf VarDeepPCA} \\
\cmidrule(lr){2-4} \cmidrule(lr){5-7} \cmidrule(lr){8-10} \cmidrule(lr){11-13} \cmidrule(lr){14-16} \cmidrule(lr){17-19}
& DSC & HD95 & ASD & DSC & HD95 & ASD & 
DSC  & HD95 & ASD & DSC & HD95 & ASD & 
DSC & HD95 & ASD & DSC & HD95 & ASD \\
\midrule
UNet & 
93.4 & 7.4 & 2.5 & 93.7 & \textbf{4.6} & \textbf{1.9} & 
84.3 & 26.8 & 8.4 & \textbf{88.0} & \textbf{7.3} & \textbf{2.9} & 
89.4 & 11.6 & 4.0 & \textbf{90.6} & \textbf{6.7} & \textbf{2.6} \\
& \textcolor{gray}{3.9} & \textcolor{gray}{13.8} & \textcolor{gray}{2.4} & \textcolor{gray}{3.4} & \textbf{\textcolor{gray}{2.2}} & \textbf{\textcolor{gray}{0.9}} &
\textcolor{gray}{7.4} & \textcolor{gray}{23.9} & \textcolor{gray}{6.2} & \textbf{\textcolor{gray}{4.6}} & \textbf{\textcolor{gray}{2.6}} & \textbf{\textcolor{gray}{1.1}} &
\textcolor{gray}{5.9} & \textcolor{gray}{13.6} & \textcolor{gray}{4.1} & \textbf{\textcolor{gray}{4.3}} & \textbf{\textcolor{gray}{2.8}} & \textbf{\textcolor{gray}{1.0}} \\
\addlinespace
AttnUNet & 
93.8 & 10.7 & 3.2 & 93.9 & \textbf{4.7} & \textbf{1.9} & 
86.8 & 18.6 & 6.4 & \textbf{89.4} & \textbf{6.7} & \textbf{2.7} & 
91.3 & 13.8 & 4.6 & \textbf{91.9} & \textbf{5.8} & \textbf{2.4} \\
& \textcolor{gray}{2.7} & \textcolor{gray}{21.1} & \textcolor{gray}{4.8} & \textcolor{gray}{2.5} & \textbf{\textcolor{gray}{1.9}} & \textbf{\textcolor{gray}{0.7}} &
\textcolor{gray}{9.9} & \textcolor{gray}{22.6} & \textcolor{gray}{8.5} & \textbf{\textcolor{gray}{4.9}} & \textbf{\textcolor{gray}{2.9}} & \textbf{\textcolor{gray}{1.2}} &
\textcolor{gray}{4.6} & \textcolor{gray}{22.2} & \textcolor{gray}{6.3} & \textbf{\textcolor{gray}{3.7}} & \textbf{\textcolor{gray}{2.5}} & \textbf{\textcolor{gray}{1.1}} \\
\addlinespace
ResUNet++ & 
93.9 & 5.1 & 1.9 & 94.1 & 4.8 & 1.9 & 
86.8 & 11.1 & 4.0 & \textbf{87.6} & \textbf{7.5} & \textbf{3.1} & 
90.7 & 8.6 & 3.2 & 90.8 & \textbf{6.7} & \textbf{2.7} \\
& \textcolor{gray}{2.6} & \textcolor{gray}{3.5} & \textcolor{gray}{1.0} & \textcolor{gray}{2.4} & \textcolor{gray}{2.0} & \textcolor{gray}{0.8} &
\textcolor{gray}{7.0} & \textcolor{gray}{11.1} & \textcolor{gray}{3.1} & \textbf{\textcolor{gray}{6.1}} & \textbf{\textcolor{gray}{3.2}} & \textbf{\textcolor{gray}{1.5}} &
\textcolor{gray}{4.5} & \textcolor{gray}{8.5} & \textcolor{gray}{2.3} & \textcolor{gray}{4.3} & \textbf{\textcolor{gray}{2.8}} & \textbf{\textcolor{gray}{1.2}} \\
\addlinespace
DeepLabV3+ & 
91.3 & 6.4 & 2.7 & 91.3 & 6.1 & 2.6 & 
82.9 & 16.5 & 6.0 & \textbf{86.0} & \textbf{8.2} & \textbf{3.5} & 
88.9 & 8.3 & 3.4 & 89.3 & \textbf{6.7} & \textbf{3.0} \\
& \textcolor{gray}{3.2} & \textcolor{gray}{2.6} & \textcolor{gray}{1.3} & \textcolor{gray}{3.2} & \textcolor{gray}{2.4} & \textcolor{gray}{1.0} &
\textcolor{gray}{13.3} & \textcolor{gray}{17.6} & \textcolor{gray}{6.2} & \textbf{\textcolor{gray}{6.5}} & \textbf{\textcolor{gray}{1.2}} & \textbf{\textcolor{gray}{0.8}} &
\textcolor{gray}{4.6} & \textcolor{gray}{6.4} & \textcolor{gray}{1.8} & \textcolor{gray}{4.0} & \textbf{\textcolor{gray}{1.8}} & \textbf{\textcolor{gray}{0.9}} \\
\addlinespace
BASNet & 
94.7 & 6.3 & 2.0 & 94.9 & \textbf{4.2} & \textbf{1.6} & 
87.7 & 22.9 & 8.0 & \textbf{91.7} & \textbf{5.5} & \textbf{2.3} & 
92.0 & 11.5 & 3.9 & \textbf{92.7} & \textbf{5.3} & \textbf{2.2} \\
& \textcolor{gray}{2.5} & \textcolor{gray}{12.0} & \textcolor{gray}{2.5} & \textcolor{gray}{2.1} & \textbf{\textcolor{gray}{1.8}} & \textbf{\textcolor{gray}{0.7}} &
\textcolor{gray}{9.6} & \textcolor{gray}{27.5} & \textcolor{gray}{10.0} & \textbf{\textcolor{gray}{3.8}} & \textbf{\textcolor{gray}{2.5}} & \textbf{\textcolor{gray}{1.1}} &
\textcolor{gray}{5.1} & \textcolor{gray}{19.5} & \textcolor{gray}{6.1} & \textbf{\textcolor{gray}{3.4}} & \textbf{\textcolor{gray}{2.4}} & \textbf{\textcolor{gray}{0.9}} \\
\addlinespace
SegAN & 
92.9 & 5.5 & 2.2 & 93.1 & \textbf{5.1} & \textbf{2.0} & 
87.6 & 9.3 & 3.5 & \textbf{88.1} & \textbf{7.4} & \textbf{3.0} & 
90.6 & 7.2 & 2.8 & 90.8 & \textbf{6.2} & \textbf{2.5} \\
& \textcolor{gray}{3.2} & \textcolor{gray}{2.7} & \textcolor{gray}{1.0} & \textcolor{gray}{3.1} & \textbf{\textcolor{gray}{2.3}} & \textbf{\textcolor{gray}{0.9}} &
\textcolor{gray}{4.7} & \textcolor{gray}{7.1} & \textcolor{gray}{2.0} & \textbf{\textcolor{gray}{4.4}} & \textbf{\textcolor{gray}{2.3}} & \textbf{\textcolor{gray}{1.0}} &
\textcolor{gray}{3.9} & \textcolor{gray}{6.1} & \textcolor{gray}{1.4} & \textcolor{gray}{3.8} & \textbf{\textcolor{gray}{2.4}} & \textbf{\textcolor{gray}{1.0}} \\
\addlinespace
MedSegDiff & 
90.3 & 24.5 & 7.4 & \textbf{92.5} & \textbf{5.5} & \textbf{2.3} & 
79.1 & 47.5 & 16.8 & \textbf{86.5} & \textbf{7.9} & \textbf{3.5} & 
72.4 & 73.2 & 28.1 & \textbf{88.2} & \textbf{7.9} & \textbf{3.5} \\
& \textcolor{gray}{5.8} & \textcolor{gray}{34.6} & \textcolor{gray}{9.9} & \textbf{\textcolor{gray}{3.6}} & \textbf{\textcolor{gray}{2.4}} & \textbf{\textcolor{gray}{1.2}} &
\textcolor{gray}{11.6} & \textcolor{gray}{39.7} & \textcolor{gray}{15.0} & \textbf{\textcolor{gray}{4.5}} & \textbf{\textcolor{gray}{2.1}} & \textbf{\textcolor{gray}{1.1}} &
\textcolor{gray}{12.5} & \textcolor{gray}{30.1} & \textcolor{gray}{14.6} & \textbf{\textcolor{gray}{3.8}} & \textbf{\textcolor{gray}{2.2}} & \textbf{\textcolor{gray}{1.1}} \\
\addlinespace
DSTransUNet & 
94.3 & 6.8 & 2.2 & 94.3 & \textbf{4.6} & \textbf{1.9} & 
91.0 & 10.3 & 3.8 & 91.4 & \textbf{6.2} & \textbf{2.5} & 
91.0 & 12.9 & 4.4 & \textbf{91.5} & \textbf{6.1} & \textbf{2.6} \\
& \textcolor{gray}{2.5} & \textcolor{gray}{10.5} & \textcolor{gray}{2.0} & \textcolor{gray}{2.4} & \textbf{\textcolor{gray}{2.0}} & \textbf{\textcolor{gray}{0.8}} &
\textcolor{gray}{4.9} & \textcolor{gray}{14.2} & \textcolor{gray}{4.9} & \textcolor{gray}{4.1} & \textbf{\textcolor{gray}{3.0}} & \textbf{\textcolor{gray}{1.3}} &
\textcolor{gray}{5.1} & \textcolor{gray}{17.8} & \textcolor{gray}{4.9} & \textbf{\textcolor{gray}{4.2}} & \textbf{\textcolor{gray}{2.5}} & \textbf{\textcolor{gray}{1.2}} \\
\addlinespace
VMUNet & 
91.5 & 5.9 & 2.6 & 91.8 & 5.6 & 2.4 & 
84.9 & 8.7 & 4.1 & \textbf{85.1} & \textbf{8.4} & 3.9 & 
87.8 & 7.6 & 3.5 & 88.0 & 7.4 & 3.5 \\
& \textcolor{gray}{3.4} & \textcolor{gray}{2.1} & \textcolor{gray}{1.0} & \textcolor{gray}{3.3} & \textcolor{gray}{2.0} & \textcolor{gray}{1.0} &
\textcolor{gray}{6.1} & \textcolor{gray}{2.1} & \textcolor{gray}{1.2} & \textbf{\textcolor{gray}{5.2}} & \textbf{\textcolor{gray}{1.9}} & \textcolor{gray}{1.1} &
\textcolor{gray}{4.8} & \textcolor{gray}{2.8} & \textcolor{gray}{1.3} & \textcolor{gray}{4.5} & \textcolor{gray}{2.2} & \textcolor{gray}{1.2} \\
\addlinespace
MedSAM & 
81.4 & 13.6 & 7.0 & 81.7 & \textbf{7.9} & \textbf{3.7} & 
82.6 & 11.3 & 5.4 & 82.7 & \textbf{7.0} & \textbf{3.5} & 
82.1 & 11.5 & 6.2 & 82.3 & \textbf{8.1} & \textbf{3.6} \\
& \textcolor{gray}{5.7} & \textcolor{gray}{4.4} & \textcolor{gray}{2.5} & \textcolor{gray}{5.8} & \textbf{\textcolor{gray}{3.9}} & \textbf{\textcolor{gray}{2.5}} &
\textcolor{gray}{5.5} & \textcolor{gray}{3.5} & \textcolor{gray}{2.2} & \textcolor{gray}{5.6} & \textbf{\textcolor{gray}{2.9}} & \textbf{\textcolor{gray}{2.1}} &
\textcolor{gray}{5.5} & \textcolor{gray}{4.1} & \textcolor{gray}{2.4} & \textcolor{gray}{5.6} & \textbf{\textcolor{gray}{3.9}} & \textbf{\textcolor{gray}{2.4}} \\
\addlinespace
PHISeg & 
93.1 & 5.2 & 2.1 & 93.3 & 5.1 & 2.1 & 
87.6 & 7.5 & 3.2 & 87.8 & 7.5 & 3.3 & 
90.7 & 6.4 & 2.8 & 90.9 & 6.3 & 2.8 \\
& \textcolor{gray}{2.5} & \textcolor{gray}{1.9} & \textcolor{gray}{0.8} & \textcolor{gray}{2.3} & \textcolor{gray}{1.8} & \textcolor{gray}{0.7} &
\textcolor{gray}{4.4} & \textcolor{gray}{2.4} & \textcolor{gray}{1.2} & \textcolor{gray}{4.3} & \textcolor{gray}{2.1} & \textcolor{gray}{1.1} &
\textcolor{gray}{3.7} & \textcolor{gray}{2.3} & \textcolor{gray}{1.1} & \textcolor{gray}{3.6} & \textcolor{gray}{2.1} & \textcolor{gray}{1.0} \\
\addlinespace
ProbUNet & 
91.4 & 11.3 & 3.6 & \textbf{91.9} & \textbf{5.8} & \textbf{2.4} & 
83.3 & 18.7 & 6.5 & \textbf{85.7} & \textbf{7.9} & \textbf{3.3} & 
86.7 & 10.9 & 4.2 & \textbf{88.1} & \textbf{7.7} & \textbf{3.2} \\
& \textcolor{gray}{3.2} & \textcolor{gray}{17.8} & \textcolor{gray}{3.8} & \textbf{\textcolor{gray}{2.8}} & \textbf{\textcolor{gray}{2.0}} & \textbf{\textcolor{gray}{0.9}} &
\textcolor{gray}{7.7} & \textcolor{gray}{17.3} & \textcolor{gray}{4.8} & \textbf{\textcolor{gray}{4.9}} & \textbf{\textcolor{gray}{2.3}} & \textbf{\textcolor{gray}{1.1}} &
\textcolor{gray}{5.1} & \textcolor{gray}{7.9} & \textcolor{gray}{2.3} & \textbf{\textcolor{gray}{4.4}} & \textbf{\textcolor{gray}{2.3}} & \textbf{\textcolor{gray}{1.0}} \\
\addlinespace
HierProbUNet & 
90.7 & 8.8 & 3.5 & 91.3 & \textbf{6.7} & \textbf{2.7} & 
77.4 & 29.4 & 11.3 & \textbf{84.8} & \textbf{7.9} & \textbf{3.7} & 
81.8 & 26.4 & 9.6 & \textbf{88.0} & \textbf{7.9} & \textbf{3.3} \\
& \textcolor{gray}{3.8} & \textcolor{gray}{7.6} & \textcolor{gray}{2.3} & \textcolor{gray}{3.5} & \textbf{\textcolor{gray}{2.5}} & \textbf{\textcolor{gray}{1.1}} &
\textcolor{gray}{10.7} & \textcolor{gray}{13.2} & \textcolor{gray}{5.6} & \textbf{\textcolor{gray}{7.4}} & \textbf{\textcolor{gray}{2.2}} & \textbf{\textcolor{gray}{1.4}} &
\textcolor{gray}{6.9} & \textcolor{gray}{12.8} & \textcolor{gray}{4.8} & \textbf{\textcolor{gray}{5.0}} & \textbf{\textcolor{gray}{2.3}} & \textbf{\textcolor{gray}{1.3}} \\
\addlinespace
SegCNN+DAE & 
92.3 & 5.3 & 2.3 & 92.7 & 5.2 & 2.1 & 
87.1 & 8.9 & 3.4 & \textbf{88.2} & \textbf{6.9} & \textbf{2.9} & 
89.4 & 7.3 & 2.9 & 89.5 & 6.8 & 2.8 \\
+TTA & \textcolor{gray}{3.4} & \textcolor{gray}{2.6} & \textcolor{gray}{1.1} & \textcolor{gray}{3.0} & \textcolor{gray}{2.5} & \textcolor{gray}{1.1} &
\textcolor{gray}{5.1} & \textcolor{gray}{7.1} & \textcolor{gray}{2.6} & \textbf{\textcolor{gray}{4.7}} & \textbf{\textcolor{gray}{2.6}} & \textbf{\textcolor{gray}{1.6}} &
\textcolor{gray}{5.3} & \textcolor{gray}{4.4} & \textcolor{gray}{2.1} & \textcolor{gray}{4.7} & \textcolor{gray}{4.3} & \textcolor{gray}{1.6} \\
\addlinespace
SegCNN+DAE & 
93.7 & 4.7 & 1.9 & 93.7 & 4.7 & 1.9 & 
88.5 & 8.0 & 3.1 & \textbf{89.3} & \textbf{6.7} & \textbf{2.8} & 
90.5 & 6.8 & 2.6 & 90.5 & 6.6 & 2.6 \\
+TTA+Atlas & \textcolor{gray}{3.0} & \textcolor{gray}{2.4} & \textcolor{gray}{0.8} & \textcolor{gray}{2.9} & \textcolor{gray}{2.2} & \textcolor{gray}{0.8} &
\textcolor{gray}{4.7} & \textcolor{gray}{6.8} & \textcolor{gray}{2.2} & \textbf{\textcolor{gray}{4.2}} & \textbf{\textcolor{gray}{2.3}} & \textbf{\textcolor{gray}{1.0}} &
\textcolor{gray}{4.9} & \textcolor{gray}{3.5} & \textcolor{gray}{1.3} & \textcolor{gray}{4.6} & \textcolor{gray}{2.8} & \textcolor{gray}{1.1} \\
\midrule
Mean of Baselines & 
92.0 & 8.5 & 3.1 & 92.3 & \textbf{5.3} & \textbf{2.2} & 
86.1 & 14.5 & 5.4 & \textbf{87.7} & \textbf{7.1} & \textbf{3.0} & 
88.5 & 12.7 & 4.9 & \textbf{89.7} & \textbf{6.7} & \textbf{2.8} \\
& \textcolor{gray}{4.7} & \textcolor{gray}{13.9} & \textcolor{gray}{3.8} & \textcolor{gray}{4.4} & \textbf{\textcolor{gray}{2.5}} & \textbf{\textcolor{gray}{1.2}} &
\textcolor{gray}{7.3} & \textcolor{gray}{17.9} & \textcolor{gray}{6.1} & \textbf{\textcolor{gray}{5.4}} & \textbf{\textcolor{gray}{2.6}} & \textbf{\textcolor{gray}{1.4}} &
\textcolor{gray}{7.1} & \textcolor{gray}{18.6} & \textcolor{gray}{6.9} & \textbf{\textcolor{gray}{5.1}} & \textbf{\textcolor{gray}{2.9}} & \textbf{\textcolor{gray}{1.4}} \\
\bottomrule
\end{tabular}
\end{minipage}%
}
\label{tab:prostate}
\end{table*}

For qualitative analysis, we selected the three strongest-performing non-variational baselines for each OOD
dataset: MedSAM, VMUNet, and ResUNet++ for ACDC (OOD); DSTransUNet, VMUNet, and BASNet for ACMRI (OOD). As
shown in Figure~\ref{fig:myo_acdc_auto} and Figure~\ref{fig:myo_acmri_auto}, these baselines often exhibit
common OOD failure modes, such as disconnected blob-like structures, potentially arising from bias-field and
other artifacts introduced by different acquisition equipment.
We also analyzed all variational baselines. We found that PHISeg, a large variational model with 99.21M
parameters (Table~\ref{tab:fetal_head}), shows considerable robustness on both ID and OOD data, which we
attribute to its complex multi-scale architecture. Conversely, ProbUNet and HierProbUNet architectures were
more sensitive to the domain shift, resulting in segmentations with highly uncertain and erroneous
boundaries. However, ProbUNet and HierProbUNet performed reasonably well on the CAP (ID)
dataset. Figure~\ref{fig:myo_acdc_auto} and Figure~\ref{fig:myo_acmri_auto} demonstrate that our VarDeepPCA
not only restores segmentation maps to make them more consistent with the underlying anatomy,
but it also produces uncertainty maps that show significantly more localization of uncertainty (aligned
with the geometry of the anatomical object of interest) and lower overall uncertainty compared to the
variational baselines.

{\bf Neuroretinal Rim Segmentation.}
We trained all models on the MAGRABI (ID) dataset, and evaluated their performance on the MAGRABI (ID) test,
G1020 (OOD), and ORIGA (OOD) datasets.
On the ID test set, most baselines yielded competitive HD95 scores (Table~\ref{tab:neuroretinal_rim}).
Exceptions included ResUNet++ that is likely constrained by its limited architectural capacity, and SegAN that
likely suffered from the known instability of adversarial optimization.
On the OOD datasets, we observed a substantial drop in DSC values. Notably, this DSC degradation was uniform
across most baselines. However, the boundary-based metrics (HD95 and ASD) showed different amounts of
performance degradations across baselines.
In this context, our VarDeepPCA helped improve the baselines segmentation maps, leading to a mean HD95 $\leq$9
on G1020 and $\leq$10 on ORIGA, significantly minimizing boundary errors compared to the baselines.
For this application, we configured the VarDeepPCA architecture with a latent dimension of $K = 8$ (refer
Section~\ref{subsec:sen} for sensitivity analysis).
Indeed, for the OOD datasets of G1020 and ORIGA, the mean HD95 values after employing the VarDeepPCA
plugin reduce from an average (across all methods; before VarDeepPCA) within 12.6-14.1 to an average (across
all methods; after VarDeepPCA) within 8.1-8.8, which is far more clinically acceptable as per our analysis
in Section~\ref{sec:UpperBoundClinUtil}.
We also present the quantitative results for the uncertainty calibration metrics, i.e., the NCC, US, and
TACE scores in Table~\ref{tab:fundus_calibration}. For the PHISeg, ProbUNet, and HPUNet baselines,
incorporating the VarDeepPCA plugin leads to a significant improvement of these scores in both ID and OOD
datasets.

\begin{figure}[!t]
\threeAcrossLabelsWidth{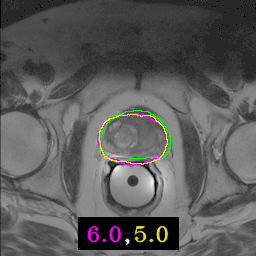}
{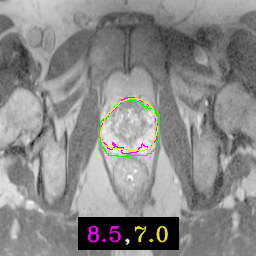}
{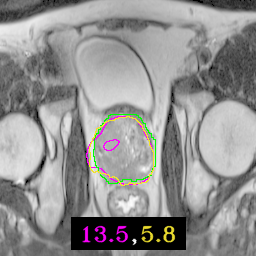}
{\scs{\bf(a)}~VMUNet}{\scs{\bf(b)}~SegAN}{\scs {\bf(c)}~SegCNN+DAE+TTA+Atlas}{0.48}
\threeAcrossLabelsWidth{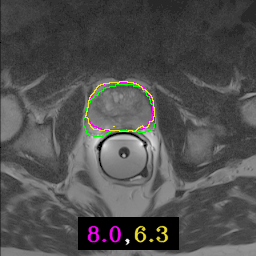}
{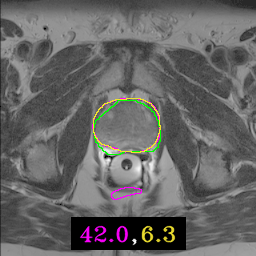}
{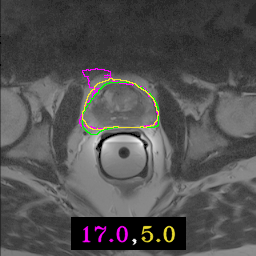}
{\scs{\bf(d)}~PHISeg}{\scs{\bf(e)}~ProbUNet}{\scs{\bf(f)}~HierProbUNet}{0.48}
\threeAcrossLabelsWidth{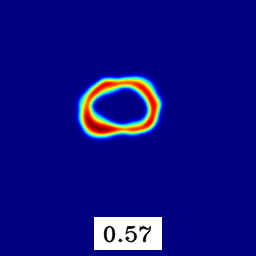}
{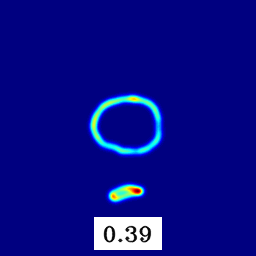}
{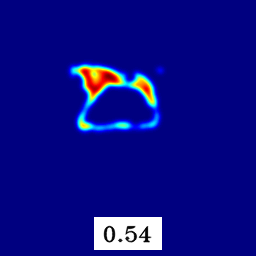}
{\scs{\bf(g)}~Unc. for (d)~PHISeg}{\scs{\bf(h)}~Unc. for (e)~ProbUNet}{\scs{\bf(i)}~Unc.  for (f)~HierProbUNet}{0.48}
\threeAcrossLabelsWidth{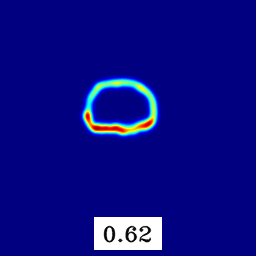}
{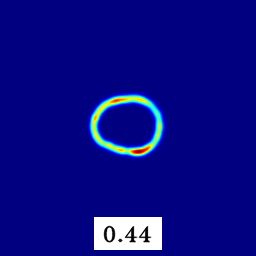}
{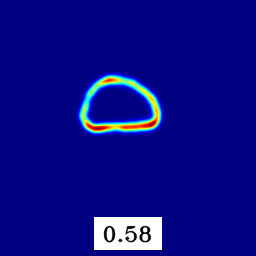}
{\scs{\bf(j)}~Unc.  for (d)+VarDeepPCA}{\scs{\bf(k)}~Unc.  for (e)+VarDeepPCA}{\scs{\bf(l)}~Unc.  for (f)+VarDeepPCA}{0.48}
\caption
{
{\bf Results--Qualitative: Prostate Segmentation Restoration on HK+I2CVB (OOD) data.}
{\bf (a)--(c)}~Results on images for the best non-variational baselines. 
{\bf (d)--(f)}~Results on images for the variational baselines.
{\bf (g)--(i)}~Uncertainty maps produced using variational baselines.
{\bf (j)--(l)}~Uncertainty maps produced using VarDeepPCA when plugged into the associated baselines
(d)--(f).
Color scheme in (a)--(f):
\textcolor{prostate_maroon}{Baseline};
\textcolor{prostate_yellow}{Baseline+VarDeepPCA (Ours)};
\textcolor{prostate_green}{Ground Truth}.
HD95 numbers in (a)--(f) indicate that the examples were representative of the test set, because the HD95
values were close to the mean HD95 reported in Table~\ref{tab:prostate}.
NCC numbers in (g)--(l) indicate that the examples were representative of the test set, because the NCC values were close to the mean of the NCC reported in Table~\ref{tab:prostate_calibration}.
}
\label{fig:prostate_hki2cvb_auto}
\end{figure}

\begin{figure}[!t]
\threeAcrossLabelsWidth{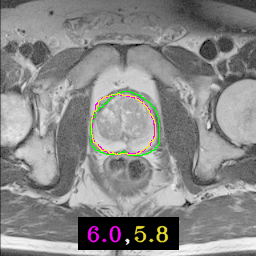}
{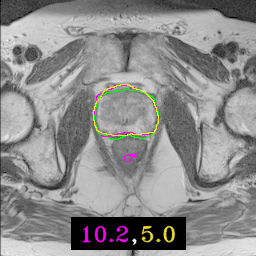}
{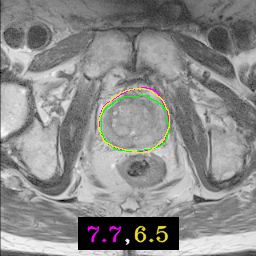}
{\scs{\bf(a)}~VMUNet}{\scs{\bf(b)}~SegAN}{\scs {\bf(c)}~SegCNN+DAE+TTA+Atlas}{0.48}
\threeAcrossLabelsWidth{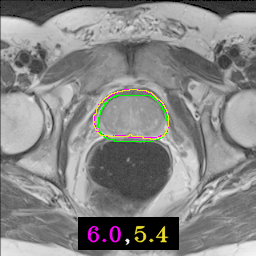}
{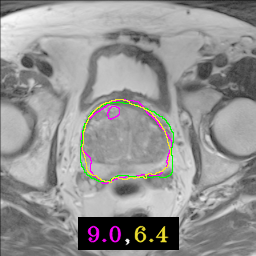}
{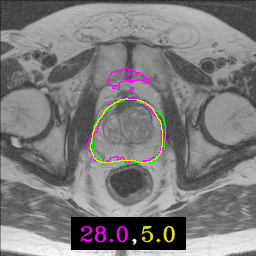}
{\scs{\bf(d)}~PHISeg}{\scs{\bf(e)}~ProbUNet}{\scs{\bf(f)}~HierProbUNet}{0.48}
\threeAcrossLabelsWidth{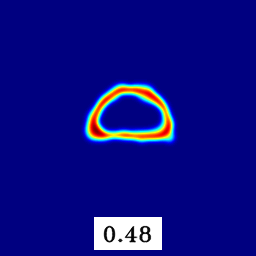}
{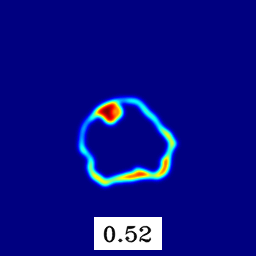}
{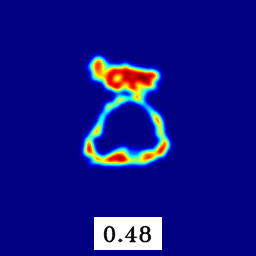}
{\scs{\bf(g)}~Unc. for (d)~PHISeg}{\scs{\bf(h)}~Unc. for (e)~ProbUNet}{\scs{\bf(i)}~Unc. for (f)~HierProbUNet}{0.48}
\threeAcrossLabelsWidth{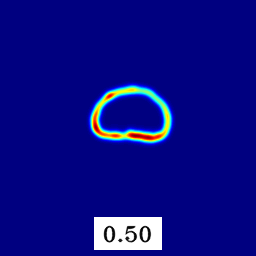}
{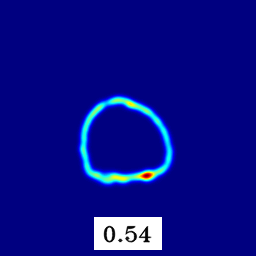}
{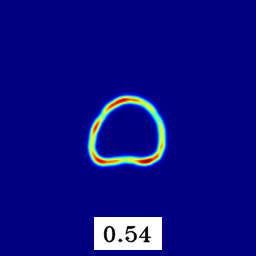}
{\scs{\bf(j)}~Unc. for (d)+VarDeepPCA}{\scs{\bf(k)}~Unc. for (e)+VarDeepPCA}{\scs{\bf(l)}~Unc. for (f)+VarDeepPCA}{0.48}
\caption
{
{\bf Results--Qualitative: Prostate Segmentation Restoration on RUNMC+UCL (OOD) data. }
{\bf (a)--(c)}~Results on images for the best non-variational baselines.
{\bf (d)--(f)}~Results on images for the variational baselines. 
{\bf (g)--(i)}~Uncertainty maps produced using variational baselines.
{\bf (j)--(l)}~Uncertainty maps produced using VarDeepPCA when plugged into the associated baselines
(d)--(f).
Color scheme in (a)--(f):
\textcolor{prostate_maroon}{Baseline};
\textcolor{prostate_yellow}{Baseline+VarDeepPCA (Ours)};
\textcolor{prostate_green}{Ground Truth}.
HD95 numbers in (a)--(f) indicate that the examples were representative of the test set, because the HD95
values were close to the mean HD95 reported in Table~\ref{tab:prostate}.
NCC numbers in (g)--(l) indicate that the examples were representative of the test set, because the NCC values were close to the mean of the NCC reported in Table~\ref{tab:prostate_calibration}.
}
\label{fig:prostate_runmcucl_auto}
\end{figure}

We compare with the strongest non-variational baselines for the G1020 (OOD) as shown in
Figure~\ref{fig:neurorim_g1020_auto}, and the ORIGA (OOD) dataset as shown in
Figure~\ref{fig:neurorim_origa_auto}; these were AttnUNet, MedSegDiff, and MedSAM (recall that MedSAM used
oracle-prompting that is actually unfair to other methods).
For variational models, PHISeg again exhibited superior robustness compared to ProbUNet and HierProbUNet.
The figures also show that the per-pixel uncertainty maps produced by our VarDeepPCA method are much more
aligned with the object anatomy as compared to those produced by the baselines.
%
%

{\bf Prostate Segmentation.}
Table~\ref{tab:prostate} shows that the performance of a few large DNN models (SegAN, VMUNet, PHISeg) does not
degrade much on OOD images with respect to the HD95 metric.
Our VarDeepPCA plugin continues to deliver consistent improvements, and reduces the mean HD95 values for all
baselines. In cases of severe degradation where baselines had high variance in HD95 values, e.g., MedSegDiff,
our method also reduced the standard deviation, improving the stability of the results.
We use a latent dimension of $K = 8$ for VarDeepPCA, as justified in our sensitivity analysis in
Section~\ref{subsec:sen}.
Indeed, for the OOD datasets, the mean HD95 values after employing the VarDeepPCA plugin reduce from an
average (across all methods; before VarDeepPCA) within 12.7-14.5 to an average (across all methods; after
VarDeepPCA) within 6.7-7.1, which is far more clinically acceptable as per our analysis in
Section~\ref{sec:UpperBoundClinUtil}.
We see minor improvements in the uncertainty calibration metrics from Table~\ref{tab:prostate_calibration}
for PHISeg, while there are major improvements for ProbUNet and HPUNet.

\begin{table*}[!t]
\caption{ 
{\bf Results--Quantitative: Fetal Head Segmentation.}
All models were trained on HC18 (ID) and evaluated on FetalPlanes (OOD).
For each method-dataset combination, we report mean (top row) and standard deviation (bottom row, in
\textcolor{gray}{gray}) for DSC($\uparrow$), HD95($\downarrow$), and ASD($\downarrow$).
Augmenting each baseline with our VarDeepPCA consistently improves performance.
{\bf Bold-font} values in the columns indicate a statistically significant improvement of
the Baseline+VarDeepPCA method over the underlying Baseline method, using a one-tailed paired-sample
t-test (p $<$ 0.05).
} \centering
\resizebox{0.99\textwidth}{!}{%
\begin{minipage}{\textwidth}
\centering
\footnotesize
\renewcommand{\arraystretch}{0.95}
\setlength{\tabcolsep}{2.5pt}
\begin{tabular}{@{}l| *{6}{c} @{\hspace{8pt}}| *{6}{c}@{}}     
\toprule
\multirow{3}{*}{\textbf{Models}} & 
\multicolumn{6}{c}{\textbf{HC18 (ID)}} & 
\multicolumn{6}{c}{\textbf{FetalPlanes (OOD)}} \\
\cmidrule(lr){2-7} \cmidrule(lr){8-13}
& \multicolumn{3}{c}{Baseline} & \multicolumn{3}{c}{\makecell{Baseline + \\ \textbf{VarDeepPCA}}} & 
\multicolumn{3}{c}{Baseline} & \multicolumn{3}{c}{\makecell{Baseline + \\ \textbf{VarDeepPCA}}} \\
\cmidrule(lr){2-4} \cmidrule(lr){5-7} \cmidrule(lr){8-10} \cmidrule(lr){11-13}
& DSC & HD95 & ASD & DSC & HD95 & ASD & 
DSC & HD95 & ASD & DSC & HD95 & ASD \\
\midrule
UNet & 
84.5 & 48.1 & 15.0 & \textbf{96.9} & \textbf{5.2} & \textbf{2.2} & 
89.4 & 38.5 & 12.1 & \textbf{96.5} & \textbf{5.2} & \textbf{2.4} \\
&
\textcolor{gray}{12.9} & \textcolor{gray}{10.9} & \textcolor{gray}{3.9} & \textbf{\textcolor{gray}{0.9}} & \textbf{\textcolor{gray}{1.4}} & \textbf{\textcolor{gray}{0.6}} &
\textcolor{gray}{8.6} & \textcolor{gray}{18.6} & \textcolor{gray}{5.8} & \textbf{\textcolor{gray}{1.2}} & \textbf{\textcolor{gray}{1.4}} & \textbf{\textcolor{gray}{0.8}} \\
\addlinespace
AttnUNet & 
96.2 & 8.4 & 2.6 & \textbf{97.0} & \textbf{5.2} & \textbf{2.1} & 
93.2 & 13.6 & 4.2 & \textbf{96.6} & \textbf{5.1} & \textbf{2.3} \\
&
\textcolor{gray}{6.6} & \textcolor{gray}{13.0} & \textcolor{gray}{3.4} & \textbf{\textcolor{gray}{0.9}} & \textbf{\textcolor{gray}{1.3}} & \textbf{\textcolor{gray}{0.6}} &
\textcolor{gray}{8.8} & \textcolor{gray}{16.0} & \textcolor{gray}{3.9} & \textbf{\textcolor{gray}{1.2}} & \textbf{\textcolor{gray}{1.4}} & \textbf{\textcolor{gray}{0.8}} \\
\addlinespace
ResUNet++ & 
85.8 & 41.1 & 16.2 & \textbf{95.8} & \textbf{6.2} & \textbf{3.0} & 
84.9 & 42.4 & 15.6 & \textbf{96.0} & \textbf{5.7} & \textbf{2.8} \\
&
\textcolor{gray}{7.4} & \textcolor{gray}{15.8} & \textcolor{gray}{7.1} & \textbf{\textcolor{gray}{1.0}} & \textbf{\textcolor{gray}{1.1}} & \textbf{\textcolor{gray}{0.8}} &
\textcolor{gray}{7.3} & \textcolor{gray}{14.1} & \textcolor{gray}{5.7} & \textbf{\textcolor{gray}{1.2}} & \textbf{\textcolor{gray}{1.3}} & \textbf{\textcolor{gray}{0.9}} \\
\addlinespace
DeepLabV3+ & 
95.7 & 10.5 & 3.5 & \textbf{96.8} & \textbf{5.4} & \textbf{2.3} & 
94.1 & 14.3 & 5.0 & \textbf{96.5} & \textbf{5.5} & \textbf{2.5} \\
&
\textcolor{gray}{3.0} & \textcolor{gray}{6.7} & \textcolor{gray}{2.3} & \textbf{\textcolor{gray}{1.0}} & \textbf{\textcolor{gray}{1.3}} & \textbf{\textcolor{gray}{0.7}} &
\textcolor{gray}{2.4} & \textcolor{gray}{6.6} & \textcolor{gray}{2.2} & \textbf{\textcolor{gray}{1.1}} & \textbf{\textcolor{gray}{1.3}} & \textbf{\textcolor{gray}{0.7}} \\
\addlinespace
BASNet & 
96.3 & 5.9 & 2.1 & 96.8 & \textbf{5.3} & \textbf{1.9} & 
96.8 & 5.9 & 2.4 & 96.8 & \textbf{5.0} & \textbf{2.3} \\
&
\textcolor{gray}{1.5} & \textcolor{gray}{8.1} & \textcolor{gray}{2.3} & \textcolor{gray}{1.0} & \textbf{\textcolor{gray}{1.4}} & \textbf{\textcolor{gray}{0.6}} &
\textcolor{gray}{1.2} & \textcolor{gray}{4.0} & \textcolor{gray}{1.1} & \textcolor{gray}{1.1} & \textbf{\textcolor{gray}{1.4}} & \textbf{\textcolor{gray}{0.8}} \\
\addlinespace
SegAN & 
96.1 & 6.6 & 2.8 & 96.8 & \textbf{5.2} & \textbf{2.2} & 
96.1 & 7.2 & 3.3 & \textbf{96.5} & \textbf{5.2} & \textbf{2.4} \\
&
\textcolor{gray}{2.3} & \textcolor{gray}{4.7} & \textcolor{gray}{1.8} & \textcolor{gray}{1.0} & \textbf{\textcolor{gray}{1.4}} & \textbf{\textcolor{gray}{0.6}} &
\textcolor{gray}{1.5} & \textcolor{gray}{3.6} & \textcolor{gray}{1.1} & \textbf{\textcolor{gray}{1.1}} & \textbf{\textcolor{gray}{1.4}} & \textbf{\textcolor{gray}{0.7}} \\
\addlinespace
MedSegDiff & 
96.2 & 5.3 & 2.7 & 97.0 & \textbf{4.9} & \textbf{2.1} & 
96.0 & 7.2 & 2.7 & \textbf{96.7} & \textbf{5.0} & \textbf{2.3} \\
&
\textcolor{gray}{1.1} & \textcolor{gray}{5.6} & \textcolor{gray}{1.1} & \textcolor{gray}{0.9} & \textbf{\textcolor{gray}{1.4}} & \textbf{\textcolor{gray}{0.6}} &
\textcolor{gray}{7.4} & \textcolor{gray}{7.7} & \textcolor{gray}{2.7} & \textbf{\textcolor{gray}{1.0}} & \textbf{\textcolor{gray}{1.4}} & \textbf{\textcolor{gray}{0.7}} \\
\addlinespace
DSTransUNet & 
95.5 & 7.9 & 2.5 & \textbf{96.4} & \textbf{5.6} & 2.5 & 
95.6 & 8.2 & 3.1 & \textbf{96.3} & \textbf{5.4} & \textbf{2.5} \\
&
\textcolor{gray}{1.4} & \textcolor{gray}{3.7} & \textcolor{gray}{1.0} & \textbf{\textcolor{gray}{1.0}} & \textbf{\textcolor{gray}{1.3}} & \textcolor{gray}{0.6} &
\textcolor{gray}{1.7} & \textcolor{gray}{3.6} & \textcolor{gray}{1.1} & \textbf{\textcolor{gray}{1.3}} & \textbf{\textcolor{gray}{1.4}} & \textbf{\textcolor{gray}{0.8}} \\
\addlinespace
VMUNet & 
96.0 & 6.4 & 2.4 & \textbf{96.9} & \textbf{5.1} & \textbf{2.2} & 
95.6 & 8.7 & 3.4 & \textbf{96.6} & \textbf{5.1} & \textbf{2.3} \\
&
\textcolor{gray}{2.4} & \textcolor{gray}{6.7} & \textcolor{gray}{2.2} & \textbf{\textcolor{gray}{0.9}} & \textbf{\textcolor{gray}{1.4}} & \textbf{\textcolor{gray}{0.6}} &
\textcolor{gray}{2.3} & \textcolor{gray}{6.1} & \textcolor{gray}{2.0} & \textbf{\textcolor{gray}{1.1}} & \textbf{\textcolor{gray}{1.4}} & \textbf{\textcolor{gray}{0.7}} \\
\addlinespace
MedSAM & 
92.3 & 13.9 & 6.4 & 92.6 & \textbf{10.8} & \textbf{5.7} & 
91.5 & 13.3 & 6.6 & \textbf{91.8} & \textbf{10.5} & \textbf{6.1} \\
&
\textcolor{gray}{4.5} & \textcolor{gray}{7.8} & \textcolor{gray}{4.0} & \textcolor{gray}{4.3} & \textbf{\textcolor{gray}{5.6}} & \textbf{\textcolor{gray}{3.6}} &
\textcolor{gray}{4.6} & \textcolor{gray}{7.2} & \textcolor{gray}{3.9} & \textbf{\textcolor{gray}{4.8}} & \textbf{\textcolor{gray}{5.6}} & \textbf{\textcolor{gray}{3.9}} \\
\addlinespace
PHISeg & 
96.0 & 5.3 & 2.1 & \textbf{96.9} & \textbf{5.3} & \textbf{1.7} & 
95.6 & 7.2 & 2.9 & \textbf{96.4} & \textbf{5.3} & \textbf{2.4} \\
&
\textcolor{gray}{1.9} & \textcolor{gray}{3.6} & \textcolor{gray}{1.3} & \textbf{\textcolor{gray}{0.9}} & \textbf{\textcolor{gray}{1.3}} & \textbf{\textcolor{gray}{0.6}} &
\textcolor{gray}{2.3} & \textcolor{gray}{3.8} & \textcolor{gray}{1.3} & \textbf{\textcolor{gray}{1.1}} & \textbf{\textcolor{gray}{1.4}} & \textbf{\textcolor{gray}{0.7}} \\
\addlinespace
ProbUNet & 
95.7 & 18.6 & 5.0 & 95.9 & \textbf{6.5} & \textbf{3.3} & 
93.9 & 14.1 & 4.9 & \textbf{94.9} & \textbf{6.6} & \textbf{3.3} \\
&
\textcolor{gray}{2.6} & \textcolor{gray}{15.3} & \textcolor{gray}{3.3} & \textcolor{gray}{1.3} & \textbf{\textcolor{gray}{1.2}} & \textbf{\textcolor{gray}{0.8}} &
\textcolor{gray}{4.7} & \textcolor{gray}{11.2} & \textcolor{gray}{3.4} & \textbf{\textcolor{gray}{1.1}} & \textbf{\textcolor{gray}{0.9}} & \textbf{\textcolor{gray}{0.7}} \\
\addlinespace
HierProbUNet & 
96.5 & 11.5 & 3.8 & \textbf{96.8} & \textbf{5.3} & \textbf{2.2} & 
94.5 & 14.1 & 4.9 & \textbf{96.4} & \textbf{5.3} & \textbf{2.4} \\
&
\textcolor{gray}{2.0} & \textcolor{gray}{7.5} & \textcolor{gray}{2.3} & \textbf{\textcolor{gray}{1.0}} & \textbf{\textcolor{gray}{1.4}} & \textbf{\textcolor{gray}{0.7}} &
\textcolor{gray}{2.8} & \textcolor{gray}{7.7} & \textcolor{gray}{2.2} & \textbf{\textcolor{gray}{1.3}} & \textbf{\textcolor{gray}{1.5}} & \textbf{\textcolor{gray}{0.8}} \\
\addlinespace
SegCNN+DAE & 
96.1 & 8.9 & 2.7 & 96.2 & \textbf{5.3} & \textbf{2.3} & 
95.3 & 8.5 & 3.1 & \textbf{96.1} & \textbf{5.7} & \textbf{2.6} \\
+TTA & \textcolor{gray}{3.1} & \textcolor{gray}{10.6} & \textcolor{gray}{3.3} & \textcolor{gray}{1.5} & \textbf{\textcolor{gray}{1.8}} & \textbf{\textcolor{gray}{0.9}} &
\textcolor{gray}{3.1} & \textcolor{gray}{6.9} & \textcolor{gray}{2.3} & \textbf{\textcolor{gray}{1.8}} & \textbf{\textcolor{gray}{1.6}} & \textbf{\textcolor{gray}{1.1}} \\
\addlinespace
SegCNN+DAE & 
96.9 & 8.6 & 2.5 & 96.9 & \textbf{5.1} & \textbf{2.2} & 
95.8 & 8.0 & 2.8 & \textbf{96.4} & \textbf{5.2} & \textbf{2.4} \\
+TTA+Atlas & \textcolor{gray}{2.9} & \textcolor{gray}{10.3} & \textcolor{gray}{2.9} & \textcolor{gray}{1.1} & \textbf{\textcolor{gray}{1.4}} & \textbf{\textcolor{gray}{0.7}} &
\textcolor{gray}{2.8} & \textcolor{gray}{6.2} & \textcolor{gray}{1.7} & \textbf{\textcolor{gray}{1.3}} & \textbf{\textcolor{gray}{1.4}} & \textbf{\textcolor{gray}{0.8}} \\
\midrule
Mean of Baselines & 
94.8 & 11.5 & 4.1 & \textbf{96.4} & \textbf{5.9} & \textbf{2.5} & 
94.2 & 12.7 & 4.7 & \textbf{96.0} & \textbf{5.8} & \textbf{2.8} \\
& \textcolor{gray}{5.7} & \textcolor{gray}{13.6} & \textcolor{gray}{4.4} & \textbf{\textcolor{gray}{2.2}} & \textbf{\textcolor{gray}{2.9}} & \textbf{\textcolor{gray}{1.8}} &
\textcolor{gray}{5.5} & \textcolor{gray}{12.9} & \textcolor{gray}{4.3} & \textbf{\textcolor{gray}{2.4}} & \textbf{\textcolor{gray}{2.8}} & \textbf{\textcolor{gray}{1.9}} \\
\bottomrule
\end{tabular}
\end{minipage}%
}
\label{tab:fetal_head}
\end{table*}

\begin{table*}[!t]
\caption
{
{\bf Results--Quantitative: Prostate -- Measuring Calibration between Per-Pixel Segmentation
Uncertainty and Per-Pixel Segmentation Error.}
All models were trained on BIDMC+BMC (ID), and evaluated on HK+I2CVB and RUNMC+UCL (both OOD).
For each method-dataset combination, we report the mean (top row) and standard deviation (bottom row; in
\textcolor{gray}{gray}) for NCC~($\uparrow$), US~($\uparrow$), and TACE~($\downarrow$) metrics.
Augmenting each baseline with our VarDeepPCA shows better calibration.
{\bf Bold-font} values in the columns indicate a statistically significant improvement of
the Baseline+VarDeepPCA method over the underlying Baseline method, using a one-tailed paired-sample
t-test (p $<$ 0.05).
}
\centering
\resizebox{0.96\textwidth}{!}{%
\begin{minipage}{\textwidth}
\centering
\footnotesize
\renewcommand{\arraystretch}{0.95}
\setlength{\tabcolsep}{2.5pt}
\begin{tabular}{@{}l| *{6}{c} @{\hspace{8pt}}| *{6}{c} @{\hspace{8pt}}| *{6}{c}@{}}        
\toprule
\multirow{3}{*}{\textbf{Models}} & 
\multicolumn{6}{c}{\textbf{BIDMC+BMC (ID)}} & 
\multicolumn{6}{c}{\textbf{HK+I2CVB (OOD)}} & 
\multicolumn{6}{c}{\textbf{RUNMC+UCL (OOD)}} \\

\cmidrule(lr){2-7} \cmidrule(lr){8-13} \cmidrule(lr){14-19} 
& \multicolumn{3}{c}{Baseline} & \multicolumn{3}{c}{Baseline +} &
\multicolumn{3}{c}{Baseline} & \multicolumn{3}{c}{Baseline +} &
\multicolumn{3}{c}{Baseline} & \multicolumn{3}{c}{Baseline +} \\

& \multicolumn{3}{c}{} & \multicolumn{3}{c}{\bf VarDeepPCA} &
\multicolumn{3}{c}{} & \multicolumn{3}{c}{\bf VarDeepPCA} &
\multicolumn{3}{c}{} & \multicolumn{3}{c}{\bf VarDeepPCA} \\
\cmidrule(lr){2-4} \cmidrule(lr){5-7} \cmidrule(lr){8-10} \cmidrule(lr){11-13} \cmidrule(lr){14-16} \cmidrule(lr){17-19}
& NCC & US & TACE & NCC & US & TACE & 
NCC & US & TACE & NCC & US & TACE & 
NCC & US & TACE & NCC & US & TACE \\
\midrule
PHISeg & 
0.66 & 0.76 & 0.39 & \textbf{0.69} & \textbf{0.82} & \textbf{0.12} & 
0.57 & 0.63 & 0.47 & \textbf{0.63} & \textbf{0.69} & \textbf{0.14} & 
0.45 & 0.66 & 0.53 & \textbf{0.52} & \textbf{0.73} & \textbf{0.09} \\
& \textcolor{gray}{0.14} & \textcolor{gray}{0.04} & \textcolor{gray}{0.26} & \textbf{\textcolor{gray}{0.10}} & \textbf{\textcolor{gray}{0.04}} & \textbf{\textcolor{gray}{0.24}} &
\textcolor{gray}{0.11} & \textcolor{gray}{0.03} & \textcolor{gray}{0.23} & \textbf{\textcolor{gray}{0.08}} & \textbf{\textcolor{gray}{0.04}} & \textbf{\textcolor{gray}{0.24}} &
\textcolor{gray}{0.05} & \textcolor{gray}{0.04} & \textcolor{gray}{0.17} & \textbf{\textcolor{gray}{0.03}} & \textbf{\textcolor{gray}{0.04}} & \textbf{\textcolor{gray}{0.12}} \\
\addlinespace
ProbUNet & 
0.45 & 0.69 & 0.44 & \textbf{0.54} & \textbf{0.73} & \textbf{0.14} & 
0.40 & 0.59 & 0.50 & \textbf{0.45} & \textbf{0.65} & \textbf{0.12} & 
0.51 & 0.68 & 0.46 & \textbf{0.55} & \textbf{0.74} & \textbf{0.08} \\
& \textcolor{gray}{0.08} & \textcolor{gray}{0.03} & \textcolor{gray}{0.31} & \textbf{\textcolor{gray}{0.08}} & \textbf{\textcolor{gray}{0.04}} & \textbf{\textcolor{gray}{0.25}} &
\textcolor{gray}{0.06} & \textcolor{gray}{0.03} & \textcolor{gray}{0.32} & \textbf{\textcolor{gray}{0.06}} & \textbf{\textcolor{gray}{0.03}} & \textbf{\textcolor{gray}{0.23}} &
\textcolor{gray}{0.03} & \textcolor{gray}{0.03} & \textcolor{gray}{0.25} & \textbf{\textcolor{gray}{0.02}} & \textbf{\textcolor{gray}{0.03}} & \textbf{\textcolor{gray}{0.12}} \\
\addlinespace
HPUNet & 
0.51 & 0.69 & 0.43 & \textbf{0.61} & \textbf{0.79} & \textbf{0.16} & 
0.54 & 0.64 & 0.65 & \textbf{0.59} & \textbf{0.75} & \textbf{0.19} & 
0.47 & 0.65 & 0.46 & \textbf{0.56} & \textbf{0.74} & \textbf{0.12} \\
& \textcolor{gray}{0.09} & \textcolor{gray}{0.03} & \textcolor{gray}{0.29} & \textbf{\textcolor{gray}{0.08}} & \textbf{\textcolor{gray}{0.03}} & \textbf{\textcolor{gray}{0.24}} &
\textcolor{gray}{0.07} & \textcolor{gray}{0.03} & \textcolor{gray}{0.25} & \textbf{\textcolor{gray}{0.06}} & \textbf{\textcolor{gray}{0.02}} & \textbf{\textcolor{gray}{0.18}} &
\textcolor{gray}{0.04} & \textcolor{gray}{0.04} & \textcolor{gray}{0.22} & \textbf{\textcolor{gray}{0.03}} & \textbf{\textcolor{gray}{0.03}} & \textbf{\textcolor{gray}{0.15}} \\
\addlinespace
\bottomrule
\end{tabular}
\end{minipage}%
}
\label{tab:prostate_calibration}
\end{table*}

\begin{table*}[!b]
\caption
{
{\bf Results--Quantitative: Fetal Head -- Measuring Calibration between Per-Pixel Segmentation
Uncertainty and Per-Pixel Segmentation Error.}
All models were trained on HC18 (ID), and evaluated on FetalPlanes (OOD).
For each method-dataset combination, we report the mean (top row) and standard deviation (bottom row; in
\textcolor{gray}{gray}) for NCC~($\uparrow$), US~($\uparrow$), and TACE~($\downarrow$) metrics.
Augmenting each baseline with our VarDeepPCA shows better calibration.
{\bf Bold-font} values in the columns indicate a statistically significant improvement of
the Baseline+VarDeepPCA method over the underlying Baseline method, using a one-tailed paired-sample
t-test (p $<$ 0.05).
}
\centering
\resizebox{0.96\textwidth}{!}{%
\begin{minipage}{\textwidth}
\centering
\footnotesize
\renewcommand{\arraystretch}{0.95}
\setlength{\tabcolsep}{2.5pt}
\begin{tabular}{@{}l| *{6}{c} @{\hspace{8pt}}| *{6}{c}@{}}        
\toprule
\multirow{3}{*}{\textbf{Models}} & 
\multicolumn{6}{c}{\textbf{HC18 (ID)}} & 
\multicolumn{6}{c}{\textbf{FetalPlanes (OOD)}} \\

\cmidrule(lr){2-7} \cmidrule(lr){8-13}
& \multicolumn{3}{c}{Baseline} & \multicolumn{3}{c}{Baseline +} &
\multicolumn{3}{c}{Baseline} & \multicolumn{3}{c}{Baseline +} \\

& \multicolumn{3}{c}{} & \multicolumn{3}{c}{\bf VarDeepPCA} &
\multicolumn{3}{c}{} & \multicolumn{3}{c}{\bf VarDeepPCA} \\
\cmidrule(lr){2-4} \cmidrule(lr){5-7} \cmidrule(lr){8-10} \cmidrule(lr){11-13}
& NCC & US & TACE & NCC & US & TACE & 
NCC & US & TACE & NCC & US & TACE \\
\midrule
PHISeg & 
0.67 & 0.69 & 0.34 & \textbf{0.73} & \textbf{0.75} & \textbf{0.14} & 
0.49 & 0.71 & 0.45 & \textbf{0.55} & \textbf{0.78} & \textbf{0.08} \\
& \textcolor{gray}{0.06} & \textcolor{gray}{0.03} & \textcolor{gray}{0.04} & \textbf{\textcolor{gray}{0.03}} & \textbf{\textcolor{gray}{0.03}} & \textbf{\textcolor{gray}{0.03}} &
\textcolor{gray}{0.05} & \textcolor{gray}{0.02} & \textcolor{gray}{0.05} & \textbf{\textcolor{gray}{0.04}} & \textbf{\textcolor{gray}{0.00}} & \textbf{\textcolor{gray}{0.04}} \\
\addlinespace
ProbUNet & 
0.54 & 0.72 & 0.37 & \textbf{0.68} & \textbf{0.82} & \textbf{0.15} & 
0.34 & 0.65 & 0.44 & \textbf{0.45} & \textbf{0.75} & \textbf{0.11} \\
& \textcolor{gray}{0.05} & \textcolor{gray}{0.03} & \textcolor{gray}{0.08} & \textbf{\textcolor{gray}{0.03}} & \textbf{\textcolor{gray}{0.04}} & \textbf{\textcolor{gray}{0.04}} &
\textcolor{gray}{0.07} & \textcolor{gray}{0.02} & \textcolor{gray}{0.03} & \textbf{\textcolor{gray}{0.05}} & \textbf{\textcolor{gray}{0.01}} & \textbf{\textcolor{gray}{0.03}} \\
\addlinespace
HPUNet & 
0.46 & 0.71 & 0.41 & \textbf{0.54} & \textbf{0.77} & \textbf{0.25} & 
0.35 & 0.65 & 0.46 & \textbf{0.53} & \textbf{0.74} & \textbf{0.14} \\
& \textcolor{gray}{0.06} & \textcolor{gray}{0.02} & \textcolor{gray}{0.04} & \textbf{\textcolor{gray}{0.03}} & \textbf{\textcolor{gray}{0.03}} & \textbf{\textcolor{gray}{0.03}} &
\textcolor{gray}{0.06} & \textcolor{gray}{0.01} & \textcolor{gray}{0.04} & \textbf{\textcolor{gray}{0.05}} & \textbf{\textcolor{gray}{0.00}} & \textbf{\textcolor{gray}{0.03}} \\
\addlinespace
\bottomrule
\end{tabular}
\end{minipage}%
}
\label{tab:fetal_calibration}
\end{table*}

For qualitative analysis on OOD data, we selected the three best-performing non-variational baselines based on
their mean HD95 metrics:
(i)~for the HK+I2CVB dataset, these were VMUNet, SegAN, and DSTransUNet
(Figure~\ref{fig:prostate_hki2cvb_auto}); (ii)~for the RUNMC+UCL dataset, these were VMUNet, SegAN, and
DeepLabV3+ (Figure~\ref{fig:prostate_runmcucl_auto}).
Compared to myocardium segmentation and neuroretinal rim segmentation, VMUNet performed much better on the
prostate dataset with relatively smaller HD95 values. However, VMUNet continued to exhibit high variance of
the HD95 values. Our VarDeepPCA, when plugged into VMUNet, not only reduced the mean HD95 values but, more
importantly, reduced the variance of the HD95 values significantly.
The analysis of variational models revealed greater instability. While PHISeg's multi-scale architecture
provided some robustness to OOD images, ProbUNet and HierProbUNet showed severe degradation. HierProbUNet, for
instance, performed adequately on ID data but failed on both OOD datasets.
As seen in Figure~\ref{fig:prostate_hki2cvb_auto} and Figure~\ref{fig:prostate_runmcucl_auto}, the
uncertainty maps from these baselines often exhibited inconsistencies with the object's geometry, unlike the
uncertainty maps produced by VarDeepPCA. Our method successfully addressed both issues, i.e., it improved
the object segmentation maps and also improved the uncertainty maps.

\begin{figure}[!t]
\threeAcrossLabelsWidth{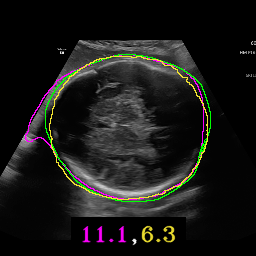}
{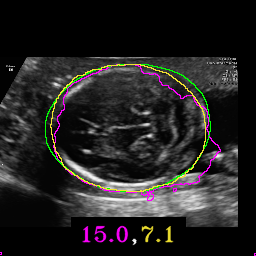}
{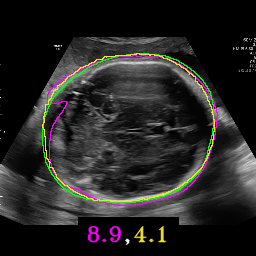}
{\scs{\bf(a)}~BASNet}{\scs{\bf(b)}~SegAN}{\scs{\bf(c)}~MedSegDiff}{0.48}
\threeAcrossLabelsWidth{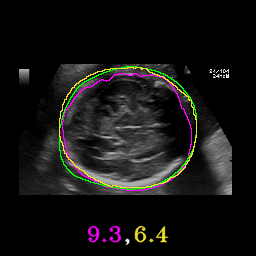}
{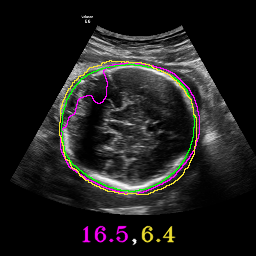}
{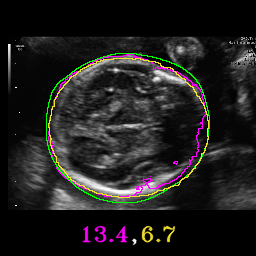}
{\scs{\bf(d)}~PHISeg}{\scs{\bf(e)}~ProbUNet}{\scs{\bf(f)}~HierProbUNet}{0.48}
\threeAcrossLabelsWidth{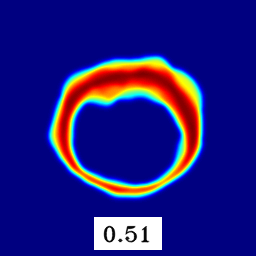}
{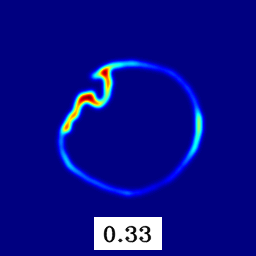}
{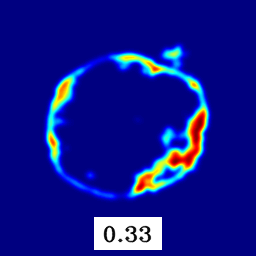}
{\scs{\bf(g)}~Unc. for (d)~PHISeg}{\scs{\bf(h)}~Unc. for (e)~ProbUNet}{\scs{\bf(i)}~Unc.  for (f)~HierProbUNet}{0.48}
\threeAcrossLabelsWidth{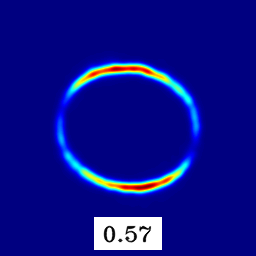}
{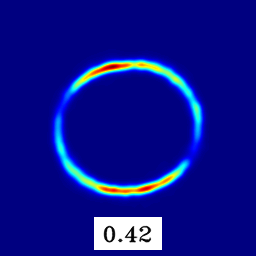}
{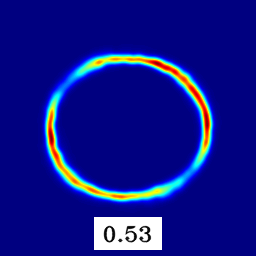}
{\scs{\bf(j)}~Unc. for (d)+VarDeepPCA}{\scs{\bf(k)}~Unc. for (e)+VarDeepPCA}{\scs{\bf(l)}~Unc. for (f)+VarDeepPCA}{0.48}
\caption
{
{\bf Results--Qualitative: Fetal Head Segmentation Restoration on FetalPlanes (OOD) data. }
{\bf (a)--(c)}~Results on images for the best non-variational baselines.
{\bf (d)--(f)}~Results on images for the variational baselines. 
{\bf (g)--(i)}~Uncertainty maps produced using variational baselines.
{\bf (j)--(l)}~Uncertainty maps produced using VarDeepPCA when plugged into the associated baselines
(d)--(f).
Color scheme in (a)--(f):
\textcolor{fetal_head_maroon}{Baseline};
\textcolor{fetal_head_yellow}{Baseline+VarDeepPCA (Ours)};
\textcolor{fetal_head_green}{Ground Truth}.
HD95 numbers in (a)--(f) indicate that the examples were representative of the test set, because the HD95
values were close to the mean HD95 reported in Table~\ref{tab:fetal_head}.
NCC numbers in (g)--(l) indicate that the examples were representative of the test set, because the NCC values were close to the mean of the NCC reported in Table~\ref{tab:fetal_calibration}.
}
\label{fig:fetalhead_fetalplanes_auto}
\end{figure}


{\bf Fetal Head Segmentation.}
Applying our VarDeepPCA framework yields consistent performance gains across both HC18 (ID) and FetalPlanes
(OOD) datasets, with the improvements being most pronounced in the HD95 metric. This demonstrates that
VarDeepPCA effectively corrects the boundary errors caused by noise and domain shift.
For these experiments, we use a latent dimension of $K = 3$ for VarDeepPCA, as justified in our sensitivity
analysis in Section~\ref{subsec:sen}.
We also note that diffusion-based models like MedSegDiff, and models with boundary-aware loss function such as
BASNet, outperformed the MedSAM foundational model (even when using oracle prompting) in this application.
Table~\ref{tab:fetal_head} shows that despite some baseline models (e.g., UNet and ResUNet++) achieving good
(85\%-90\%) DSC scores (overlap based), their HD95 values (boundary based) are poor (around 40), even on the
HC18 (ID) test set. This stems from the well-known limitation of DSC in segmenting large near-convex objects.
Many baseline models exhibiting poor mean HD95 scores also showed high standard deviations
(Table~\ref{tab:fetal_head}) in HD95, indicating inconsistent performance across images.
In contrast, our method not only achieves a lower (better) mean HD95 but also simultaneously reduces the
standard deviation, demonstrating higher accuracy and precision. In all cases, our method consistently
restores degraded segmentation maps towards valid anatomical geometries.
Indeed, for the OOD datasets, the mean HD95 values after employing the VarDeepPCA plugin reduce from an
average (across all methods; before VarDeepPCA) within 11.5-12.7 to an average (across all methods; after
VarDeepPCA) within 5.8-5.9, which is far more clinically acceptable as per our analysis in
Section~\ref{sec:UpperBoundClinUtil}.
From Table \ref{tab:fetal_calibration}, we see that the models ProbUNet and HPUNet are not well calibrated
and do not provide good uncertainty estimates, whereas, PHISeg is relatively well calibrated and provides
better uncertainty estimates.

For qualitative analysis on the FetalPlanes (OOD) dataset, we show examples from three best-performing
non-variational baselines: BASNet, SegAN, and MedSegDiff.
These baselines frequently produce severe mispredictions, with segmentation boundaries extending well beyond
the anatomical boundaries (Figure~\ref{fig:fetalhead_fetalplanes_auto}). In contrast, our method successfully
restores these degraded masks to anatomically valid geometries, much closer to the ground truth.
Regarding the variational models, PHISeg again performs better than ProbUNet and HierProbUNet on the OOD
dataset, probably because PHISeg leverages a more sophisticated larger backbone DNN than UNet.
Figure~\ref{fig:fetalhead_fetalplanes_auto} also shows that our method generates superior uncertainty maps;
the uncertainty is suitably localized and respects the geometry of the fetal head, unlike the more diffuse or
inaccurate uncertainty maps from the other variational baselines.

\begin{figure}[!t]
\centering
%
\twoAcrossLabelsHeight[0]{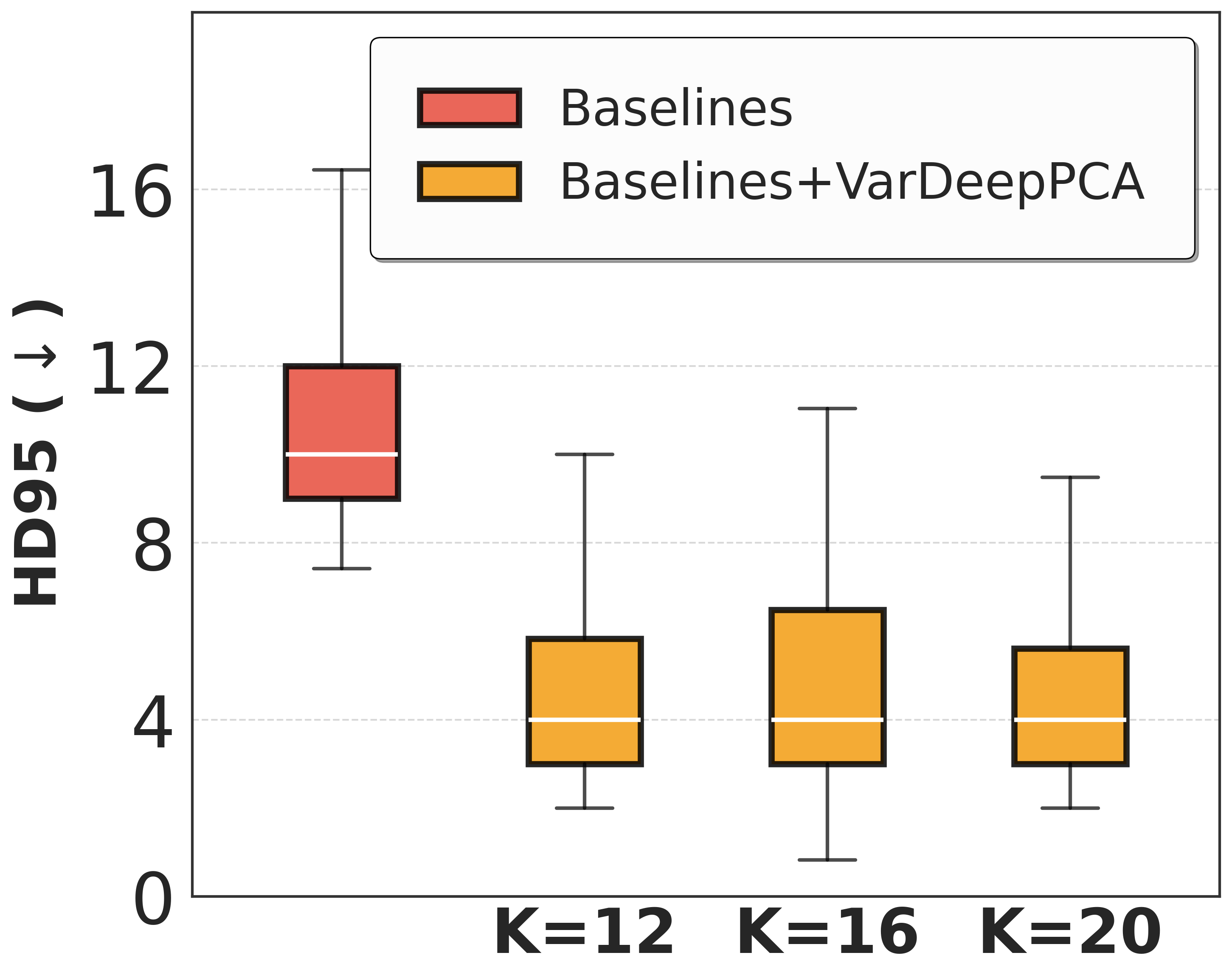}{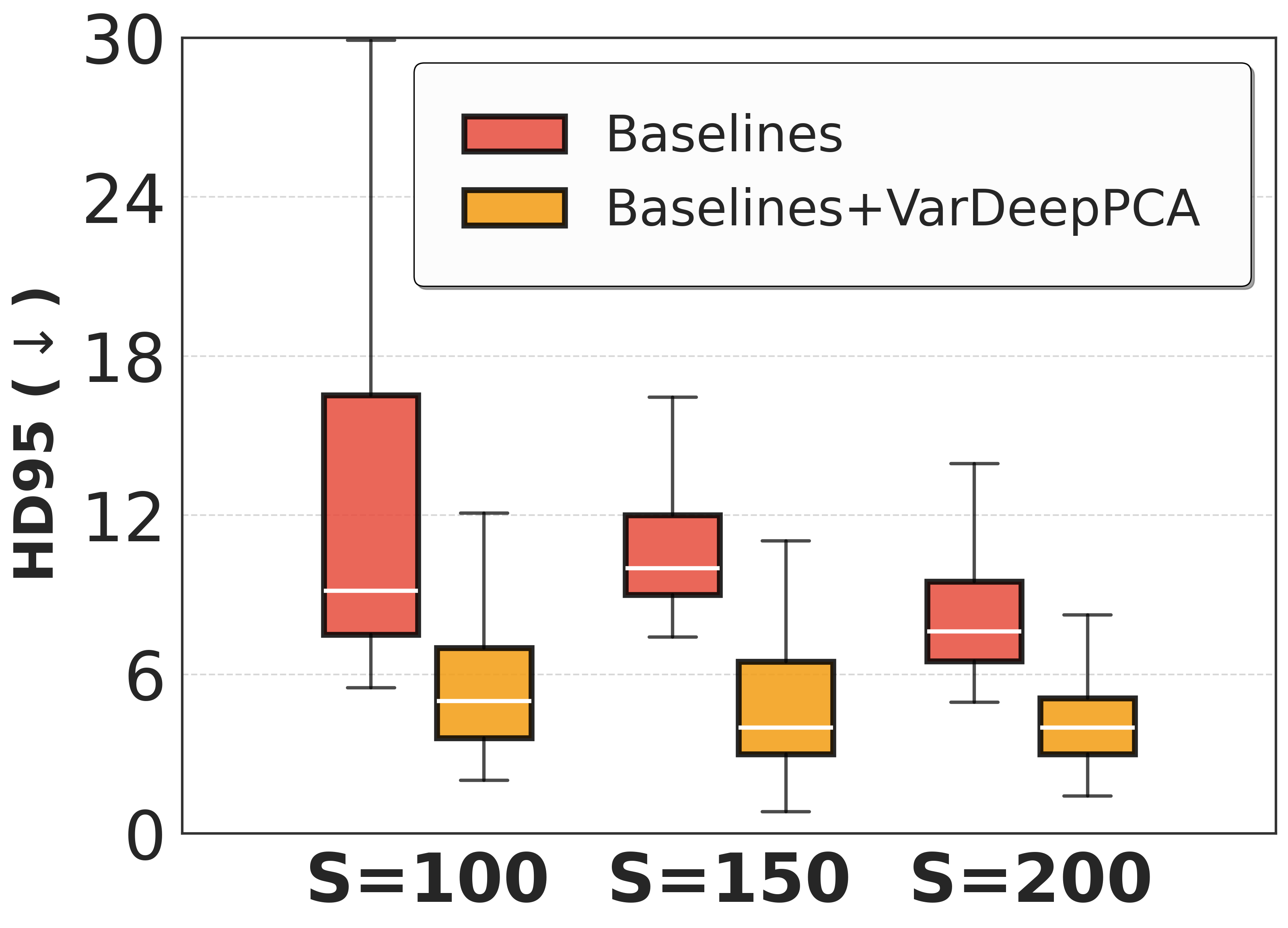}
{{\bf (a)~Myocardium}}{{\bf (e)~Myocardium}}{0.57}
%
\twoAcrossLabelsHeight[0]{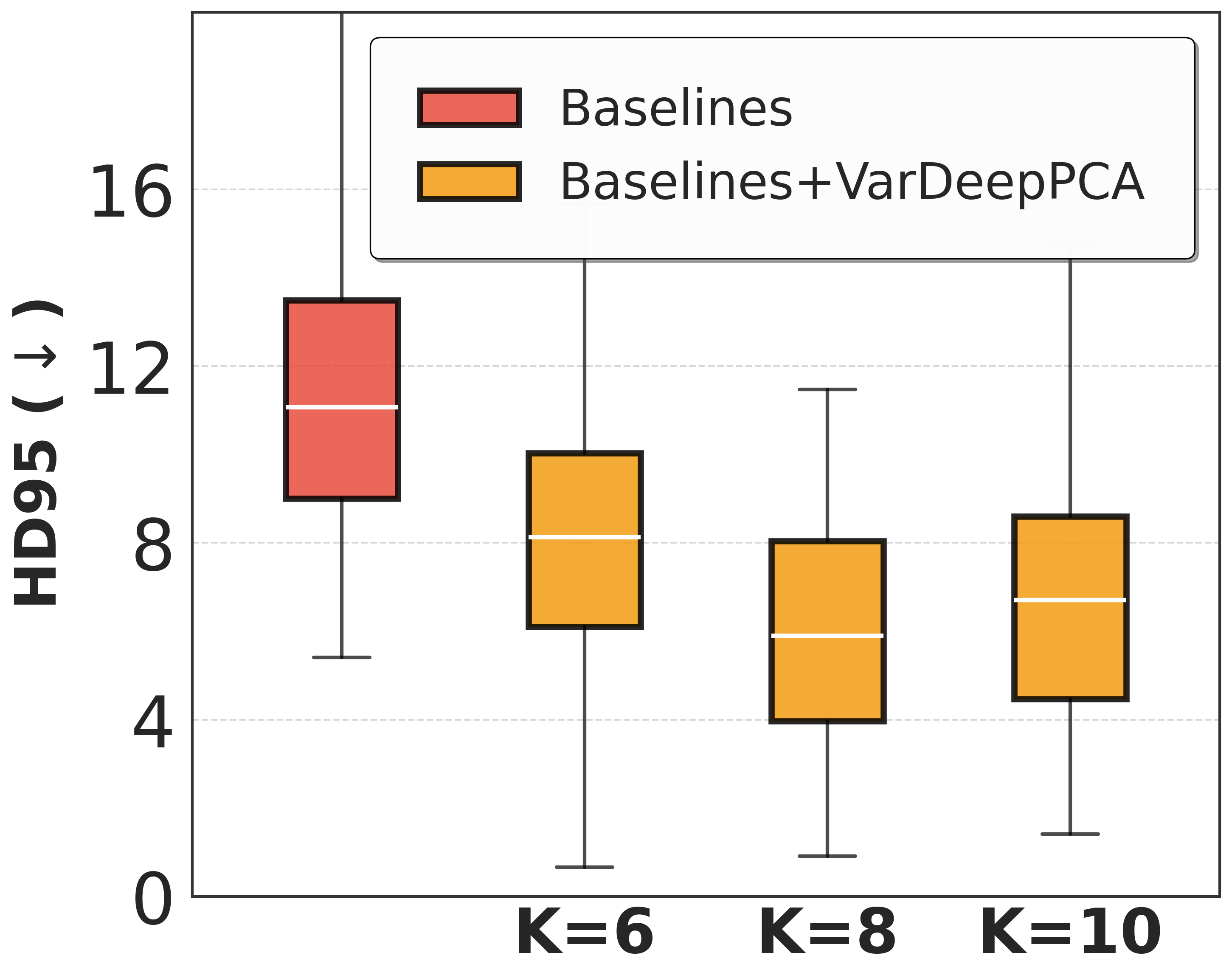}{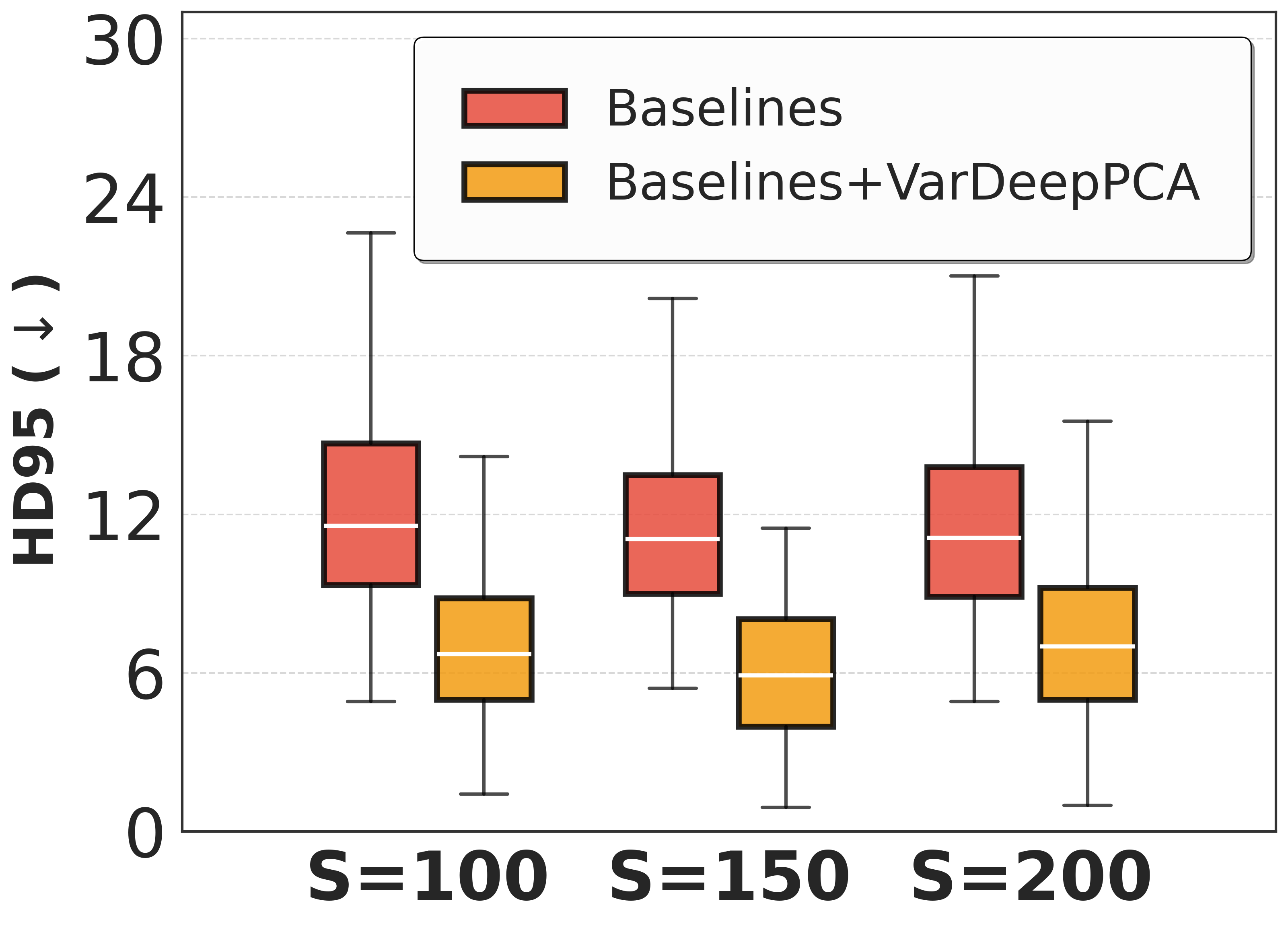}
{{\bf (b)~Neuroretinal Rim}}{{\bf (f)~Neuroretinal Rim}}{0.57}
%
\twoAcrossLabelsHeight[0]{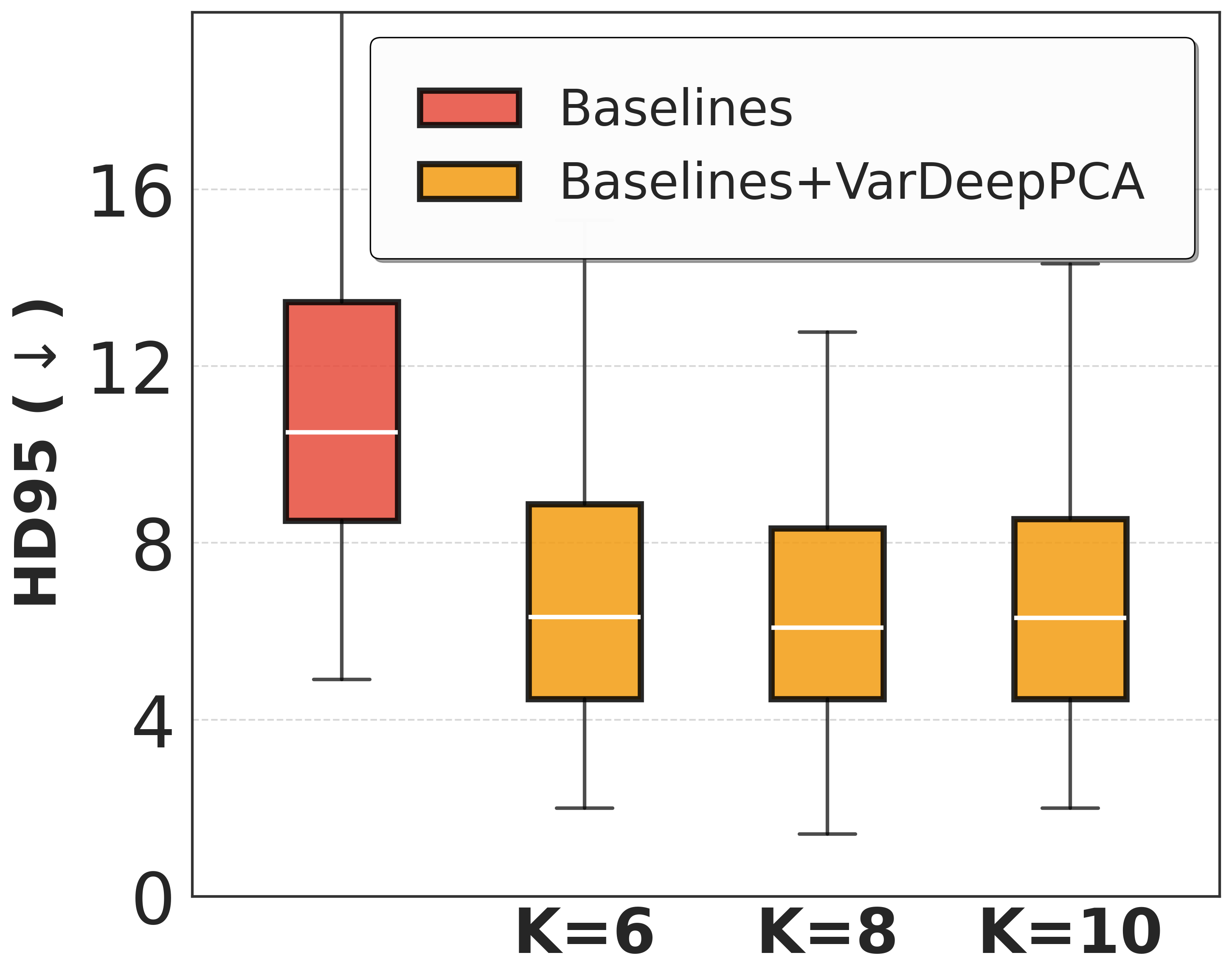}{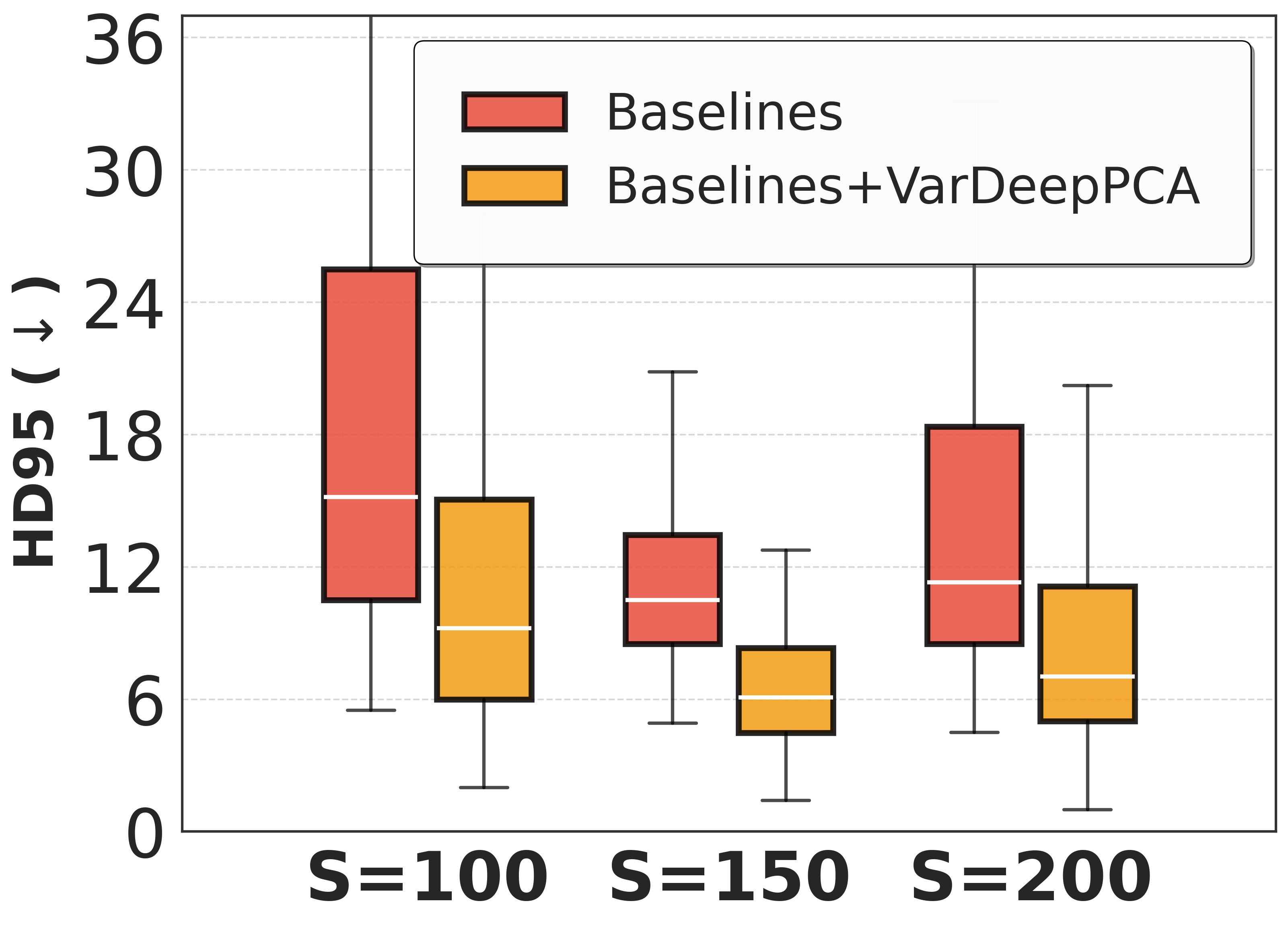}
{{\bf (c)~Prostate}}{{\bf (g)~Prostate}}{0.57}
%
\twoAcrossLabelsHeight[0]{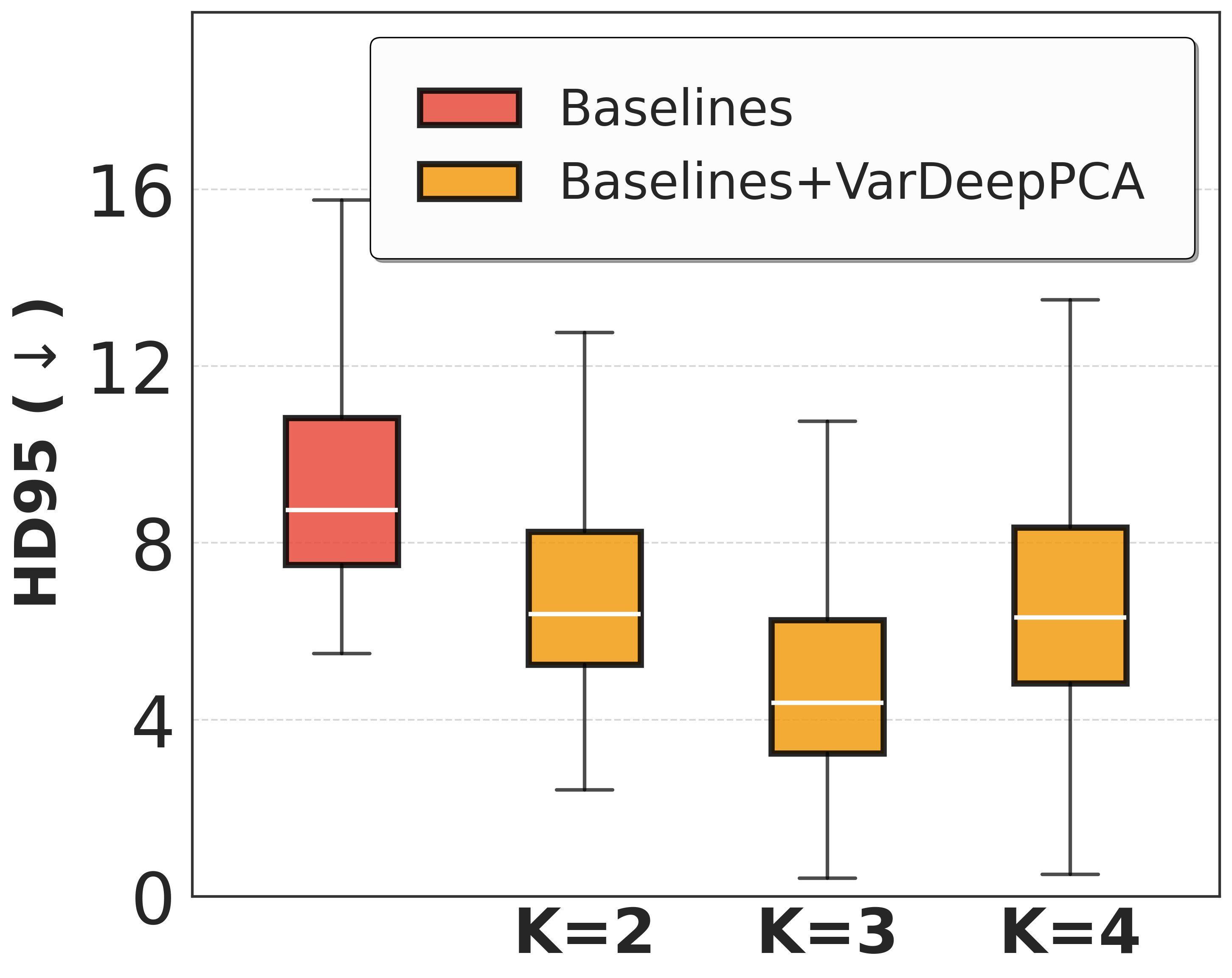}{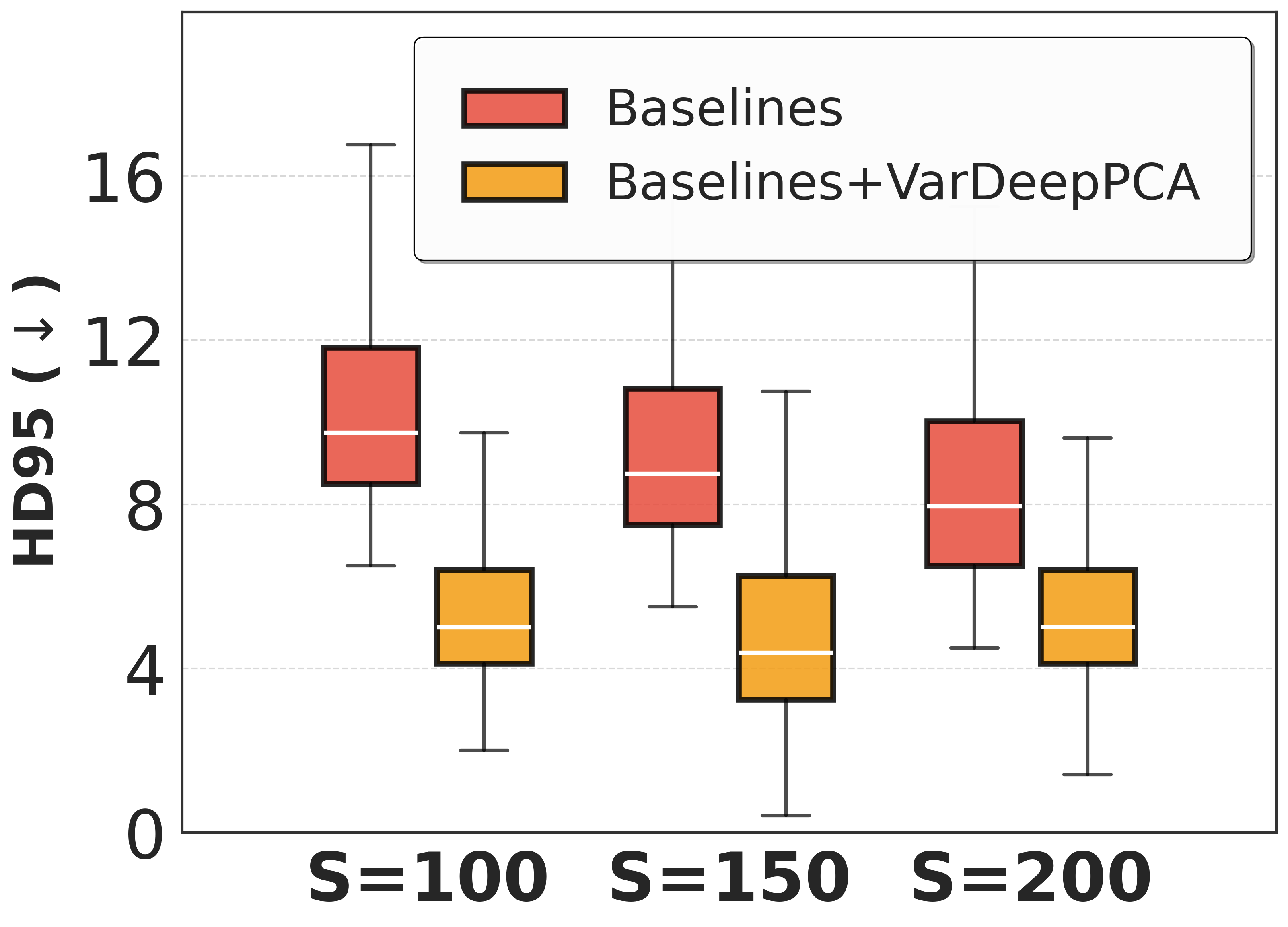}
{{\bf (d)~Fetal Head}}{{\bf (h)~Fetal Head}}{0.57}
\caption
{
{\bf Sensitivity of Results to Choice of Latent Dimension $K$ and Training Set Size $S$.}
We pooled results using three baselines (UNet, VMUNet, PHISeg) and their corresponding versions with
VarDeepPCA plugged in.
\textbf{(a)--(d)} Boxplots of HD95 values for varying latent dimension $K$.
\textbf{(e)--(h)} Boxplots of HD95 values for varying training-set sample size $S$.
HD95 values on pooled ID and OOD test sets for:
\textbf{(a,e)} Myocardium (CAP + ACDC);
\textbf{(b,f)} Neuroretinal Rim (MAGRABI + G1020);
\textbf{(c,g)} Prostate (BIDMC+BMC + HK+I2CVB);
\textbf{(d,h)} Fetal Head (HC18 + FetalPlanes).
}
\label{fig:sensitivity_boxplots}
\end{figure}

\subsection{Sensitivity Analysis}
\label{subsec:sen}

We evaluated the sensitivity of our VarDeepPCA plugin to the choice of the latent dimension $K$ and the
training set size $S$.
We selected VMUNet and PHISeg as the baselines specifically because our earlier experiments demonstrated their
relative robustness to OOD data in all of the applications. We also selected UNet as an early DNN method.
We report our analysis on a pooled test set comprising the ID test set and one representative OOD dataset for
each application.

{\bf Changing Latent Dimension $K$ Within VarDeepPCA.}
We trained different versions of VarDeepPCA (training set size $S = 150$) across a range of $K$ values for
each application.
For myocardium segmentation on CAP (ID) and ACDC (OOD) datasets, the baselines yielded a median HD95 of
approximately 10. Our VarDeepPCA models with $K \in \{ 12, 16, 20 \}$ consistently reduced this to a mean of
5--6, as shown in Figure~\ref{fig:sensitivity_boxplots}(a). We find the performance to be relatively
insensitive to to small changes in $K$.
We observed similar trends for all other datasets, i.e., 
for the neuroretinal rim segmentation on MAGRABI (ID) + ORIGA (OOD) datasets
(Figure~\ref{fig:sensitivity_boxplots}(b)),
for prostate segmentation on BIDMC+BMC (ID) + HK+I2CVB (OOD) datasets
(Figure~\ref{fig:sensitivity_boxplots}(c))
for fetal head segmentation on HC18 (ID) + FetalPlanes (OOD) datasets
(Figure~\ref{fig:sensitivity_boxplots}(d)).
Across all applications and all tested latent dimensions, our VarDeepPCA plugin consistently improved the
baseline segmentations, demonstrating significant robustness regardless of the specific $K$ value chosen.

{\bf Changing Size of Training Set $S$.}
To analyze sensitivity to training-data availability, we used training sets of three sizes: $S_{100}$,
$S_{150}$, and $S_{200}$. We designed them to be nested such that $S_{100} \subset S_{150} \subset S_{200}$.
The baselines (UNet, VMUNet, PHISeg) were trained on each subset and evaluated on the pooled ID and OOD test
set (Figure~\ref{fig:sensitivity_boxplots}(e)--(h)).
Across all applications and datasets, we find the performances of the methods to be quite insensitive to small
changes in $S$. Moreover, across all training-set sample sizes, our plugin consistently and significantly
improved over the baselines.

\section{Conclusion}
\label{sec:conclusion}

We have presented VarDeepPCA, a novel framework designed to restore degraded segmentation masks produced by
existing DNNs on OOD images.
By explicitly learning the principal modes of anatomical geometric variation from a small ID dataset of
segmentation maps, our framework neither requires access to any OOD data nor to any medical/acquired image.
Our novel variational learning framework leverages a reinterpretation of the softmax mapping to implicitly
perform exact distribution modeling, thereby enabling computationally efficient, sampling-free learning and
inference.
Our framework is characterized by a lightweight architecture (ranging from 1.02M to 2.72M parameters depending
on $K$), modality independence, and the ability to generate reliable/geometrically-consistent uncertainty
estimates.
Extensive empirical validation across 14 publicly available datasets and 15 DNN segmenters confirms that when
VarDeepPCA is plugged in to existing DNN methods, it consistently improves existing methods in OOD object
segmentation as well as uncertainty estimation. Indeed, the VarDeepPCA plugin may also be extended to improve
poor OOD-image segmentations resulting from traditional non-DNN methods.
While some of the baselines exhibit robustness sporadically on specific datasets, our framework provides a
much more reliable and consistent improvement in terms of boundary-delineation accuracy and the plausibility
of anatomical geometry.

We acknowledge a few constraints of our framework.
First, the method relies on consistency of geometry of the anatomical objects of interest; therefore, it is
inherently unsuitable for segmenting pathologies with highly non-uniform or amorphous/unpredictable
geometries, e.g., lesions and tumors.
%
%
Second, if a segmentation map (input to VarDeepPCA) lies within the learned distribution of valid geometries
but is still incorrect, then VarDeepPCA would be unable to improve the segmentation.
Finally, in scenarios where the error in the segmentation produced by an existing method is extremely large,
then our (or any such other) framework may be unable to restore the correct segmentation. Nevertheless, even
in such cases, VarDeepPCA succeeds in projecting the segmentation map to one with a valid anatomical geometry.
%
Future work may include extensions of VarDeepPCA to multi-class segmentation problems in 2D and 3D
images.
Future work may also explore the impact of patient demographics and clinical metadata on model
  generalization and uncertainty.


\acks{Supported by the Prime Minister's Research Fellowship from the Government of India.}

%
\ethics{The work follows appropriate ethical standards in conducting research and writing the manuscript,
following all applicable laws and regulations regarding treatment of animals or human subjects.}

\coi{We declare we don't have conflicts of interest.}

\data{To facilitate reproducibility, the source code, datasets, and pre-trained model weights will be made
publicly available upon publication.}

\bibliography{references_compressed}





\end{document}